%% file: Extension.tex
\documentclass[letterpaper]{article} 
\usepackage{aaai2026}  
\usepackage{times}  
\usepackage{helvet}  
\usepackage{courier}  
\usepackage[hyphens]{url}  
\usepackage{graphicx} 
\usepackage{natbib}  
\usepackage{caption} 
\usepackage{algorithm}
\usepackage{algorithmic}
\usepackage{amsmath}
\usepackage{booktabs}
\usepackage{amssymb}
\usepackage{multirow}
\usepackage{subfigure}
\usepackage{xcolor}
\usepackage{array}
\usepackage{tabularx}
\usepackage{ragged2e}
\usepackage{enumitem}

\usepackage{tabularx,array}
\newcolumntype{Y}{>{\raggedright\arraybackslash}X} 

\usepackage{cuted}      
\usepackage{capt-of}    

\usepackage{newfloat}
\usepackage{listings}
\DeclareCaptionStyle{ruled}{labelfont=normalfont,labelsep=colon,strut=off} 
\floatstyle{ruled}
\newfloat{listing}{tb}{lst}{}
\floatname{listing}{Listing}
\title{Mobile-Agent-RAG: Driving Smart Multi-Agent Coordination with Contextual Knowledge Empowerment for Long-Horizon Mobile Automation}
\author{
    Yuxiang Zhou$^{1,2,4}$\thanks{Equal Contribution.},
    Jichang Li$^{2*}$,
    Yanhao Zhang$^{3}$,
    Haonan Lu$^{3}$,
    Guanbin Li$^{1,2,4}$\thanks{Corresponding author is Guanbin Li.}
}
\affiliations{
    \textsuperscript{\rm 1}School of Computer Science and Engineering, Sun Yat-sen University\\
    \textsuperscript{\rm 2}Pengcheng Laboratory \\
    \textsuperscript{\rm 3}OPPO AI Center, OPPO Inc., China \\
    \textsuperscript{\rm 4}Guangdong Key Laboratory of Big Data Analysis and Processing \\
    
    zhouyx95@mail2.sysu.edu.cn, li.jichang@pcl.ac.cn, \{zhangyanhao,luhaonan\}@oppo.com, liguanbin@mail.sysu.edu.cn
}

\usepackage{bibentry}

\begin{document}

\maketitle

\begin{abstract}
Mobile agents show immense potential, yet current state-of-the-art (SoTA) agents exhibit inadequate success rates on real-world, long-horizon, cross-application tasks. We attribute this bottleneck to the agents' excessive reliance on static, internal knowledge within MLLMs, which leads to two critical failure points: 1) strategic hallucinations in high-level planning and 2) operational errors during low-level execution on user interfaces (UI). The core insight of this paper is that high-level planning and low-level UI operations require fundamentally distinct types of knowledge. Planning demands high-level, strategy-oriented experiences, whereas operations necessitate low-level, precise instructions closely tied to specific app UIs.
Motivated by these insights, we propose \texttt{Mobile-Agent-RAG}, a novel hierarchical multi-agent framework that innovatively integrates dual-level retrieval augmentation. At the planning stage, we introduce \textbf{Manager-RAG} to reduce strategic hallucinations by retrieving human-validated comprehensive task plans that provide high-level guidance. At the execution stage, we develop \textbf{Operator-RAG} to improve execution accuracy by retrieving the most precise low-level guidance for accurate atomic actions, aligned with the current app and subtask. To accurately deliver these knowledge types, we construct two specialized retrieval-oriented knowledge bases. Furthermore, we introduce \textbf{Mobile-Eval-RAG}, a challenging benchmark for evaluating such agents on realistic multi-app, long-horizon tasks. Extensive experiments demonstrate that 
 \texttt{Mobile-Agent-RAG} significantly outperforms SoTA baselines, improving task completion rate by 11.0\% and step efficiency by 10.2\%, establishing a robust paradigm for context-aware, reliable multi-agent mobile automation. 
\end{abstract}

\begin{links}
    \link{Code\&Supp.}{https://github.com/george13zyx/MAR}
\end{links}

\section{Introduction}

\begin{figure*}
    \centering

    \includegraphics[width=1.0\textwidth]{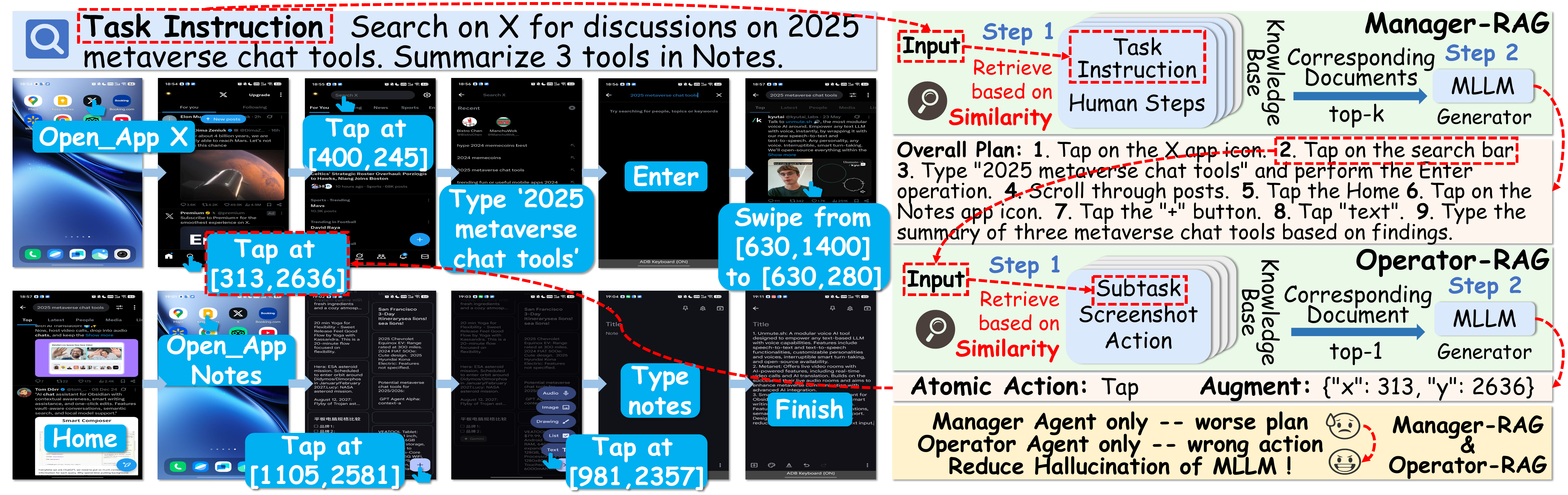}

    \captionof{figure}{\texttt{Mobile-Agent-RAG} is a novel hierarchical multi-agent framework explicitly designed for long-horizon, multi-app mobile automation tasks. \texttt{Mobile-Agent-RAG}'s proposed \textbf{Manager-RAG} retrieves human-verified task demonstrations to guide high-level plans, while \textbf{Operator-RAG} retrieves UI-grounded examples for atomic action generation.}
    \label{fig1}

\end{figure*}

Mobile agents have demonstrated immense potential in intelligent automation, promising to autonomously accomplish complex user tasks by interacting with smartphone interfaces \cite{you2024ferret,yan2023gpt}. However, current state-of-the-art (SoTA) agents still exhibit inadequate success rates when handling real-world, long-horizon, cross-application tasks \cite{bai2024hallucination}. We attribute this bottleneck to a fundamental issue: the agents' excessive reliance on static, internal knowledge embedded in multimodal large language models (MLLMs) \cite{tang2025survey,xie2025gui}. This reliance leads to two critical failure points: 1) strategic hallucinations in high-level planning requiring multi-step reasoning \cite{xie2025revealing,ji2024testing}, and 2) operational errors during low-level execution involving specific elements for user interfaces (UI) \cite{song2024visiontasker,guo2025maple}.

To address these challenges, hierarchical frameworks, such as \texttt{Mobile-Agent-E} \cite{wang2025mobile}, have been introduced, which decouple high-level planning from low-level execution. Although this architecture improves task performance to some extent, both planning and execution modules still depend on the inherent reasoning capabilities of the models themselves \cite{hou2024coact}. This dependence does not fundamentally resolve the hallucination issue caused by static knowledge, resulting in error accumulation during task execution. Retrieval-Augmented Generation (RAG) \cite{lewis2020retrieval} provides a new approach to mitigate this issue by dynamically retrieving information from external knowledge bases, which has shown success in domains \cite{yao2023react,feng2024enabling,zhu2024retrieval}. However, the effective application of RAG in mobile automotion to specifically address the aforementioned issues at the planning and execution levels remains unexplored.

The core insight of this paper is that high-level planning and low-level UI operations require fundamentally distinct types of knowledge. Planning demands high-level, strategy-oriented experiences, whereas operations necessitate low-level, precise instructions closely tied to specific app UIs. Motivated by these insights, we propose \texttt{Mobile-Agent-RAG}, a novel hierarchical multi-agent framework explicitly designed for long-horizon, multi-app mobile automation tasks. Our framework innovatively integrates dual-level retrieval augmentation, significantly boosting agent performance through real-time, contextually relevant external knowledge. At the planning stage, we introduce \textbf{Manager-RAG}, retrieving human-validated comprehensive task plans to provide high-level guidance for long-term strategies, effectively reducing strategic hallucinations. At the execution stage, we develop \textbf{Operator-RAG}, which retrieves most precise low-level guidance for accurate atomic actions aligned with the current app and subtask, substantially improving execution accuracy.

{To accurately deliver these knowledge types, we construct two specialized retrieval-oriented knowledge bases.} The framework dynamically leverages these two core RAG components to systematically overcome the limitations of current agents. Furthermore, we introduce \textbf{Mobile-Eval-RAG}, a challenging benchmark featuring realistic multi-app, long-horizon tasks. Extensive experiments demonstrate that our approach significantly surpasses existing state-of-the-art baselines, e.g. \texttt{Mobile-Agent-E}~{\cite{wang2025mobile}}, achieving notable improvements in task completion rates, operator accuracy, step efficiency, and success rates.

In summary, this paper makes the following contributions.
\begin{itemize}
    \item We propose \texttt{Mobile-Agent-RAG}, a hierarchical multi-agent framework enpowered with contextually relevant external knowledge through {dual-level retrieval augmentation}, namely {Manager-RAG} and {Operator-RAG},  for robust long-horizon mobile automation.
    \item  We design two retrieval-oriented knowledge bases, specialized to support {Manager-RAG} and {Operator-RAG}, for reducing agent hallucinations and execution errors, and release {Mobile-Eval-RAG}, a benchmark tailored for evaluating such retrieval-augmented mobile agents.
     \item Extensive experiments demonstrate that 
    our proposed \texttt{Mobile-Agent-RAG} significantly outperforms SoTA baseline algorithms, improving task completion rate by 11.0\% and step efficiency by 10.2\%, establishing a robust paradigm for context-aware, reliable multi-agent mobile automation.
\end{itemize}

\begin{figure*}[t]
\centering
\includegraphics[width=0.9\textwidth]{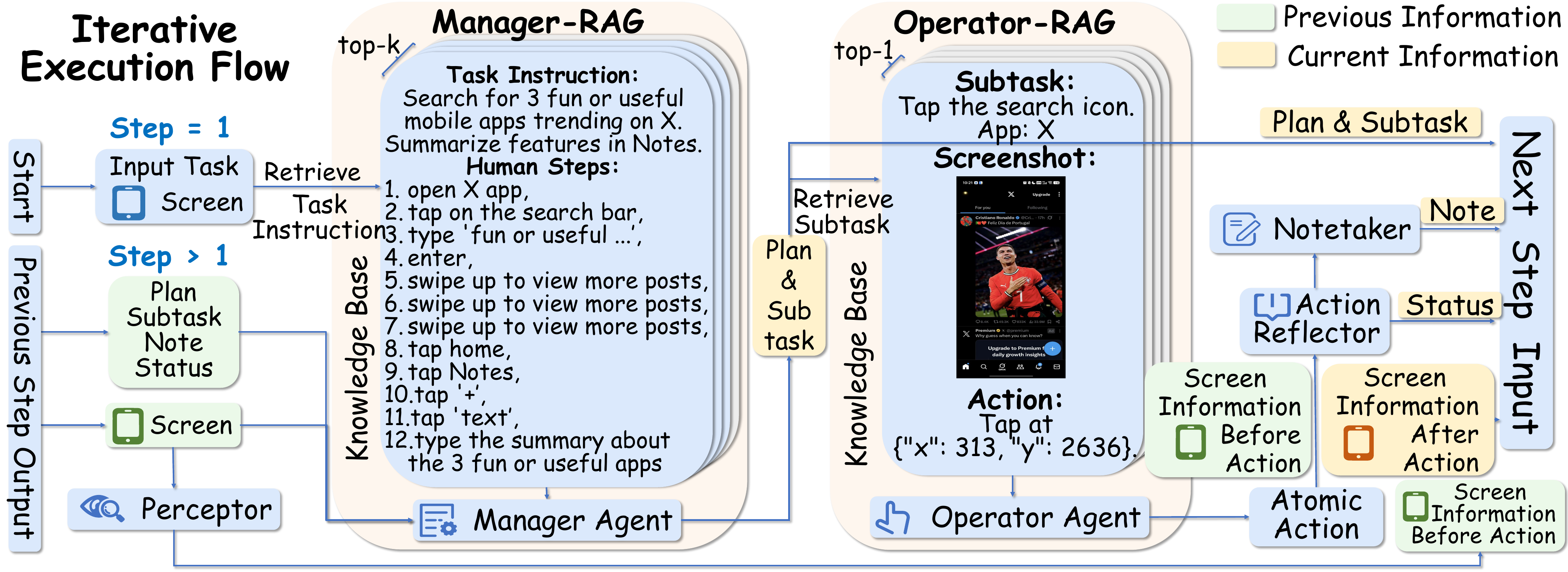}

\caption{{Framework overview of} \texttt{Mobile-Agent-RAG}: A hierarchical multi-agent system empowered by dedicated knowledge providers, namely \textbf{Manager-RAG} and \textbf{Operatoer-RAG}, for enhanced high-level planning and precise atomic action generation.  Its core operational loop is illustrated: Manager plans, Operator executes, with Perceptor, Action Reflector, and Notetaker providing comprehensive support for mobile task automation.}
\label{fig2}

\end{figure*}

\section{Related Work}

\paragraph{Mobile Agents for UI-Based Task Automation}
Early mobile task automation focused on enhancing agents' context understanding for graphic user interfaces (UI) through tools and exploratory actions, e.g., \texttt{Mobile-Agent} \cite{wang2024mobile} and \texttt{AppAgent} \cite{zhang2025appagent}. These single-agent systems struggled with complex tasks, prompting the shift to multi-agent approaches \cite{xie2024large}. For instance, \texttt{M3A} \cite{rawles2024androidworld} improves task completion by employing multiple agents for collaborative planning and decision-making. \texttt{Mobile-Agent-v2} \cite{wang2024mobile2} further decomposes tasks into planning, decision, and reflection agents. However, its planning component functions more as a tracker than a strategic planner, and decision-making spans multiple levels. Hence, MLLM-based agents like \texttt{DroidBot-GPT} \cite{wen2023droidbot} and \texttt{AutoDroid} \cite{wen2024autodroid} automate UI interactions by converting GUI states into natural language prompts. Additionally, \texttt{MobileGPT} \cite{lee2024mobilegpt} introduces a human-like memory system that modularizes tasks into reusable sub-modules via an ``explore, select, derive, recall'' model. Despite advancements, existing MLLM-based agents face hallucinations \cite{zheng2024thinking,li2025mitigating}, undermining their reliability in multi-step mobile tasks. Therefore, we instead propose \texttt{Mobile-Agent-RAG}, a retrieval-augmented generation framework that integrates external knowledge to improve context accuracy and reduce hallucinations.

\paragraph{Retrieval-Augmented Agents \textit{vs.} Memory-Driven Agents}
Retrieval-Augmented Generation (RAG) enhances MLLMs in knowledge-intensive tasks by combining parametric generation with non-parametric memory through semantic retrieval \cite{lewis2020retrieval}. In agent-based systems, RAG facilitates dynamic access to external knowledge during inference, improving factual consistency and planning \cite{singh2025agentic}. This is evident in web-based agents like \texttt{WebGPT} \cite{nakano2021webgpt} and reasoning-action frameworks like \texttt{ReAct} \cite{yao2023react}. Meanwhile, memory-driven agents, such as \texttt{MemGPT} \cite{packer2023memgpt}, support long-horizon tasks by retaining context across sessions, ensuring stable reasoning over time \cite{song2025towards}.
Recent mobile UI agents combine these techniques to address multi-step interactions. Representative works such as ~\texttt{AppAgent-v2} \cite{li2024appagent} builds a structured knowledge base during offline exploration for retrieval-driven decision-making, while \texttt{AppAgentX} \cite{jiang2025appagentx} logs execution history to enhance adaptability. 
Although \texttt{Mobile-Agent-E} \cite{wang2025mobile} improves task efficiency through agent-driven heuristic summarization, it remains susceptible to hallucinations and lacks explicit task abstraction. 
To overcome this challenge, our proposed \texttt{Mobile-Agent-RAG} integrates a hierarchical multi-agent framework, combining retrieval at both planning and execution levels. By utilizing both structured UI knowledge and external documents, this approach enhances context grounding, reduces hallucination risks, and supports reliable long-horizon mobile automation.

\section{Methodology}

This section details \texttt{Mobile-Agent-RAG}, a novel framework addressing long-horizon mobile automation by empowering a hierarchical multi-agent system with contextual knowledge. We first elaborate on its multi-agent architecture and execution flow. Subsequently, we detail how our proposed knowledge components enhance core agent decision-making, followed with their correpsonding knowledge base collection, providing a robust solution to challenges like hallucination. As illustrated in Figure \ref{fig2}, \texttt{Mobile-Agent-RAG} integrates our proposed dedicated knowledge providers, i.e. {\textbf{Manager-RAG}} and {\textbf{Operator-RAG}}, into the Manager and Operator agents, facilitating high-level task planning and low-level precise atomic actions. Notation definitions used in \texttt{Mobile-Agent-RAG} are provided in \underline{\textbf{Appendix A}}.

\subsection{Hierarchical Multi-Agent Architecture}

\texttt{Mobile-Agent-RAG} is built upon a {hierarchical multi-agent framework}, inherited and extended from \texttt{Mobile-Agent-E}~\cite{wang2025mobile}. This architecture is designed to achieve a clear separation between high-level planning and low-level execution in mobile task automation, ensuring robustness and adaptability across diverse mobile environments. The framework comprises a central decision-making loop facilitated by a \textbf{Manager Agent} and an \textbf{Operator Agent}, supported by several specialized modules including a \textbf{Perceptor}, an \textbf{Action Reflector}, and a \textbf{Notetaker}.

\subsubsection{Core Agents: Manager and Operator}

The interaction between the Manager and Operator constitutes the backbone of task execution, coordinating high-level strategic planning and low-level fine-grained operations.

\begin{itemize}

    \item \textbf{Manager Agent ($M$) {empowered by} \texttt{Manager-RAG}:} The Manager Agent is responsible for high-level strategic planning and subtask decomposition. At each timestep $t$, it devises the overall plan ($P_t$) to achieve the user’s task instruction ($I$) and generates the next subtask ($T_t^\text{app}$) with the relevant application.
    $$P_t, T_t^\text{app} =
    \begin{cases}
        M(I, D_{MR}, S_{t-1})\text{,} \quad \text{if } t=1 \\
        M(I, P_{t-1}, T_{t-1}^\text{app}, S_{t-1}, G_{t-1}, \\ N_{t-1}) \text{,} \quad \text{if } t>1 \text{ and } F_{t-1}=\text{False}\\
        M(I, P_{t-1}, T_{t-1}^\text{app}, S_{t-1}, G_{t-1},  L^E_{[-k:]}, \\ N_{t-1}) \text{,} \quad  \text{if } t>1 \text{ and } F_{t-1}=\text{True}\\
    \end{cases}$$
    At the initial timestep ($t=1$), the Manager leverages the input task instruction ($I$), a \textit{retrieved Manager-RAG document} ($D_{MR}$), and the initial screenshot ($S_{t-1}$) to formulate the first overall plan ($P_1$) and subtask ($T_1^\text{app}$). For subsequent timesteps ($t>1$), it refines its plan based on previous states. In case of consecutive errors ($F_{t-1}=\text{True}$), it consults recent error logs ($L^E_{[-k:]}$) for error recovery. The Manager primarily focuses on abstract planning, thus avoiding fine-grained visual information ($V_{t-1}$) that might introduce noise for high-level decision-making. \\

    \textbf{Operator Agent ($O$) {empowered by} \texttt{Operator-RAG}:} The Operator Agent translates the Manager's subtasks into concrete, executable atomic actions. It interacts directly with the mobile environment via Android Debug Bridge (ADB).
    $$
    \begin{aligned}
    A_t,S_t = O(&I, D_{OR}, P_t, T_t^\text{app}, S_{t-1}, V_{t-1}, \\
                &G_{t-1}, L^E_{[-k:]}, L^A_{[-k:]}, N_{t-1})
    \end{aligned}
    $$
    The Operator generates an atomic action ($A_t$) based on the current context, including the input task instruction ($I$), a \textit{retrieved Operator-RAG document }($D_{OR}$), the overall plan ($P_t$), the current subtask ($T_t^\text{app}$), the screenshot before action ($S_{t-1}$), and crucially, the fine-grained visual information before action ($V_{t-1}$). The combination of $S_{t-1}$ and $V_{t-1}$ enables precise action parameter generation (e.g., specific tap coordinates). The atomic actions $A_t$ are selected from a predefined set, including \textit{Open\_App}, \textit{Tap}, \textit{Swipe}, \textit{Type}, \textit{Enter}, \textit{Back}, \textit{Home}, and \textit{Wait}. Refer to \underline{\textbf{Appendix B}} for more details. \\
\end{itemize}

\subsubsection{Supporting Modules: Perceptor, Action Reflector, and Notetaker}

These modules augment the core Manager-Operator loop by providing essential information and feedback, ensuring the system's awareness and adaptability within the mobile environment.

\begin{itemize}
    \item \textbf{Perceptor ($P$): Fine-grained Visual Perception.} The Perceptor analyzes screenshots ($S_t$) to convert raw visual information into {structured, fine-grained information} ($V_t$). This information includes identified text, icons, their precise locations, and contextual descriptions, providing the crucial visual context for other agents. Refer to \underline{\textbf{Appendix C}} for details.
    $$V_t = P(S_t)$$

    \item \textbf{Action Reflector ($R$): Reflection on Action Outcome.} The Action Reflector evaluates the outcome of the Operator's most recent action ($A_t$), providing critical feedback to the system. It compares the UI states before ($S_{t-1}, V_{t-1}$) and after ($S_t, V_t$) the action, considering the current subtask ($T_t^\text{app}$) and global progress status ($G_{t-1}$). It classifies outcomes (e.g., successful, failed due to wrong page, failed due to no change) and updates the global progress status ($G_t$), action logs ($L^A_t$), and error logs ($L^E_t$). This feedback loop is vital for error detection and recovery.
    $$
    \begin{aligned}
        \text{Outcome}, G_t, L^A_t, L^E_t  =    R(&I, T_t^\text{app}, A_t, S_{t-1},\\
        &   S_t,  V_{t-1}, V_t, G_{t-1})
    \end{aligned}
    $$

    \item \textbf{Notetaker ($N$): Information Aggregation.} The Notetaker maintains and updates task-critical information ($N_t$) throughout execution. It aggregates dynamic details (e.g., search results, extracted phone numbers) from the current state ($S_t, V_t, G_t$) and previous notes ($N_{t-1}$). This ensures persistent tracking of information across multiple steps, which is crucial for long-horizon tasks to maintain contextual awareness.
    $$N_t = N(I, P_t, T_t^\text{app}, S_t, V_t, G_t, N_{t-1})$$
\end{itemize}

\subsubsection{Iterative Execution Flow}

As illustrated in Figure~\ref{fig2}, the system operates in a \textbf{dynamic iterative loop}, where agents and modules continuously interact to achieve the user's task instruction by processing information and executing actions. This process ensures adaptive responses to the mobile environment, evolving step-by-step toward task completion. The general flow at each timestep $t$ is as follows:

\begin{enumerate}
    \item \textbf{Perception:} The \textbf{Perceptor} module analyzes the current screenshot to extract fine-grained visual information.
    \item \textbf{High-Level Planning:} The \textbf{Manager Agent} uses raw screenshot, along with the overall task and contextual knowledge, to refine the global plan and determine the next high-level subtask.
    \item \textbf{Low-Level Action:} Based on the Manager Agent's subtask, the \textbf{Operator Agent} generates and executes a precise atomic UI action on the mobile device.
    \item \textbf{Post-Action Perception:} Immediately after an action is performed, the \textbf{Perceptor} captures the updated fine-grained visual information for Action Reflector to reflect the changes.
    \item \textbf{Reflection:} The \textbf{Action Reflector} evaluates the outcome of the executed action by comparing raw and fine-grained visual information before and after, providing critical feedback for progress tracking and error detection.
    \item \textbf{Information Aggregation:} Concurrently, the \textbf{Notetaker} module continuously updates and aggregates essential dynamic information as notes, maintaining a persistent and comprehensive context for the ongoing task.
\end{enumerate}

\texttt{Mobile-Agent-RAG} is built upon this hierarchical multi-agent architecture. To further enhance this framework, we introduce \textbf{Contextual Knowledge Empowerment}, a mechanism utilizing RAG to incorporate external, task-specific knowledge into both the Manager and Operator agents. By mitigating the limitations associated with exclusive reliance on internal representations from MLLMs, our approach significantly improves \texttt{Mobile-Agent-RAG}'s capability to address complex and previously unseen mobile tasks. Detailed implementation is described in the subsequent section.

\begin{figure*}[t]
\centering
\includegraphics[width=1.0\textwidth]{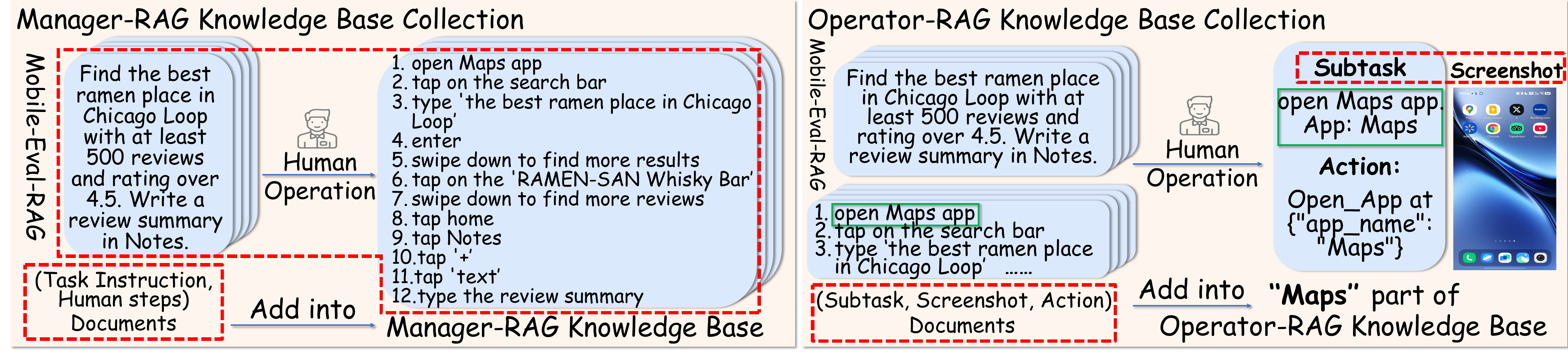}
\caption{A flow of knowledge base collection for Manager-RAG and Operater-RAG.}

\label{fig3}
\end{figure*}

\subsection{Contextual Knowledge Empowerment via RAG}

This section details how external knowledge is retrieved and integrated to empower the Manager and Operator agents, forming the core of \texttt{Mobile-Agent-RAG}'s innovation. This process allows the agents to leverage a rich repository of past experiences and domain-specific insights, significantly enhancing their performance beyond what is achievable with internal MLLM knowledge alone.

\begin{figure}[t]
    \centering
    \includegraphics[width=0.45\textwidth]{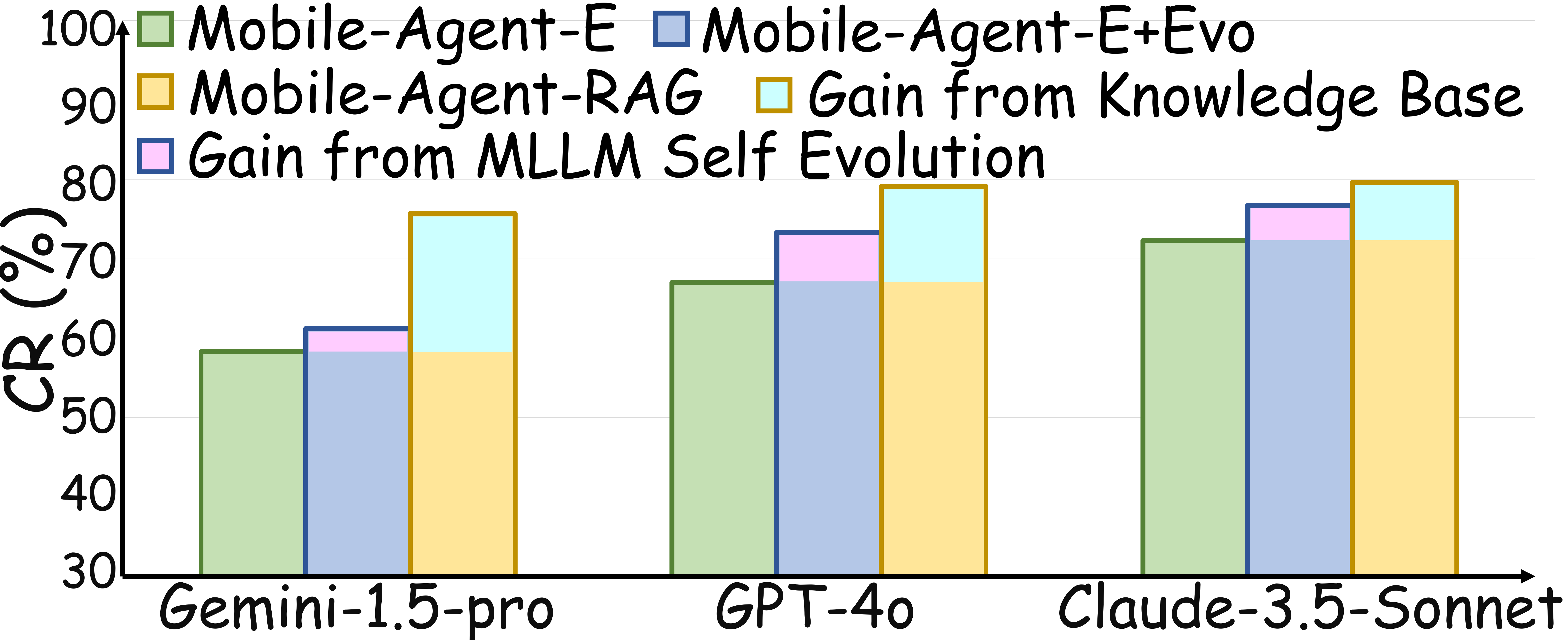}

    \caption{Comparison of performance gain sources for \texttt{Mobile-Agent-RAG}, \texttt{Mobile-Agent-E}, and \texttt{Mobile-Agent-E+Evo} across different MLLMs.}

    \label{figure4}
\end{figure}

\begin{table*}[t]
    \centering
    \small 

    \setlength{\tabcolsep}{4mm}
    \begin{tabular}{lcccccccccc}
        \toprule
        \textbf{Method} & \textbf{Framework Type} & {\textbf{CR}} & {\textbf{OA}} & {\textbf{RA}} & {\textbf{Steps}} & {\textbf{Efficiency}} & {\textbf{SR}} \\
        \midrule[\heavyrulewidth]
        \multicolumn{8}{c}{{Single-App Task Execution}} \\
        \midrule[\heavyrulewidth] 
        AutoDroid        & Single-Agent & 24.4 & 59.7 & {--} & 30.0 & 0.81 & 0.0  \\
        Appagent (Auto)  & Single-Agent & 25.4 & 63.5 & 91.0 & 30.0 & 0.85 & 0.0  \\
        Appagent (Demo)    & Single-Agent & 29.2 & 73.7 & 92.4 & 30.0 & 0.97 & 0.0  \\
        \midrule[\heavyrulewidth] 
        \multicolumn{8}{c}{{Multi-App Task Execution}} \\
        \midrule[\heavyrulewidth]
        Mobile-Agent     & Single-Agent & 33.7 & 60.3 & {--} & 29.0 & 1.16 & 12.0 \\
        Mobile-Agent-v2     & Multi-Agent & 38.5 & 61.9 & 92.5 & 23.5 & 1.64 & 44.0 \\
        Mobile-Agent-E   & Multi-Agent & 58.3 & 74.1 & 89.3 & 22.4 & 2.60 & 48.0 \\
        Mobile-Agent-E + Evo  & Multi-Agent & 61.2 & 77.2 & 91.0 & 21.8 & 2.81 & 56.0 \\
        Mobile-Agent-RAG (Ours)    & Multi-Agent & \textbf{75.7} & \textbf{90.1} & \textbf{94.7} & \textbf{18.8} & \textbf{4.03} & \textbf{76.0} \\
        \bottomrule
    \end{tabular}

    \caption{
    Comparison results of \texttt{Mobile}\texttt{-}\texttt{Agent}\texttt{-}\texttt{RAG} with previous SoTA algorithms  for single-app  and multi-app task execution with both single-agent and multi-agent frameworks. \textbf{Bold} indicates the best-performing results; the same applies hereafter. }
    \label{tab:exp_results}
             
\end{table*}

\subsubsection{Manager-RAG ($MR$)}

Manager-RAG aims to guide high-level planning, especially during the initial stages of a task.

\begin{itemize}
    \item \textbf{Retrieval:} At the beginning of a task ($t=1$), the Manager-RAG module takes the input task instruction ($I$) and retrieves the top-$k$ most relevant (task instruction $I_{MR}$, human steps $H_{MR}$) documents ($D_{MR}$) from its manually curated knowledge base ($K_{MR}$). These documents are selected based on the semantic similarity between the input task instruction and indexed task instructions, as determined by the Contriever-MSMARCO model \cite{izacard2022unsupervised}. The detailed retrieval algorithm is provided in Algorithm 1 in \underline{\textbf{Appendix D}}.

    \item \textbf{Generation:} The Manager-RAG module receives the overall task instruction $I_{\text{query}}$ and the retrieved set of $k$ relevant documents $\mathcal{R}_M$ from its retrieval mechanism. The MLLM generates the overall plan $P$ as follows:
    $$P = \text{PromptGen}_{\text{M}}(I_{\text{query}}, \mathcal{R}_M)$$
    where $\text{PromptGen}_{\text{M}}(\cdot)$ is a function that formulates the input prompt for the MLLM by combining $I_{\text{query}}$ with the retrieved examples in $\mathcal{R}_M$ as few-shot examples. 

\end{itemize}

\begin{table}[t]
    \centering
    \setlength{\tabcolsep}{1mm} 
    \small 

    \begin{tabularx}{\linewidth}{l *{6}{>{\centering\arraybackslash}X}}
        \toprule
        \textbf{Method} & \textbf{CR} & \textbf{OA } & \textbf{RA} & \textbf{Steps} & \textbf{Effic.} & \textbf{SR} \\
        \midrule[\heavyrulewidth]
        \multicolumn{7}{c}{Simple Operation Tasks} \\
        \midrule[\heavyrulewidth]
        Mobile-Agent-E & 63.4 & 81.8 & 89.7 & 20.3 & 3.12 & 80.0 \\
        Mobile-Agent-E + Evo & 68.3 & 85.9 & 86.3 & 16.8 & 4.07 & 80.0 \\
        Mobile-Agent-RAG (Ours) & \textbf{78.0} & \textbf{91.1} & \textbf{97.3} & \textbf{14.6} & \textbf{5.34} & \textbf{90.0} \\
        \midrule[\heavyrulewidth]
        \multicolumn{7}{c}{Complex Operation Tasks} \\
        \midrule[\heavyrulewidth]
        Mobile-Agent-E & 54.9 & 69.7 & 89.1 & 23.8 & 2.31 & 26.7 \\
        Mobile-Agent-E + Evo & 56.5 & 81.3 & 93.2 & 25.1 & 2.25 & 40.0 \\
        Mobile-Agent-RAG (Ours) & \textbf{74.2} & \textbf{89.6} & \textbf{93.5} & \textbf{21.6} & \textbf{4.30} & \textbf{66.7} \\
        \bottomrule
    \end{tabularx}

    \caption{
    Comparison resutls of \texttt{Mobile}\texttt{-}\texttt{Agent}\texttt{-}\texttt{RAG} with previous SoTA algorithms on tasks with
    varying complexity.
    }
    \label{tab:difficulty}

\end{table}

\subsubsection{Operator-RAG ($OR$)}

Operator-RAG is responsible for retrieving app-specific knowledge to support the generation of atomic actions required to accomplish a given subtask.

\begin{itemize}
    \item \textbf{Retrieval:} When the Operator is about to execute an atomic action, it sends the current subtask with the identified app name $T^{\text{app}}_{\text{query}}$ to Operator-RAG. The system then restricts retrieval to the app-specific knowledge base $K_{OR}^{\text{app}}$ for (subtask $T_{OR}$, screenshot $S_{OR}$, action $A_{OR}$) document ($D_{OR}$). The subtask is embedded via the same embedding function $f(\cdot)$ (Contriever-MSMARCO) into a vector representation. The system performs retrieval within the app-specific embedding space, selecting the entry with the highest cosine similarity between the query and indexed subtasks. The detailed retrieval algorithm is provided in Algorithm 2 in \underline{\textbf{Appendix D}}.

    \item \textbf{Generation:} For the Operator-RAG, the reasoning model receives the current subtask, a representation of the current screenshot, and the top-1 most relevant document $\mathcal{R}_O$ from its app-specific knowledge base. This detailed context allows the MLLM to accurately generate the correct atomic action needed to address the current subtask and the current screen state. The generation process can be expressed as:
    $$A = \text{PromptGen}_{\text{O}}(T^{\text{app}}_{\text{query}}, S_{\text{current}}, \mathcal{R}_O)$$
    where $\text{PromptGen}_{\text{O}}(\cdot)$ is a function that formulates the input prompt for the MLLM by integrating $T^{\text{app}}_{\text{query}}$, $S_{\text{current}}$, and the retrieved example $\mathcal{R}_O$. 

\end{itemize}

\begin{table}[t]
    \centering
    \small 
    \setlength{\tabcolsep}{1mm}
    \begin{tabularx}{\linewidth}{l *{6}{>{\centering\arraybackslash}X}}
        \toprule
        \textbf{Method} & \textbf{CR} & \textbf{OA} & \textbf{RA} & \textbf{Steps} & \textbf{Effic.} & \textbf{SR} \\
        \midrule[\heavyrulewidth]
        \multicolumn{7}{c}{Gemini-1.5-Pro} \\
        \midrule[\heavyrulewidth]
        Mobile-Agent-E & 58.3 & 74.1 & 89.3 & 22.4 & 2.60 & 48.0 \\
        Mobile-Agent-E + Evo & 61.2 & 77.2 & 91.0 & 21.8 & 2.81 & 56.0 \\
        Mobile-Agent-RAG (Ours) & \textbf{75.7} & \textbf{90.1} & \textbf{94.7} & \textbf{18.8} & \textbf{4.03} & \textbf{76.0} \\
        \midrule[\heavyrulewidth]
        \multicolumn{7}{c}{GPT-4o} \\
        \midrule[\heavyrulewidth]
        Mobile-Agent-E & 67.0 & 81.5 & 90.2 & 18.6 & 3.60 & 60.0 \\
        Mobile-Agent-E + Evo & 73.3 & 85.6 & 92.4 & \textbf{17.8} & 4.12 & 76.0 \\
        Mobile-Agent-RAG (Ours) & \textbf{79.1} & \textbf{88.4} & \textbf{97.8} & 18.4 & \textbf{4.30} & \textbf{84.0} \\
        \midrule[\heavyrulewidth]
        \multicolumn{7}{c}{Claude-3.5-Sonnet} \\
        \midrule[\heavyrulewidth]
        Mobile-Agent-E & 72.3 & 91.0 & 94.8 & 17.3 & 4.18 & 60.0 \\
        Mobile-Agent-E + Evo & 76.7 & 91.3 & \textbf{95.6} & \textbf{17.1} & \textbf{4.49} & 72.0 \\
        Mobile-Agent-RAG (Ours) & \textbf{79.6} & \textbf{92.7} & 95.1 & 18.7 & 4.26 & \textbf{84.0} \\
        \bottomrule
    \end{tabularx}

    \caption{
    Comparison results of \texttt{Mobile}\texttt{-}\texttt{Agent}\texttt{-}\texttt{RAG} with previous SoTA algorithms, evaluated by different MLLMs.
    }
    \label{tab:model_comparison}

\end{table}

\subsection{Retrieval-Oriented Knowledge Base Collection}

To enable retrieval-augmented reasoning for both high-level task planning and low-level action execution, we construct dedicated knowledge bases for Manager-RAG and Operator-RAG, as illustrated in Figure~\ref{fig3}. By combining automated logging with human validation, our data collection pipeline produces high-quality, context-rich supervision signals that effectively ground model inference. For additional technical details, please refer to \underline{\textbf{Appendix E}}.

\paragraph{Manager-RAG Knowledge Base ($K_{MR}$)}
This knowledge base is designed to support high-level task planning. Each entry in the knowledge base consists of a natural language task instruction paired with its corresponding human-annotated operation steps. These (task instruction, human steps) documents ($D_{MR}$), are directly extracted from the \textbf{Mobile-Eval-RAG} dataset \textit{we construct below}. For each task in Mobile-Eval-RAG, the task instruction is treated as a retrieval query, and the human execution trace is simplified and structured into reference steps for downstream use by MLLMs. All data is manually collected and curated to ensure task-level coherence and accuracy.

\paragraph{Operator-RAG Knowledge Base ($K_{OR}^{\text{app}}$)} 
This knowledge base  is intended to support atomic action generation during low-level app interactions. Its construction follows a semi-automated process. During agent execution, we dynamically log three types of information: the current subtask being performed, the corresponding screenshot at that step, and the atomic action generated by the Operator. These (subtask, screenshot, action) documents ($D_{OR}$) are collected in real time and subsequently verified by human annotators. Invalid or low-quality entries are manually corrected or discarded, ensuring data quality and consistency. To avoid cross-app interference and enhance retrieval precision, we maintain separate retrieval libraries for each mobile application.

\section{Experiments}

This section develops the evaluation benchmark \textbf{Mobile-Eval-RAG} and designs experiments to thoroughly assess and analyze the performance, module roles, and generalization capability of \texttt{Mobile-Agent-RAG}. Due to space constraints, further details on the experiments, e.g. case study and more analysis, can be found in \underline{\textbf{Appendix F-M}}.

\subsection{Benchmark Dataset: Mobile-Eval-RAG}

We introduce \textbf{Mobile-Eval-RAG}, a benchmark dataset of 50 diverse and challenging tasks designed specifically to evaluate Retrieval-Augmented Generation (RAG) performance in mobile agent systems. Unlike existing benchmarks such as Mobile-Eval-E \cite{wang2025mobile}, which have limited suitability for assessing RAG's generalization due to progressively increasing difficulty and insufficient task similarity, Mobile-Eval-RAG emphasizes cross-application coordination and long-horizon planning.
Specifically, Mobile-Eval-RAG expands upon Mobile-Eval-E’s five core categories—Information Searching, What’s Trending, Restaurant Recommendation, Online Shopping, and Travel Planning—by introducing cross-application tasks averaging 16.9 steps across 2-3 applications. These tasks closely reflect realistic scenarios demanding substantial coordination and planning. Additionally, region-specific constraints of application have been removed to enhance general applicability. Queries are MLLM-generated and human-verified to ensure feasibility and consistency. Tasks are divided into simple operations (Information Searching, What’s Trending), involving basic searches, and complex operations (Restaurant Recommendation, Online Shopping, Travel Planning), requiring detailed interactions and multi-step coordination, maintaining consistent difficulty levels within each category. 
In summary, Mobile-Eval-RAG effectively evaluates retrieval alongside long-horizon planning and inter-task coordination, providing a concise yet comprehensive platform for assessing RAG models in complex, multi-app environments.

\subsection{Experimental Setups}

\paragraph{Baselines}  

To evaluate \texttt{Mobile-Agent-RAG}, we compare it with state-of-the-art open-source mobile agent frameworks, including \texttt{AutoDroid}~\cite{wen2024autodroid}, \texttt{AppAgent (Auto)}~\cite{zhang2025appagent}, \texttt{AppAgent (Demo)}~\cite{zhang2025appagent}, \texttt{Mobile-Agent}~\cite{wang2024mobile}, \texttt{Mobile-Agent-v2}~\cite{wang2024mobile2}, \texttt{Mobile-Agent-E}~\cite{wang2025mobile}, and \texttt{Mobile-Agent-E+Evo}~~\cite{wang2025mobile}. These frameworks include both single-agent and multi-agent architectures, all utilizing MLLMs for mobile task automation. Specifically, \texttt{Mobile-Agent-E+Evo} enhances \texttt{Mobile-Agent-E} by integrating a ``Self-Evolution'' strategy, while \texttt{AppAgent (Auto)} and \texttt{AppAgent (Demo)} represent two variants proposed by ~\cite{zhang2025appagent} for autonomous exploration and human-demonstration modes, respectively.

\paragraph{Reasoning Models}  
We use several MLLMs as reasoning backbones in our framework, including \textbf{GPT-4o}~\cite{hurst2024gpt}, \textbf{Claude-3.5-Sonnet}~\cite{anthopic2024introducing}, and \textbf{Gemini-1.5-Pro}~\cite{team2024gemini}.
{Among them, \textit{Gemini-1.5-Pro} achieves the best performance–cost balance, and we use it as the default backbone for efficiency and scalability.
}

\paragraph{Evaluation Metrics}

To assess \texttt{Mobile-Agent-RAG}'s mobile agent performance on complex tasks, we first use the standard evaluation metrics from \texttt{Mobile-Agent-E} {and \texttt{Mobile-Agent-v2}}, including \textbf{Success Rate} (\textbf{SR}, \%), \textbf{Completion Rate} (\textbf{CR}, \%), \textbf{Operator Accuracy} (\textbf{OA}, \%), and \textbf{Reflector Accuracy} (\textbf{RA}, \%), \textcolor{black}{and then newly add }\textbf{Steps} and \textbf{Efficiency} \textcolor{black}{to evaluate task execution efficiency}; Refer to \underline{\textbf{Appendix G}} for details.

        
        


\begin{figure}[t]
    \centering
    \includegraphics[width=0.48\textwidth]{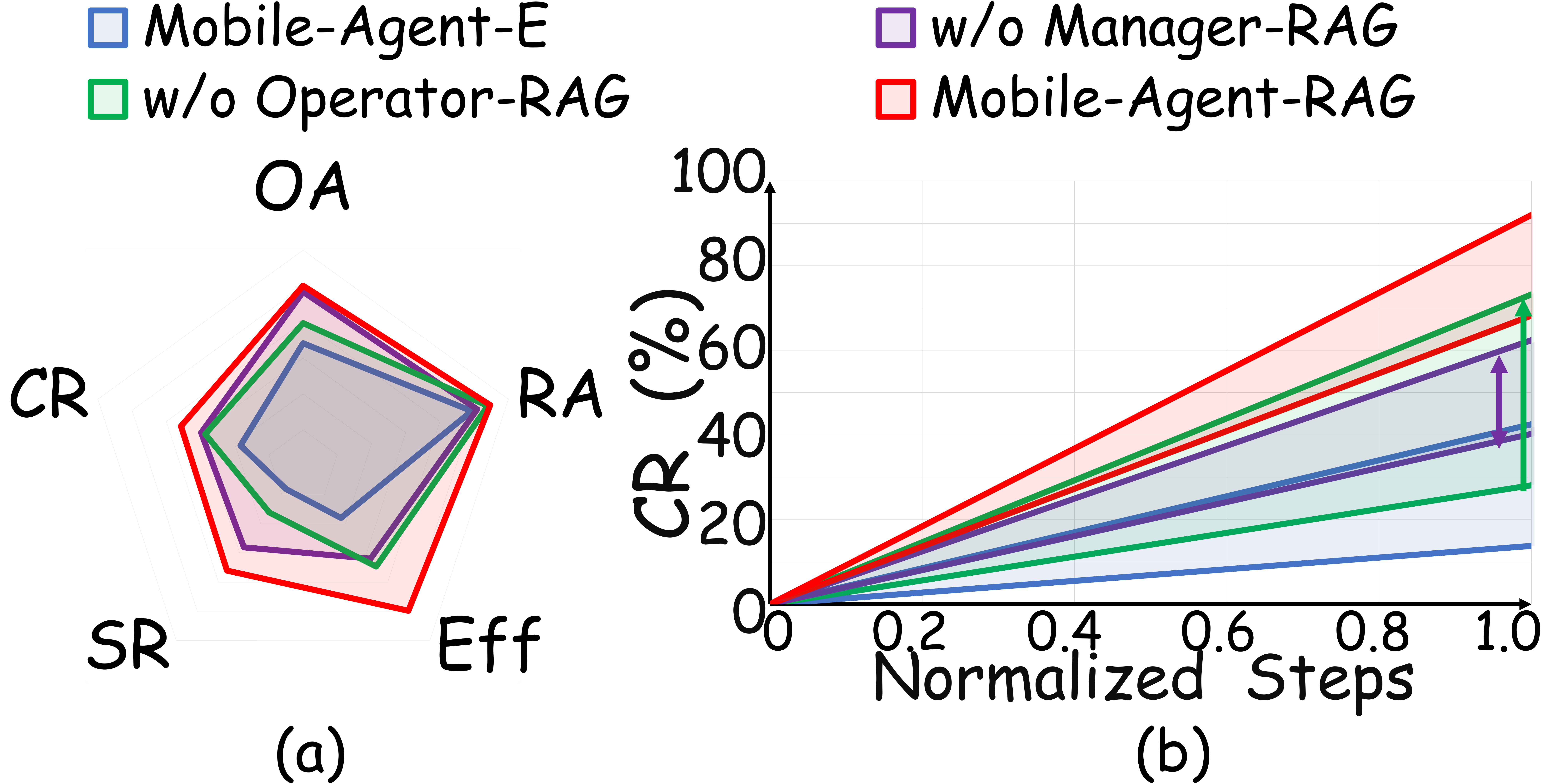}
    \caption{
    Ablation study results of \texttt{Mobile-Agent-RAG}, with (a) showing a radar chart comparing evaluation metrics across different ablation variants, and (b) presenting CR variations over 10 trials for each variant.
    }
    \label{fig:ablation}
\end{figure}

\begin{figure}[t]
    \centering
    \includegraphics[width=0.48\textwidth]{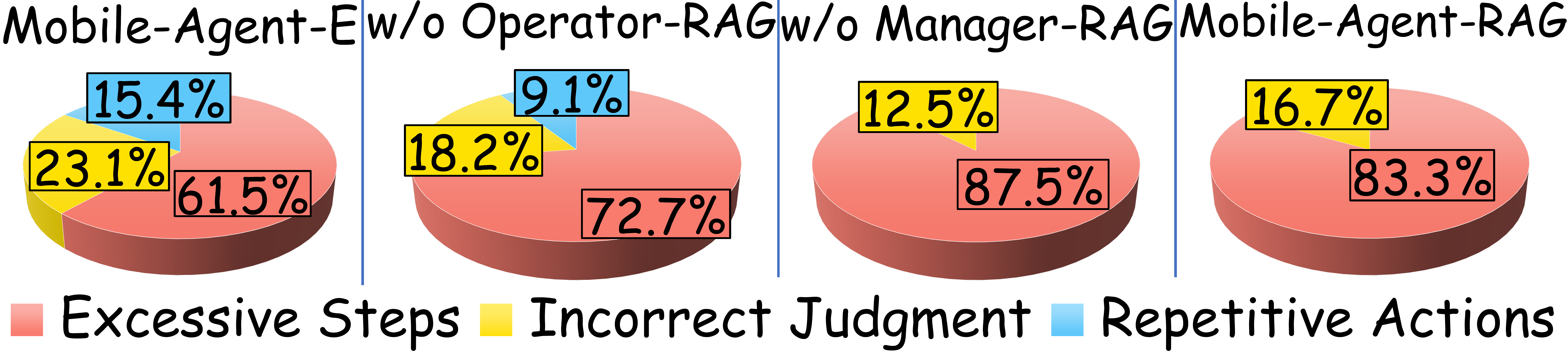}
    
    \caption{Error type distribution over three task failure modes corresponding to the three evaluation criteria of SR.}
    
    \label{fig:error}
\end{figure}

\subsection{Main Results}

To evaluate \texttt{Mobile-Agent-RAG}, we first benchmarked it against SoTA single- and multi-app frameworks, as summarized in Table~\ref{tab:exp_results}. Results show that \texttt{Mobile-Agent-RAG} consistently outperforms baselines, particularly in complex multi-app tasks, primarily due to accurate human-annotated knowledge rather than less reliable self-generated summaries. Furthermore, we assessed its robustness across tasks of varying complexity in Table~\ref{tab:difficulty}. While methods like \texttt{Mobile-Agent-E+Evo} yield modest CR improvements of 7.7\% on simple and 2.9\% on complex tasks, our approach significantly boosts CR by 23.0\% (simple) and 35.2\% (complex) through effective knowledge augmentation. 
Furthermore, to confirm \texttt{Mobile-Agent-RAG}'s generalization across multiple MLLMs, we conducted additional experiments shown in Table~\ref{tab:model_comparison}. 
Compared to \texttt{Mobile-Agent-E}, \texttt{Mobile-Agent-RAG} achieves a 23.6\% CR gain on weaker models (Gemini-1.5-Pro) and a 5.8\% advantage on stronger ones (GPT-4o \& Claude-3.5-Sonnet). Finally, Figure~\ref{figure4} confirms that the RAG's benefit negatively correlates with model strength—providing greater compensation for limited reasoning in weaker MLLMs while continuing to improve performance on more capable models.

\subsection{Ablation Analysis}

We conducted an ablation study on \texttt{Mobile-Agent-RAG} to examine the roles of its two core modules, \textbf{Manager-RAG} and \textbf{Operator-RAG}. Results in Figures \ref{fig:ablation}(a), \ref{fig:ablation}(b) and \ref{fig:error} show that removing Operator-RAG significantly lowers OA, Efficiency, and SR by increasing repetitive and erroneous low-level actions. Conversely, removing Manager-RAG reduces maximum achievable CR, highlighting its essential role in long-horizon planning, although this benefit requires accurate execution provided by Operator-RAG. Further analysis illustrated by Figure~\ref{fig:error} confirms their complementarity: Manager-RAG focuses on global task decomposition and provides a more concise strategy, while Operator-RAG ensures precise and context-aware task execution, reducing repetitive local errors. Together, they enable effective mobile automation for complex tasks.

\section{Conclusions}

We present \texttt{Mobile-Agent-RAG}, a hierarchical multi-agent framework that introduces dual-level retrieval-augmented generation (RAG) to address the core challenges of long-horizon, multi-app mobile automation. Our approach mitigates the limitations of conventional mobile agents, which often suffer from strategic hallucinations during high-level planning and operational errors in low-level execution due to static MLLM knowledge. By combining Manager-RAG for high-level, human-validated strategic planning and Operator-RAG for low-level, app-specific action grounding, our framework achieves robust and context-aware decision-making across both planning and execution layers. To support this design, we constructed two dedicated retrieval-oriented knowledge bases and released Mobile-Eval-RAG, a challenging benchmark dataset of realistic, long-horizon, multi-app tasks. Extensive experiments demonstrate that \texttt{Mobile-Agent-RAG} achieves significant gains in task completion rate, operator accuracy, and step efficiency over the state-of-the-art baselines, validating dual-level contextual knowledge empowerment for multi-agent coordination.

\section{Acknowledgments}

This work is supported in part by the National Key R\&D Program of China (2024YFB3908503), and in part by the National Natural Science Foundation of China (62322608). 

\bibliography{aaai2026}

\input{Supplementary}

\end{document}

%% file: Supplementary.tex
\appendix

\section*{Appendix Overview}
\label{appendix:overview}

This appendix provides comprehensive details and additional analyses that complement our main paper on the proposed \texttt{Mobile-Agent-RAG} framework. We offer an in-depth look into the technical components, evaluation metrics, and a detailed case study to provide a holistic understanding of the proposed algorithm. Specifically, we present the full list of notations, a description of the Operator Agent's action space, detailed implementation specifics for our multi-agent system, the algorithms underpinning our retrieval mechanisms, and a thorough explanation of our knowledge base construction process. We also provide a complete list of the tasks in the Mobile-Eval-RAG benchmark, the full completion criteria used for task completion rate evaluation, and an expanded analysis of a case study demonstrating the \texttt{Mobile-Agent-RAG}'s effectiveness.
Specifically, it includes:
\begin{itemize}
    \item \textbf{A. Notation Definitions}
    \item \textbf{B. Further Details on Experimental Setups}
    \item \textbf{C. Further Details on the Hierarchical Multi-agent Framework}
    \item \textbf{D. Retrieval Algorithms for Manager-RAG and Operator-RAG}
    \item \textbf{E. Further Details for Retrieval-Oriented Knowledge Base Collection}
    \item \textbf{F. Further Details for Mobile-Eval-RAG Construction}
    \item \textbf{G. Further Details on Evaluation Metrics}
    \item \textbf{H. Completion Rate Evaluation Criteria}
    \item \textbf{I. Further Details on Experimental Implementations}
    \item \textbf{J. Computational Cost}    
    \item \textbf{K. Additional Ablation Study on Core
Components}
    \item \textbf{L. Case Study}
    \item \textbf{M. More Analysis and Limitations}
\end{itemize}
\vspace{0.5em}
\noindent We provide tables, algorithms, and figures in-place to keep each section self-contained for replication.

\section{A. Notation Definitions}
\label{appendix:notation_definitions}

This appendix provides the main notations used to describe the \texttt{Mobile-Agent-RAG} framework and its components. The definitions, presented in Table~\ref{tab:notation_appendix}, clarify the roles of each element and the information flow within our hierarchical multi-agent system for robust long-horizon multi-app mobile automation tasks.

\begin{table}[t!]
\vskip 0.1in
\begin{center}
\begin{small}
\centering
\setlength{\tabcolsep}{4pt} 
\begin{tabularx}{\linewidth}{@{}l Y@{}}

\toprule
\textbf{Notation} & 
\textbf{Description} \\
\midrule
\multicolumn{2}{l}{\textit{Environment}} \\
\midrule
$I$ & User task instruction \\
$A_t$ & Atomic action executed at timestep $t$ \\
$S_t$ & UI screenshot (Raw visual information) captured at timestep $t$ \\
\midrule
\multicolumn{2}{l}{\textit{Agents}} \\
\midrule
$M$ & Manager Agent\\
$O$ & Operator Agent\\
$P$ & Perceptor\\
$R$ & Action Reflector\\
$N$ & Notetaker\\
\midrule
\multicolumn{2}{l}{\textit{Working Memory}} \\
\midrule
$V_t$ & Fine-grained visual information from $P$ at timestep $t$\\
$P_t$ & Overall plan at timestep $t$\\
$T_t^\text{app}$ & Current subtask with identified app name at timestep $t$\\
$G_t$ & Progress status at timestep $t$\\
$N_t$ & Important notes at timestep $t$\\
$F_t$ & Consecutive-error flag at timestep $t$\\
$L^A$ & Action logs with outcome status\\
$L^E$ & Error logs with feedback\\
\midrule
\multicolumn{2}{l}{\textit{Retrieval-Augmented Components}} \\
\midrule
$MR$ & Manager-RAG\\
$OR$ & Operator-RAG\\
$K_{MR}$ & Manager-RAG knowledge base: set of $D_{MR}$ with $(I_{MR}, H_{MR})$ pairs\\
$K_{OR}^{\text{app}}$ & App-specific Operator-RAG knowledge base : set of $D_{OR}$ with $(T_{OR}^{\text{app}}, S_{OR}, A_{OR})$ triplets\\

$D_{MR}$ & A document with $(I_{MR}, H_{MR})$ pair in $K_{MR}$ \\
$D_{OR}$ & A document with $(T_{OR}^{\text{app}}, S_{OR}, A_{OR})$ triplet in $K_{OR}^{\text{app}}$ \\

$n_{MR}$ & The number of document $D_{MR}$ in $K_{MR}$\\
$n_{OR}^{\text{app}}$ & The number of document $D_{OR}$ in one of the $K_{OR}^{\text{app}}$\\

$I_{MR}$ & Task instruction text in $D_{MR}$ \\
$H_{MR}$ & Human operation steps in $D_{MR}$ \\

$T_{OR}^{\text{app}}$ & Subtask text in $D_{OR}$ \\
$S_{OR}$ & Reference screenshot in $D_{OR}$ \\
$A_{OR}$ & Atomic action (with arguments) in $D_{OR}$ \\
$\mathcal{R}_M$ & Top-$k$ retrieved $(I_{MR}, H_{MR})$ for $MR$ \\
$\mathcal{R}_O$ & Top-1 retrieved $(T_{OR}^{\text{app}}, S_{OR}, A_{OR})$ for $OR$ \\
$f(\cdot)$ & Text encoder for embeddings by Contriever-MSMARCO \\

\bottomrule
\end{tabularx}
\end{small}
\end{center}
\caption{Symbols used in \texttt{Mobile-Agent-RAG}, grouped by \emph{Environment}, \emph{Agents}, 
\emph{Working Memory}, and \emph{Retrieval-Augmented Components}. Index $t$ denotes the interaction timestep.
$\mathcal{R}_M$ returns top-$k$ matches from $K_{MR}$, while $\mathcal{R}_O$ returns the top-1 match within the active app-specific $K_{OR}^{\text{app}}$.
Embeddings $f(\cdot)$ are computed with Contriever-MSMARCO.}

\label{tab:notation_appendix}
\end{table}

\section{B. Further Details on Experimental Setups}

To ensure a fair and consistent experimental setting with previous work \texttt{Mobile-Agent-E}, we synchronize the atomic action space, initial shortcuts, and initial tips of \texttt{Mobile-Agent-RAG} with those of the baseline model, \texttt{Mobile-Agent-E}. The shared action space is detailed in Table~\ref{action space}.
The shared initial tips are as follows:
\begin{enumerate}
    \item Do not add any payment information. If you are asked to sign in, ignore it or sign in as a guest if possible. Close any pop-up windows when opening an app.
    \item By default, no APPs are opened in the background.
    \item Screenshots may show partial text in text boxes from your previous input; this does not count as an error.
    \item When creating new Notes, you do not need to enter a title unless the user specifically requests it.
\end{enumerate}

\begin{table*}[t]
    \centering
    \begin{tabular}{ll}
        \toprule
        \textbf{Operation} & \textbf{Description} \\
        \midrule
        Open\_App(app\_name) & Opens the app ``app\_name'' from the Home screen. \\
        Tap(x, y) & Taps the current screen at position (x, y). \\
        Swipe(x1, y1, x2, y2) & Swipes from (x1, y1) to (x2, y2) for scrolling content or navigating apps. \\
        Type(text) & Types ``text'' into an active input box. \\
        Enter($\cdot$) & Presses the Enter key. \\
        Back($\cdot$) & Returns to the previous screen or state. \\
        Home($\cdot$) & Returns to the  home page. \\
        Wait($\cdot$) & Pauses execution for 10 seconds to allow for page loading. \\
        Tap\_Type\_Enter(x, y, text) & A composite action: taps an input box at (x, y), types the ``text'', then presses Enter.\\
        \bottomrule
    \end{tabular}
    \caption{A detailed list of the atomic actions and composite shortcuts available to the agent. This identical action space is adopted from the baseline model, \texttt{Mobile-Agent-E},  to ensure that performance gains are attributed to our retrieval-augmented components rather than a more expressive action set.}

    \label{action space}
\end{table*}

\section{C. Further Details on the Hierarchical Multi-agent Framework}
\label{appendix:multi-agent_implementation_details}
This section provides a more detailed look into the implementation of our proposed \texttt{Mobile-Agent-RAG}'s hierarchical multi-agent framework, which is inherited from \texttt{Mobile-Agent-E}. Notably, all core agents and support modules, \textit{with the exception of the Perceptor}, are categorized as reasoning agents.

\paragraph{Reasoning Agents: Manager, Operator, Action Reflector, and Notetaker} 
Our framework's four reasoning agents are powered by specific API versions of leading large language models. For our experiments, we use Gemini-1.5-pro-latest, GPT-4o, and Claude-3.5-Sonnet-latest as the underlying inference engines.

\paragraph{Perceptor} 
The Perceptor module is largely based on the implementation in \texttt{Mobile-Agent-E}. For fine-grained text information, we leverage the DBNet\footnote{\url{https://modelscope.cn/models/iic/cv_resnet18_ocr-detection-db-line-level_damo}} model from the ModelScope platform for Optical Character Recognition (OCR) text detection, and the ConvNextViT-document\footnote{\url{https://modelscope.cn/models/iic/cv_convnextTiny_ocr-recognition-document_damo}} model for character recognition. For fine-grained icon information, we employ GroundingDINO for icon grounding and use Qwen-VL-Plus to generate descriptive captions for each cropped icon.

\section{D. Retrieval Algorithms for Manager-RAG and Operator-RAG}

This section provides a detailed description of the retrieval algorithms central to our framework. The \texttt{Mobile-Agent-RAG} system leverages two distinct retrieval components: {Manager-RAG} and {Operator-RAG}. The former is designed to retrieve high-level strategic guidance to inform the agent's overall plan, while the latter focuses on retrieving app-specific operational knowledge to enable precise atomic actions. We outline the algorithms for each component below.

\paragraph{Manager-RAG}
\label{appendix:manager_rag_retrieval}

The Manager-RAG's retrieval algorithm is designed to provide high-level guidance for plan generation. Its knowledge base ($K_{MR}$), which is manually curated, contains $n_{MR}$ ($n_{MR}=25$ in our experiment) documents ($D_{MR}$). Each document consists of a mobile task instruction ($I_{MR}$) and a sequence of human-annotated operation steps ($H_{MR}$). When a new task instruction ($I_{\text{query}}$) is provided, the system retrieves the top-$k$ (We set $k=3$ in our experiment) most semantically similar documents from $K_{MR}$ using embeddings from Contriever-MSMARCO. The retrieval process is formally described in Algorithm~\ref{algorithm1}.

\begin{algorithm}[t]
\caption{Manager-Retrieve Algorithm}
\label{algorithm1}
\textbf{Input:} Task instruction $I_{\text{query}}$ \\
\textbf{Parameter:} Manager-RAG knowledge base $K_{MR} = \{(I_{MR}^{(i)}, H_{MR}^{(i)})\}_{i=1}^{n_{MR}}$; embedding function $f(\cdot)$; number $k$ \\
\textbf{Output:} Top-$k$ retrieved results $\mathcal{R}_M = \{(I_{MR}^{(j)}, H_{MR}^{(j)})\}_{j=1}^k$ \\
\begin{algorithmic}[1]
\STATE $v_{\text{query}} \leftarrow f(I_{\text{query}})$
\STATE Initialize $V_{MR}^{\text{sim}} \leftarrow [\,]$
\FORALL{$(I_{MR}^{(i)}, H_{MR}^{(i)}) \in K_{MR}$}
    \STATE $v_{MR}^{(i)} \leftarrow f(I_{MR}^{(i)})$
    \STATE $\text{sim}^{(i)}_{MR} \leftarrow \cos(v_{\text{query}}, v_{MR}^{(i)})$
    \STATE Append $(\text{sim}^{(i)}_{MR}, I_{MR}^{(i)}, H_{MR}^{(i)})$ to $V_{MR}^{\text{sim}}$
\ENDFOR
\STATE Sort $V_{MR}^{\text{sim}}$ by descending $\text{sim}^{(i)}_{MR}$
\STATE $\mathcal{R}_M \leftarrow$ Top-$k$ documents from $V_{MR}^{\text{sim}}$
\RETURN $\mathcal{R}_M$
\end{algorithmic}
\end{algorithm}

\paragraph{Operator-RAG}
\label{appendix:operator_rag_retrieval}

The Operator-RAG's retrieval algorithm is responsible for retrieving app-specific operational knowledge to support the generation of atomic actions. Its knowledge base ($K_{OR}^{\text{app}}$) is semi-automatically constructed and partitioned by application domain (e.g., YouTube, Maps) to prevent cross-app interference. Each knowledge base contains $n_{OR}^{\text{app}}$. This parameter's value is specific to each application. Each knowledge base document ($D_{OR}$), contains a subtask description ($T_{OR}^{\text{app}(i)}$), a reference screenshot ($S_{OR}^{(i)}$), and the corresponding atomic action and arguments ($A_{OR}^{(i)}$). Given a current subtask, the system retrieves the top-1 relevant document from the active app's knowledge base. The process is outlined in Algorithm~\ref{algorithm2}.

\begin{algorithm}[t]
\caption{Operator-Retrieve Algorithm}
\label{algorithm2}
\textbf{Input:} Current subtask and identified app name $T^{\text{app}}_{\text{query}}$ \\
\textbf{Parameter:} App-specific knowledge base $K_{OR}^{\text{app}} = {\{(T_{OR}^{\text{app}(i)}, S_{OR}^{(i)}, A_{OR}^{(i)})}\}_{i=1}^{n^{\text{app}}_{OR}}$; 
embedding function $f(\cdot)$ \\
\textbf{Output:} Top-1 retrieved result $\mathcal{R}_O=(T_{OR}^{app(j)}, S_{OR}^{(j)}, A_{OR}^{(j)})$ \\

\begin{algorithmic}[1]

\STATE $v_{\text{query}} \leftarrow f(T^{\text{app}}_{\text{query}})$
\STATE Initialize $V_{OR}^{\text{sim}} \leftarrow [\,]$

\FORALL{$(T_{OR}^{app(i)}, S_{OR}^{(i)}, A_{OR}^{(i)}) \in K_{OR}^{\text{app}}$}
\STATE $v_{OR}^{(i)} \leftarrow f(T_{OR}^{app(i)})$
\STATE $\text{sim}^{(i)}_{OR} \leftarrow \cos(v_{\text{query}}, v_{OR}^{(i)})$
\STATE Append $(\text{sim}^{(i)}_{OR}, T_{OR}^{app(i)}, S_{OR}^{(i)}, A_{OR}^{(i)})$ to $V_{OR}^{\text{sim}}$
\ENDFOR

\STATE Sort $V_{OR}^{\text{sim}}$ by descending $\text{sim}^{(i)}_{OR}$
\STATE $\mathcal{R}_O \leftarrow$ Top-1 document from $V_{OR}^{\text{sim}}$
\RETURN $\mathcal{R}_O$
\end{algorithmic}
\end{algorithm}

\section{E. Further Details for Retrieval-Oriented Knowledge Base Collection}
\label{appendix:knowledge_base_construction}
This section details the construction and collection process for the two distinct knowledge bases, which are critical for the retrieval-augmented components of our proposed framework. Both the Manager-RAG and Operator-RAG knowledge bases (KB) are built through a combination of manual and semi-automated strategies to ensure high quality and relevance. 

\paragraph{Manager-RAG Knowledge Base Collection}
The Manager-RAG knowledge base is meticulously compiled to provide high-level strategic guidance. Its construction involves a multi-stage process:
\begin{enumerate}
    \item \textbf{Task Execution by Experimenters:} 
    As illustrated in Table~\ref{tab:Mobile-Agent-RAG_bench_part1} and Table \ref{tab:Mobile-Agent-RAG_bench_part2}, we engage multiple human experimenters to perform, on real mobile devices, the top 50\% of Mobile-Eval-RAG tasks from each category. 
    This setup ensures that operations are effective and transferable to real-world conditions. 
    During each run, we log screens, timestamps, and atomic actions to produce raw, verifiable trajectories.

    \item \textbf{Rigorous Sequence Filtering:} 
    From the raw trajectories, we remove erroneous or incomplete trials, deduplicate near-identical runs, and apply a \emph{minimal-success} criterion: 
    for each task, we retain the shortest real-world action sequence that reliably completes the task. 
    This filtering keeps optimized pathways and eliminates suboptimal or redundant operations.

    \item \textbf{Data Structuring:} 
    We structure the curated data into a schema retrievable by the Manager-RAG module. Each knowledge-base document includes: (1) \textbf{Task Instruction}, representing the overall task objective in text; and (2) \textbf{Human Steps}, representing the corresponding concise operation steps that provide high-level guidance.

\end{enumerate}

We normalize terminology, encode the \textit{Task Instruction} field into embeddings, and store entries in a vector database to enable efficient similarity search during inference.

Table~\ref{tab:restaurant_tasks} shows several examples of the structured data entries for the Manager-RAG knowledge base.

\paragraph{Operator-RAG Knowledge Base Collection}

The Operator-RAG knowledge base is carefully constructed to provide accurate, executable operational guidance. Its construction involves a multi-stage process:

\begin{enumerate}
    \item \textbf{Data Collection:}  As shown in Table~\ref{tab:Mobile-Agent-RAG_bench_part1} and Table~\ref{tab:Mobile-Agent-RAG_bench_part2}, we collect Operator-RAG data by recording precise agent actions while executing Mobile-Eval-RAG tasks. \textbf{Strategic Minimization of Human Intervention:} To reduce manual overhead and better reflect autonomous behavior, we strategically minimize human involvement. Human guidance is only provided when the agent fails to perform correctly, ensuring correctness while maintaining a high degree of autonomy in the recorded behavior. \textbf{Instance Recording:} Each recorded instance includes (1) the current subtask described in text, (2) a screenshot of the corresponding UI state, and (3) the atomic action required to achieve the subtask. A representative subset of correctly executed operational instances for the Maps App is presented in Table~\ref{operator-rag}.
    
    \item \textbf{Rigorous Data Cleansing:}  
    Following initial data collection, we perform strict cleansing to eliminate erroneous and redundant entries. This step ensures that the knowledge base maintains only high-quality, accurate, and reliable operational examples.

    \item \textbf{Data Structuring:}  
    We structure the cleaned data into a schema retrievable by the Operator-RAG module. Each knowledge-base document comprises three key components: (1) \textbf{Subtask}, a textual description of the specific subtask objective; (2) \textbf{Screenshot}, the visual UI state corresponding to the subtask, stored as a local file path; and (3) \textbf{Action}, the atomic action along with its arguments required to accomplish the subtask.

\end{enumerate}

We encode the \textit{Subtask} field into embeddings and store the structured entries in a vector database, enabling efficient similarity-based retrieval during inference.

\begin{table*}[t]
\setlength{\tabcolsep}{1mm} 
\centering
\begin{tabular}{p{0.35\textwidth}|p{0.6\textwidth}}
\toprule
\textbf{Task Instruction} & \textbf{Human steps} \\
\midrule
Find the best ramen place in Chicago Loop with at least 500 reviews and rating over 4.5. Write a review summary in Notes. & open Maps app, tap on the search bar, type ``the best ramen place in Chicago Loop'', enter, swipe down to find more results, tap on the ``RAMEN-SAN Whisky Bar'', swipe down to find more reviews, tap home, tap Notes, tap ``+'', tap ``text'', type the review summary \\
\midrule
Look for a family-friendly restaurant in Urbana suitable for kids. Write a short summary in Notes. & open Maps app, tap on the search bar, type ``family-friendly restaurant in Urbana'', enter, tap the filter, tap ``good for kids'', tap apply, tap on the first result, swipe down to find more information, swipe down to find more information, swipe down to find more information, tap home, tap Notes, tap ``+'', tap ``text'', type the short summary \\
\midrule
Search for breakfast buffet places near me with good reviews. Compare 2 and write in Notes. & open Maps app, tap on the search bar, type ``breakfast buffet places'', enter, tap filter, tap ``Distance'', tap ``Apply'', tap on the first result has review, swipe down to find more information, back, tap on the second result has review, swipe down to find more information, tap home, tap Notes, tap ``+'', tap ``text'', type the summary to compare the two restaurant \\
\midrule
Find a hotpot restaurant near a university campus. Write the address and the average user rating into Notes. & open Maps app, tap on the search bar, type ``hotpot restaurant near a university campus'', enter, tap the first result, swipe down to find more information, tap home, tap Notes, tap ``+'', tap ``text'', type the address and the average user rating \\
\midrule
Find a Chinese restaurant in Chicago with rating over 4.5 that offers takeout. Save 3 dishes and their prices in Notes. & open Maps app, tap on the search bar, type ``Chinese restaurant in Chicago'', enter, tap filter, tap ``4.5 star'', tap ``Takeaway'', tap ``Apply'', tap the first result, tap menu, swipe down to find more information, swipe down to find more information, swipe down to find more information, tap home, tap Notes, tap ``+'', tap ``text'', type the 3 dishes and their prices \\
\bottomrule
\end{tabular}
\caption{
Representative documents with \textbf{task instructions} and \textbf{human steps} from \textit{Restaurant Recommendation} tasks used to construct the \textbf{Manager-RAG} knowledge base.}

\label{tab:restaurant_tasks}
\end{table*}

\begin{table*}[t]
\small
\setlength{\tabcolsep}{2mm}
\centering
\begin{tabular}{lcccccc}
\toprule
\textbf{Benchmark} & \textbf{Multi-App / All} & \textbf{Apps} & \textbf{Avg Steps} & \textbf{CR Crit.} & \textbf{LongH.} & \textbf{RAG Eval.} \\
\midrule
Mobile-Eval           & 3 / 33         & 10 & 5.55  & -           & -           & - \\
DroidTask             & 0 / \textbf{158} & 13 & 5.56  & -           & -           & - \\
Mobile-Eval-v2        & 4 / 44         & 10 & 5.57  & -           & -           & - \\
AppAgent (General)    & 0 / 45         & 9  & 6.31  & -           & -           & \checkmark \\
AndroidWorld          & 10 / 116       & \textbf{20} & 9.13 & -    & -           & - \\
Mobile-Eval-E         & 19 / 25        & 15 & 14.56 & \checkmark  & \checkmark  & - \\
AppAgent (Long)       & 0 / 5          & 5  & 15.40 & -           & \checkmark  & \checkmark \\
\midrule[1pt]
\textbf{Mobile-Eval-RAG (simple)}  & 20 / 20        & 4  & 14.05 & \checkmark  & \checkmark  & \checkmark \\
\textbf{Mobile-Eval-RAG (complex)} & \textbf{30} / 30 & 7  & \textbf{18.80} & \checkmark  & \checkmark  & \checkmark \\
\bottomrule
\end{tabular}
\caption{Comparison of \textbf{Mobile-Eval-RAG} (ours) with existing mobile automation benchmarks. \textbf{Mobile-Eval-RAG (simple)} and \textbf{Mobile-Eval-RAG (complex)} represent the simple and complex task subsets of our proposed benchmark, respectively. \textbf{Multi-App / All} denotes the number of cross-app tasks versus the total number of tasks. \textbf{Apps} is the number of unique applications involved. \textbf{Avg Steps} is the average number of steps per task. \textbf{CR Crit.} indicates whether Completion Rate Evaluation Criteria (see \underline{\textbf{Appendix I}}) are defined. \textbf{LongH.} denotes whether the benchmark supports long-horizon task execution. \textbf{RAG Eval.} shows if the benchmark is suitable for RAG-based evaluation.}
\label{tab:benchmark_comparison}
\end{table*}

\begin{table*}[t]
\centering
\setlength{\tabcolsep}{1mm} 
\small
\begin{tabular}{@{}p{0.23\textwidth}@{\hspace{0.01\textwidth}}p{0.23\textwidth}@{\hspace{0.01\textwidth}}p{0.23\textwidth}@{\hspace{0.01\textwidth}}p{0.23\textwidth}@{}}
\toprule

\begin{minipage}[t]{\linewidth}
    \centering
    \includegraphics[width=0.23\linewidth]{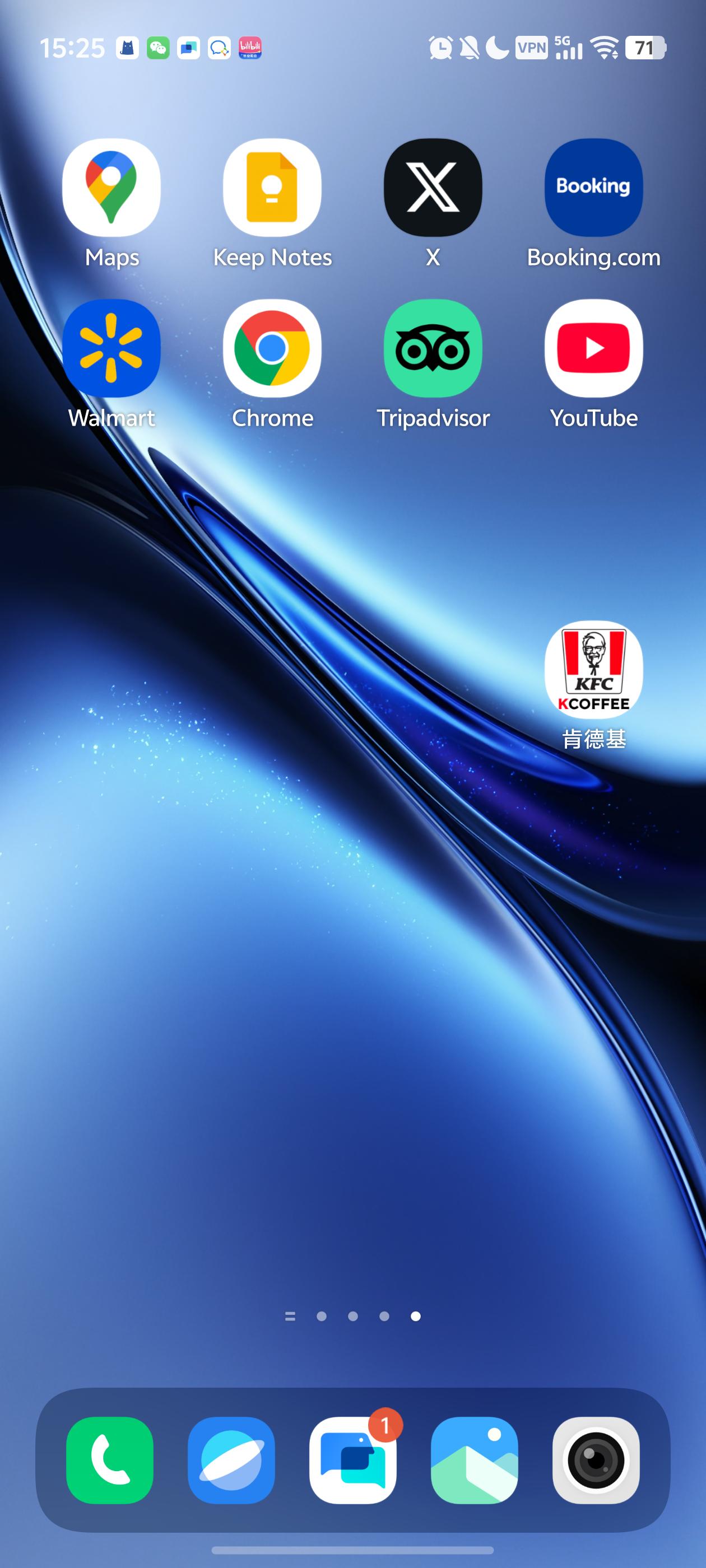}
    
    \textbf{Subtask:} Tap Maps app. \\ \textbf{Action:} Open\_App at \{``app\_name'': ``Maps''\}
\end{minipage} &
\begin{minipage}[t]{\linewidth}
    \centering
    \includegraphics[width=0.23\linewidth]{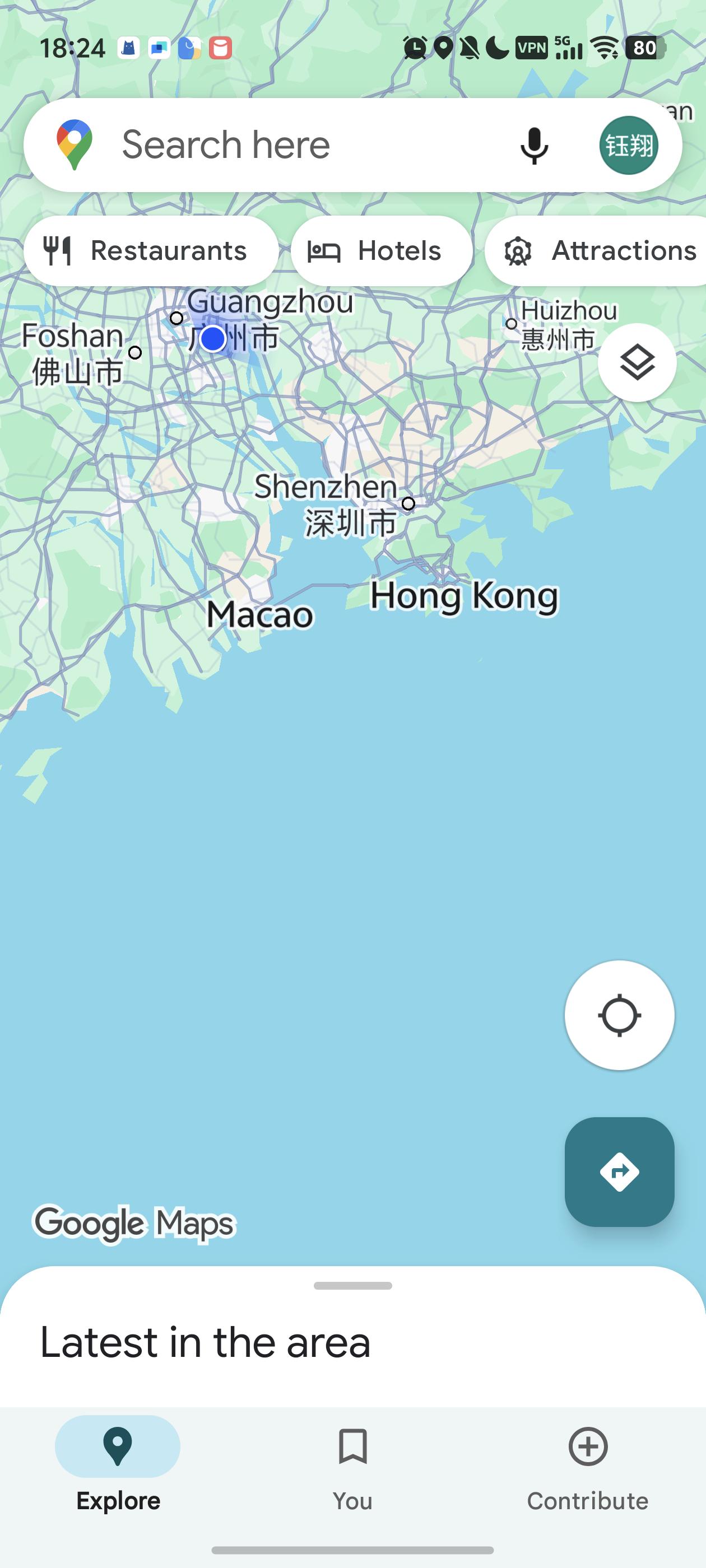}
    
    \textbf{Subtask:} Tap the search bar. \\ \textbf{Action:} Tap at \{``x'': 404, ``y'': 260\}
\end{minipage} &
\begin{minipage}[t]{\linewidth}
    \centering
    \includegraphics[width=0.23\linewidth]{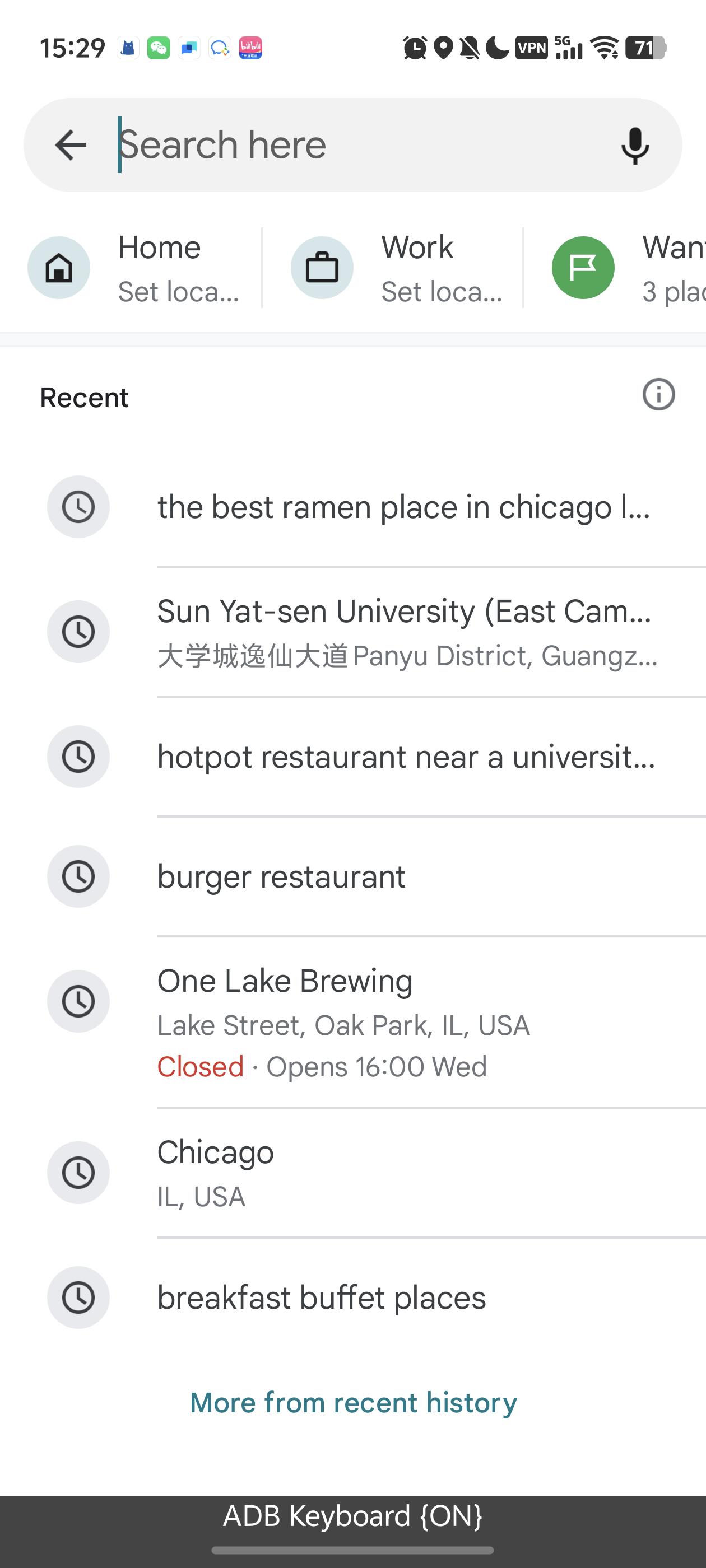}
    
    \textbf{Subtask:} Type ``ramen in Chicago Loop''. \\ \textbf{Action:} Type at \{``text'': ``ramen in Chicago Loop''\}
\end{minipage} &
\begin{minipage}[t]{\linewidth}
    \centering
    \includegraphics[width=0.23\linewidth]{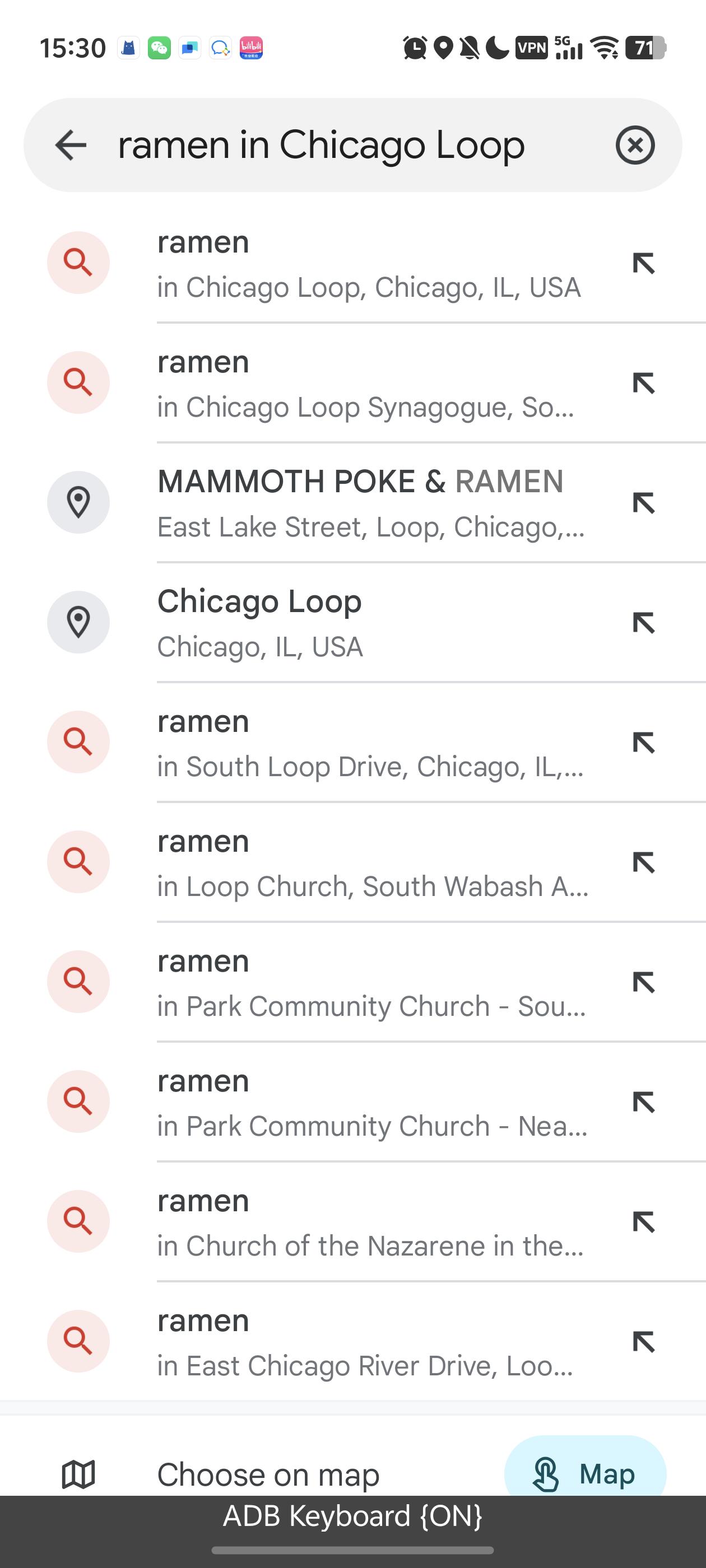}
    
    \textbf{Subtask:} Tap Enter. \\ \textbf{Action:} Enter at null
\end{minipage} \\
\midrule

\begin{minipage}[t]{\linewidth}
    \centering
    \includegraphics[width=0.23\linewidth]{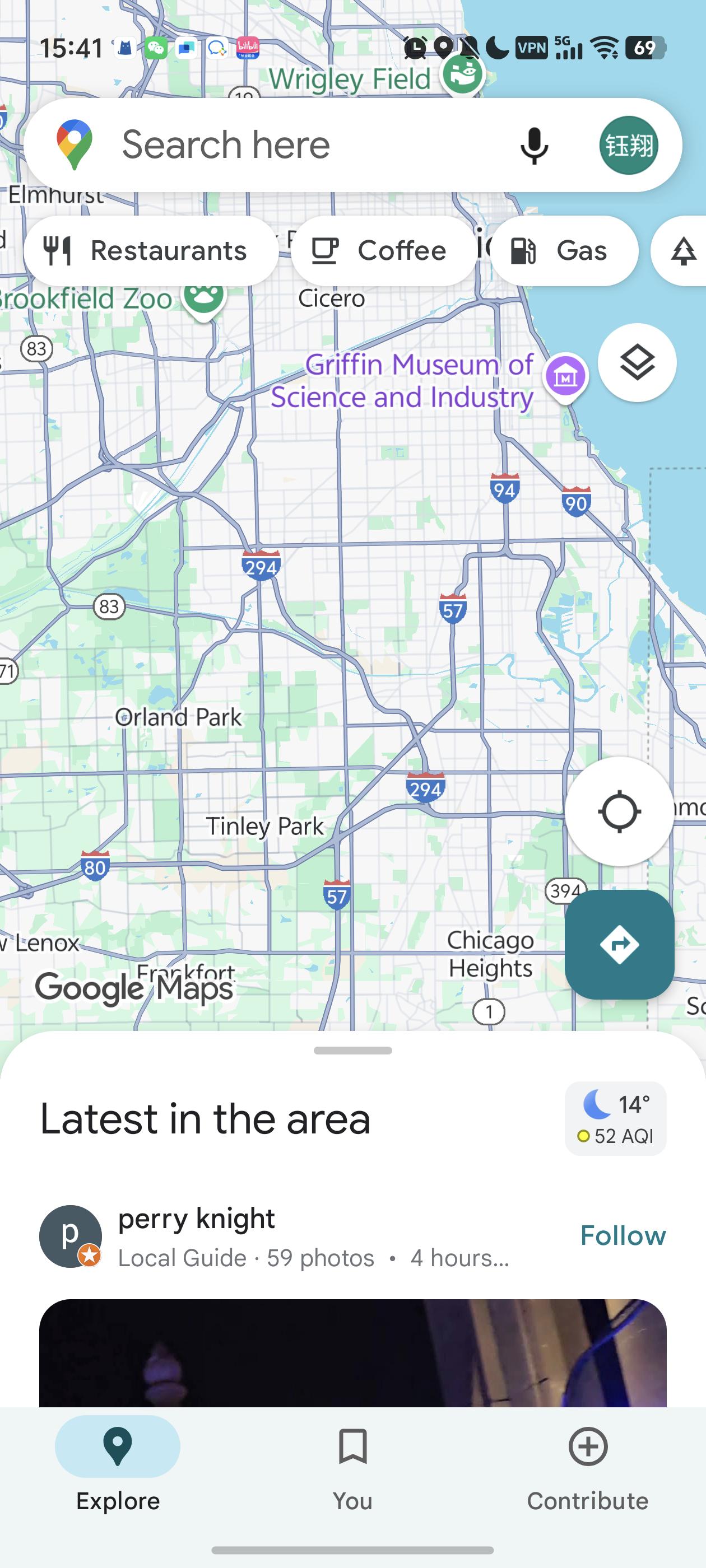}
    
    \textbf{Subtask:} Tap the search bar. Type ``ramen in Chicago Loop''. Tap Enter. \\ \textbf{Action:} Tap\_Type\_and\_Enter at \{``x'': 200, ``y'': 250, ``text'': ``ramen in Chicago Loop''\}
\end{minipage} &
\begin{minipage}[t]{\linewidth}
    \centering
    \includegraphics[width=0.23\linewidth]{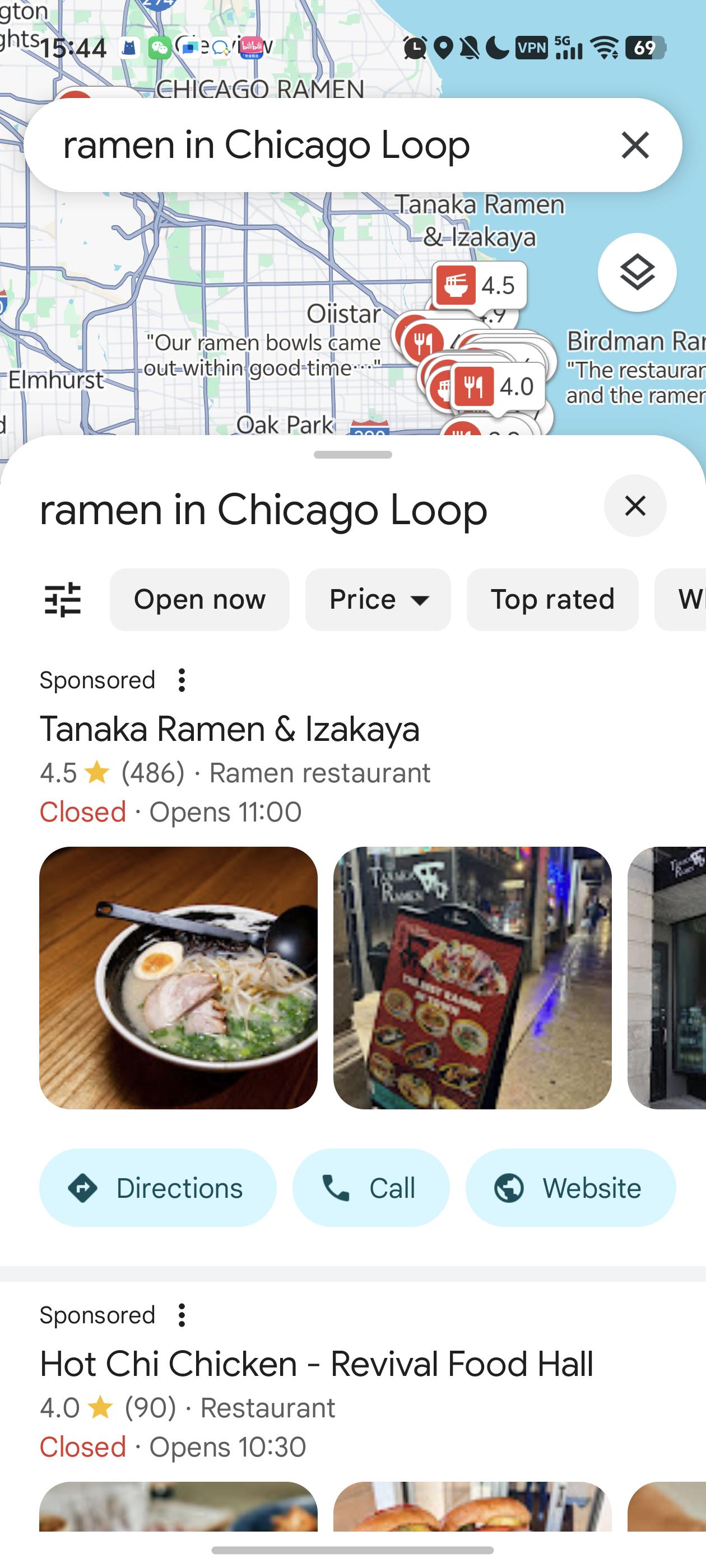}
    
    \textbf{Subtask:} Tap ``Filter''. \\ \textbf{Action:} Tap at \{``x'': 110, ``y'': 1068\}
\end{minipage} &
\begin{minipage}[t]{\linewidth}
    \centering
    \includegraphics[width=0.23\linewidth]{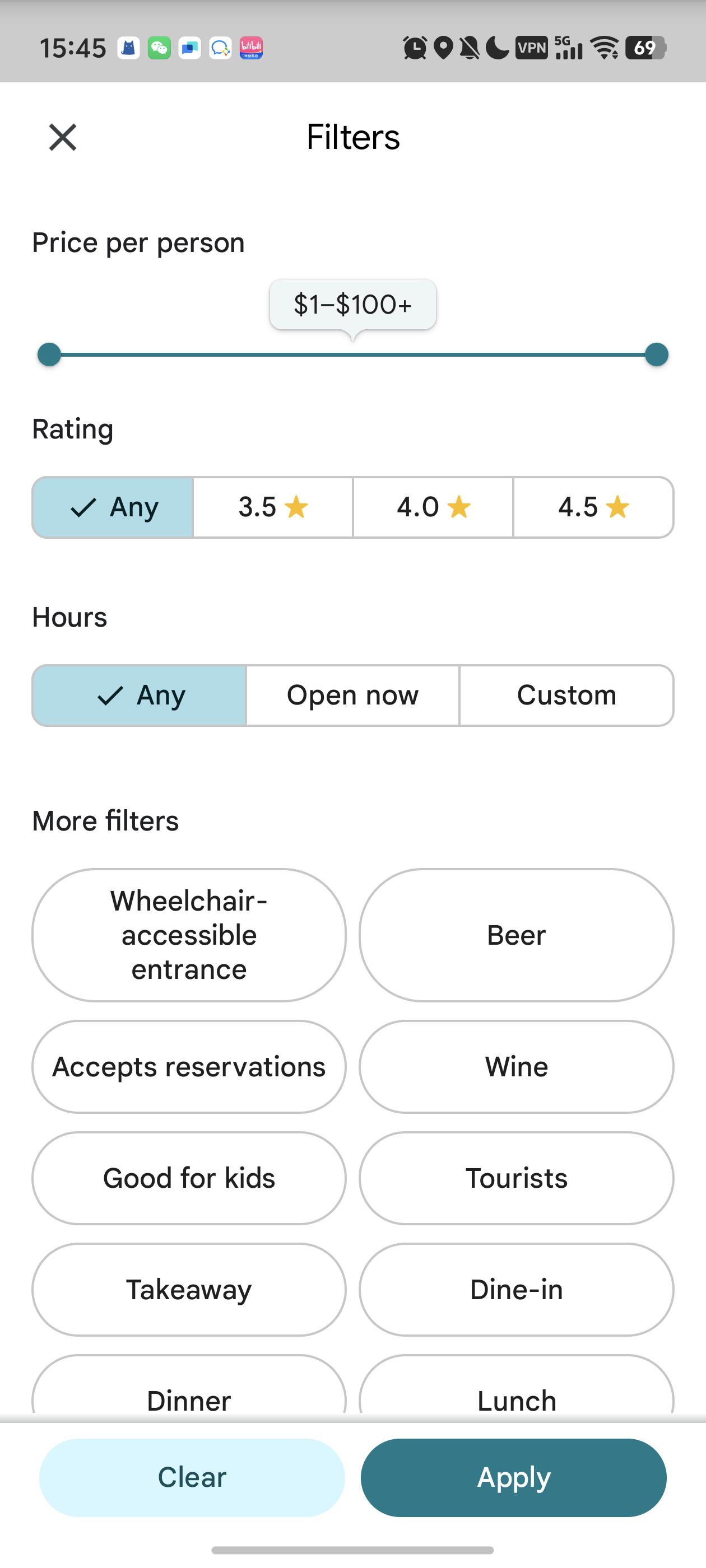}
    
    \textbf{Subtask:} Tap ``4.5 Stars \& up''. \\ \textbf{Action:} Tap at \{``x'': 1034, ``y'': 902\}
\end{minipage} &
\begin{minipage}[t]{\linewidth}
    \centering
    \includegraphics[width=0.23\linewidth]{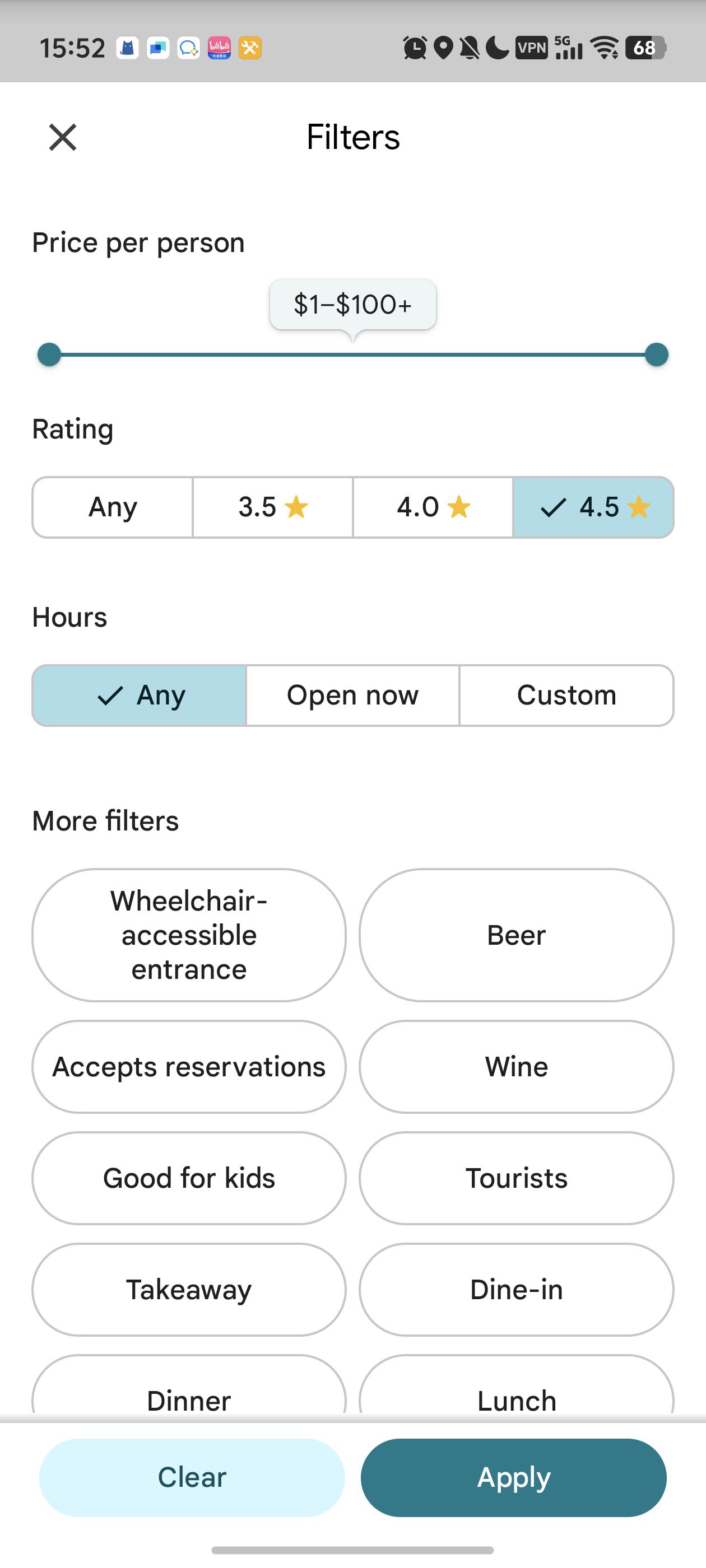}
    
    \textbf{Subtask:} Tap ``Apply. \\ \textbf{Action:} Tap at \{``x'': 917, ``y'': 2642\}
\end{minipage} \\
\midrule

\begin{minipage}[t]{\linewidth}
    \centering
    \includegraphics[width=0.23\linewidth]{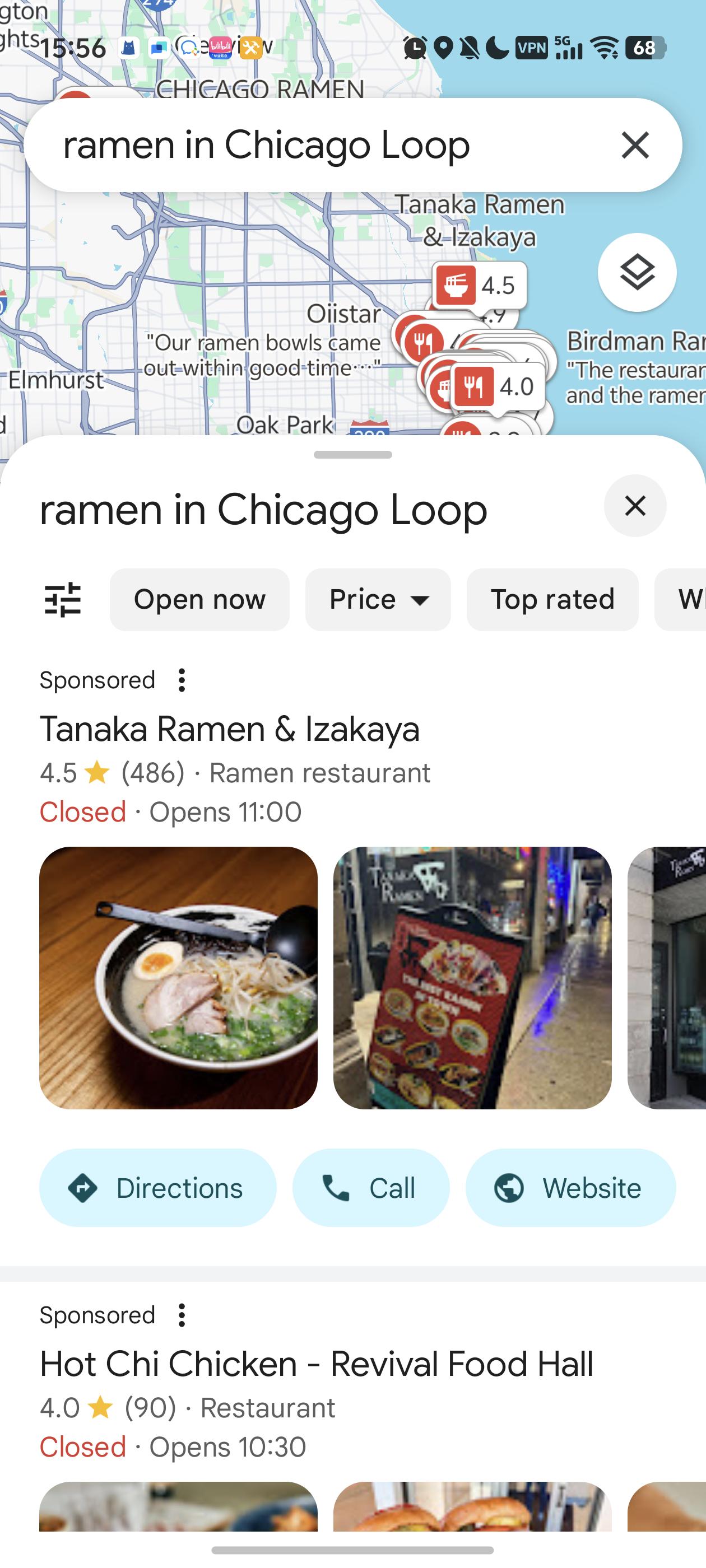}
    
    \textbf{Subtask:} Swipe up to see more results if needed. \\ \textbf{Action:} Swipe at \{``x1'': 630, ``y1'': 1400, ``x2'': 630, ``y2'': 280\}
\end{minipage} &
\begin{minipage}[t]{\linewidth}
    \centering
    \includegraphics[width=0.23\linewidth]{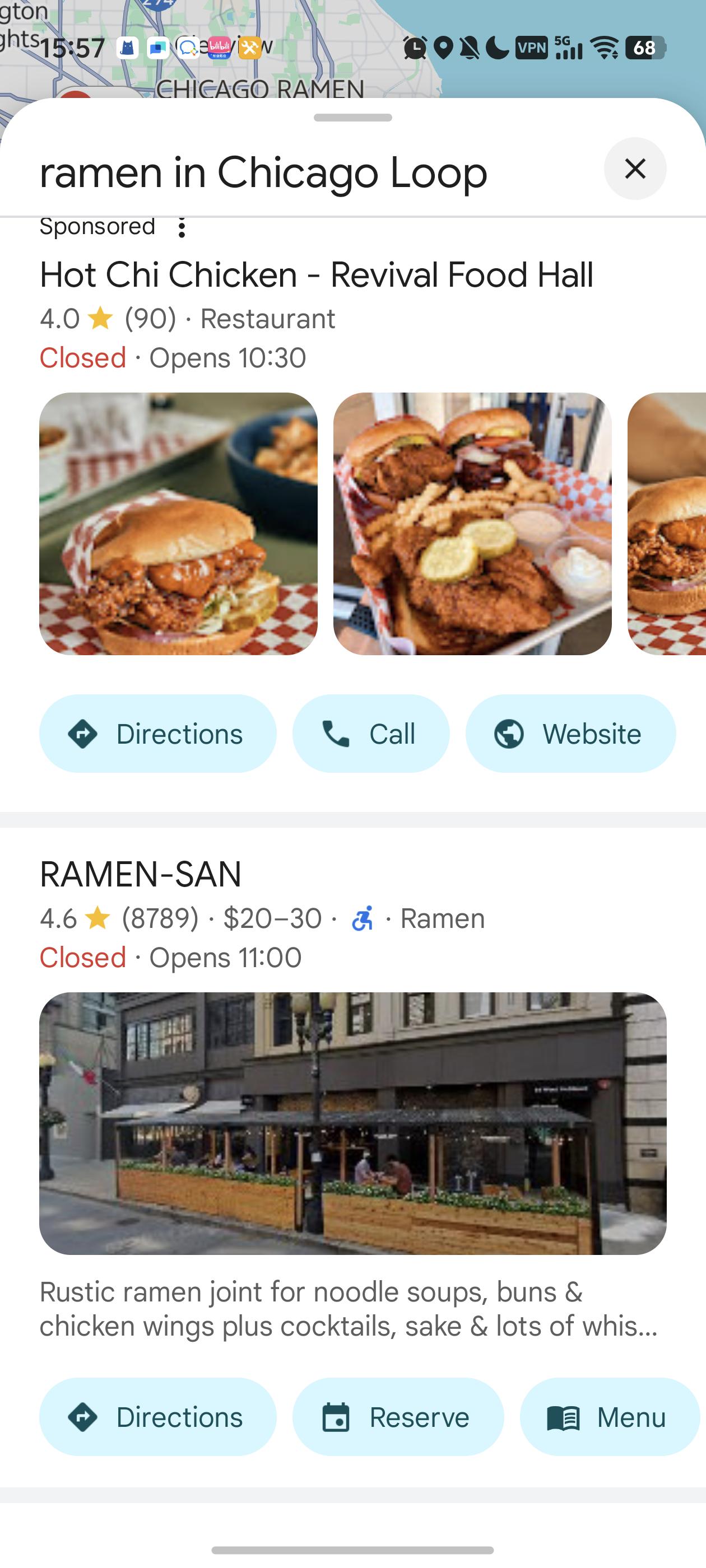}
    
    \textbf{Subtask:} Tap on a ramen place with at least 500 reviews and rating over 4.5. \\ \textbf{Action:} Tap at \{``x'': 250, ``y'': 1600\}
\end{minipage} &
\begin{minipage}[t]{\linewidth}
    \centering
    \includegraphics[width=0.23\linewidth]{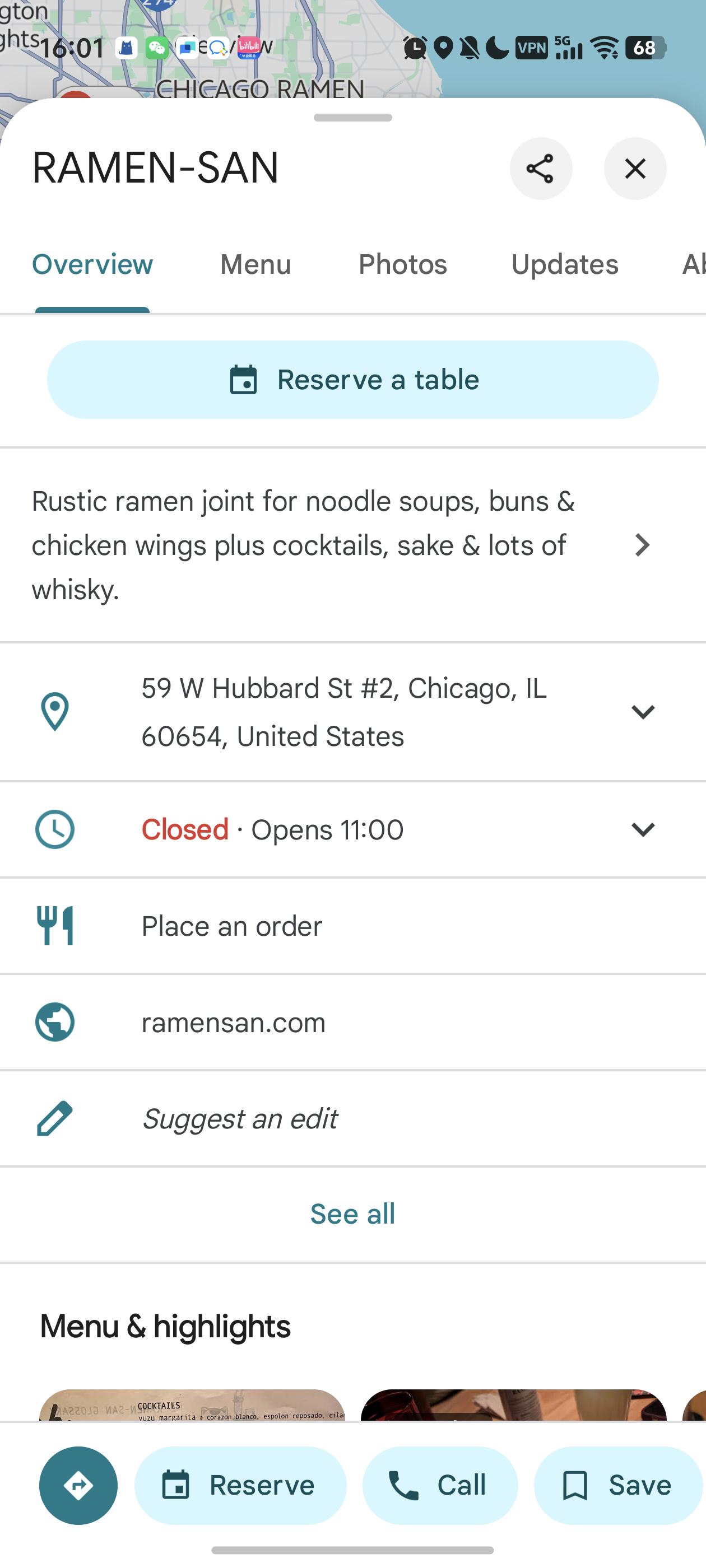}
    
    \textbf{Subtask:} Tap Home. \\ \textbf{Action:} Home at null
\end{minipage} &
\begin{minipage}[t]{\linewidth}
    \centering
    \includegraphics[width=0.23\linewidth]{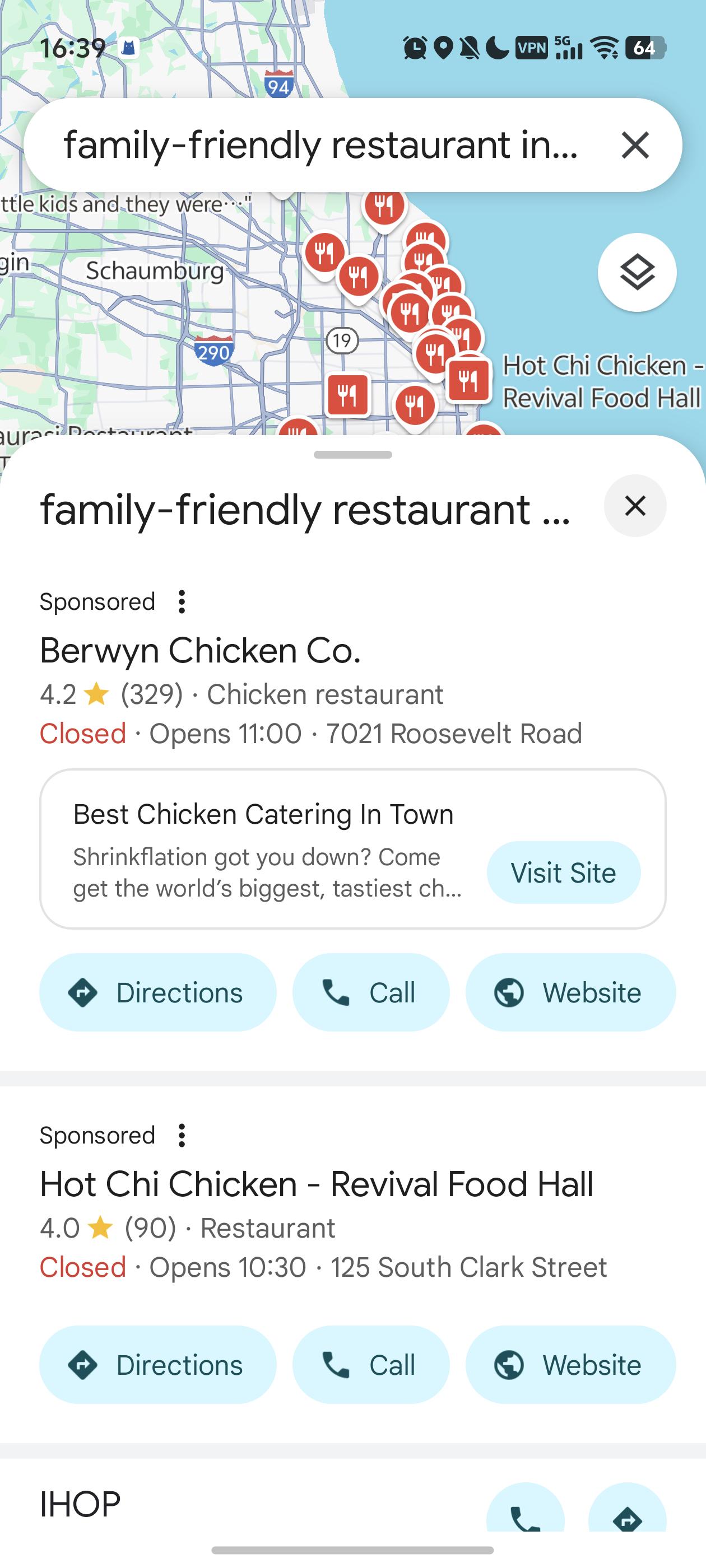}
    
    \textbf{Subtask:} Tap a restaurant result. \\ \textbf{Action:} Tap at \{``x'': 355, ``y'': 1162\}
\end{minipage} \\
\midrule

\begin{minipage}[t]{\linewidth}
    \centering
    \includegraphics[width=0.23\linewidth]{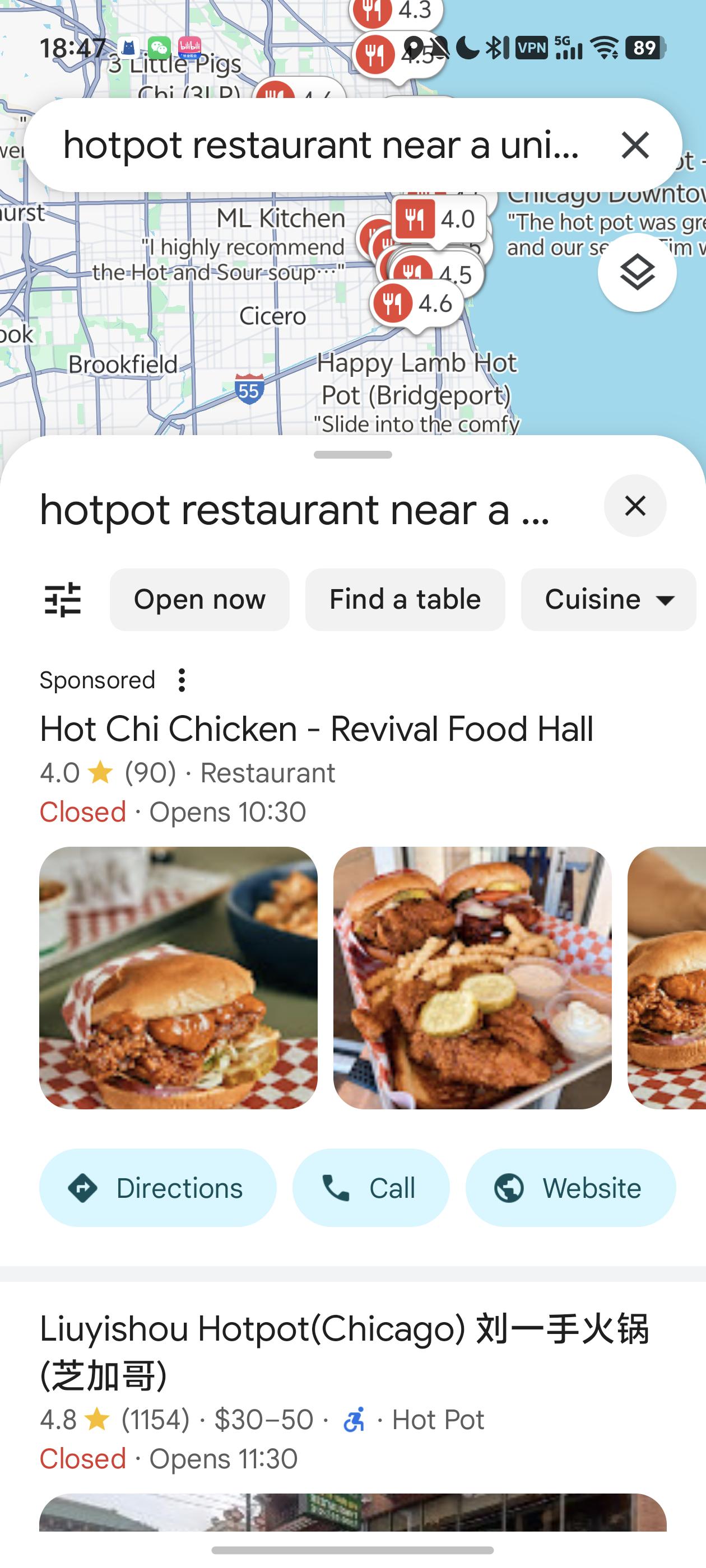}
    
    \textbf{Subtask:} Tap Liuyishou Hotpot(Chicago). \\ \textbf{Action:} Tap at \{``x'': 609, ``y'': 2420\}
\end{minipage} &
\begin{minipage}[t]{\linewidth}
    \centering
    \includegraphics[width=0.23\linewidth]{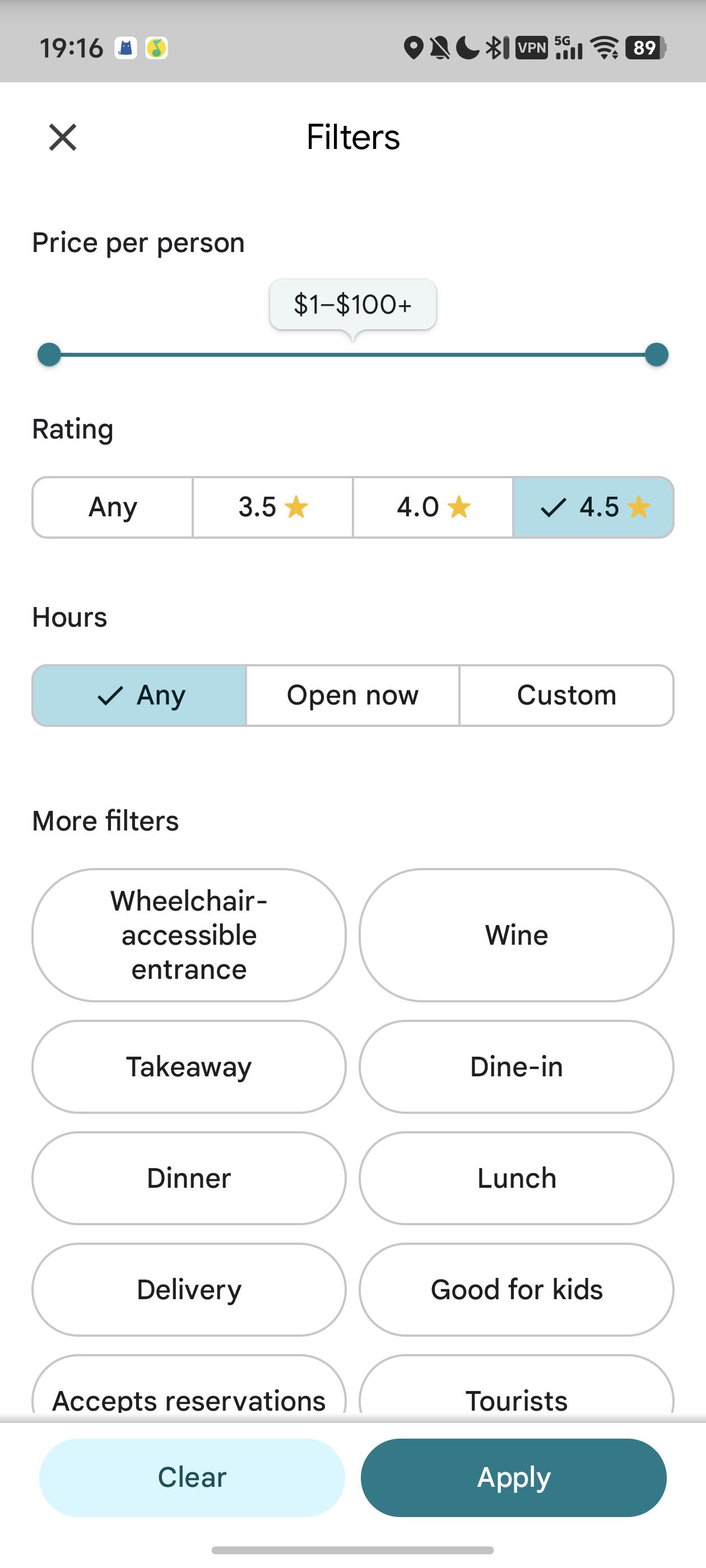}
    
    \textbf{Subtask:} Tap ``Takeaway''. \\ \textbf{Action:} Tap at \{``x'': 335, ``y'': 1905\}
\end{minipage} &
\begin{minipage}[t]{\linewidth}
    \centering
    \includegraphics[width=0.23\linewidth]{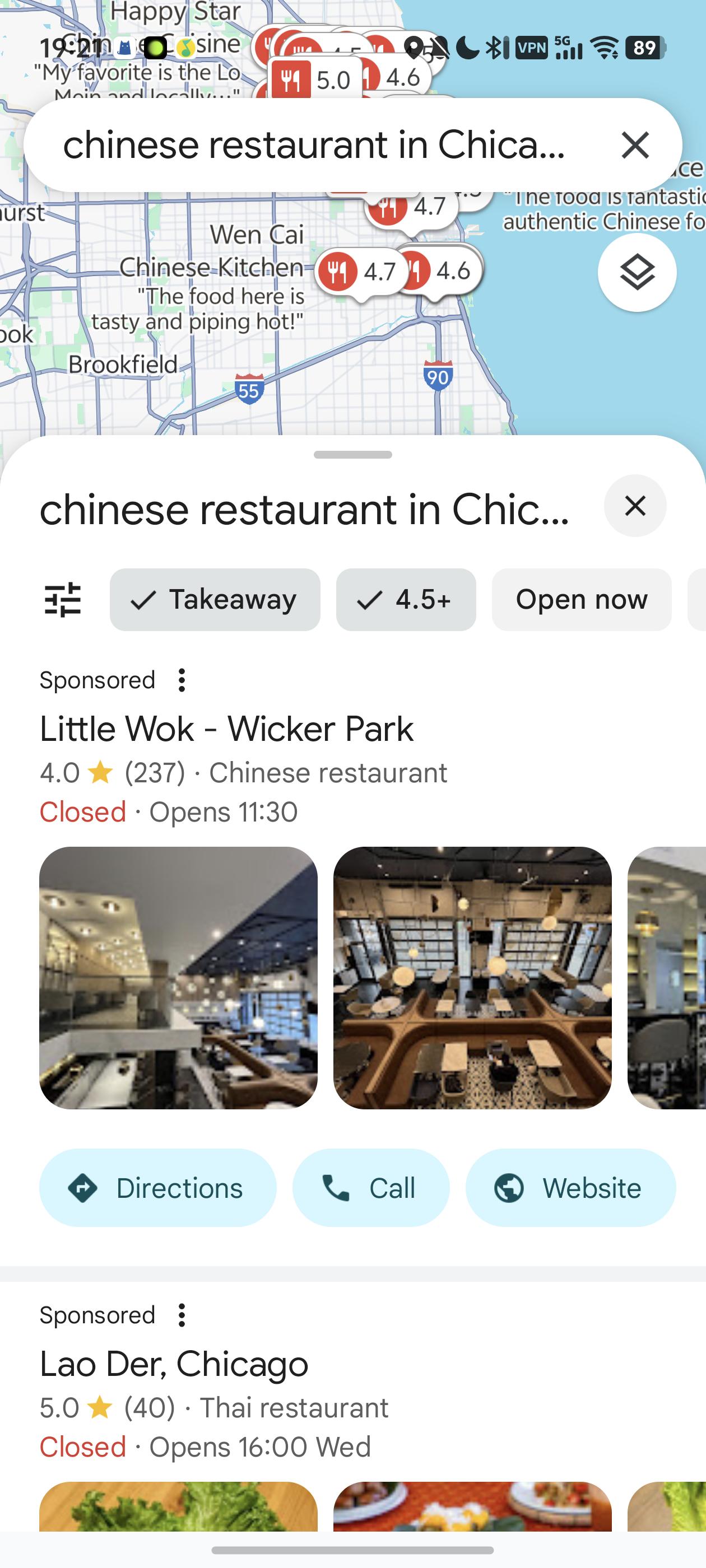}
    
    \textbf{Subtask:} Tap the first restaurant in the results. \\ \textbf{Action:} Tap at \{``x'': 300, ``y'': 1300\}
\end{minipage} &
\begin{minipage}[t]{\linewidth}
    \centering
    \includegraphics[width=0.23\linewidth]{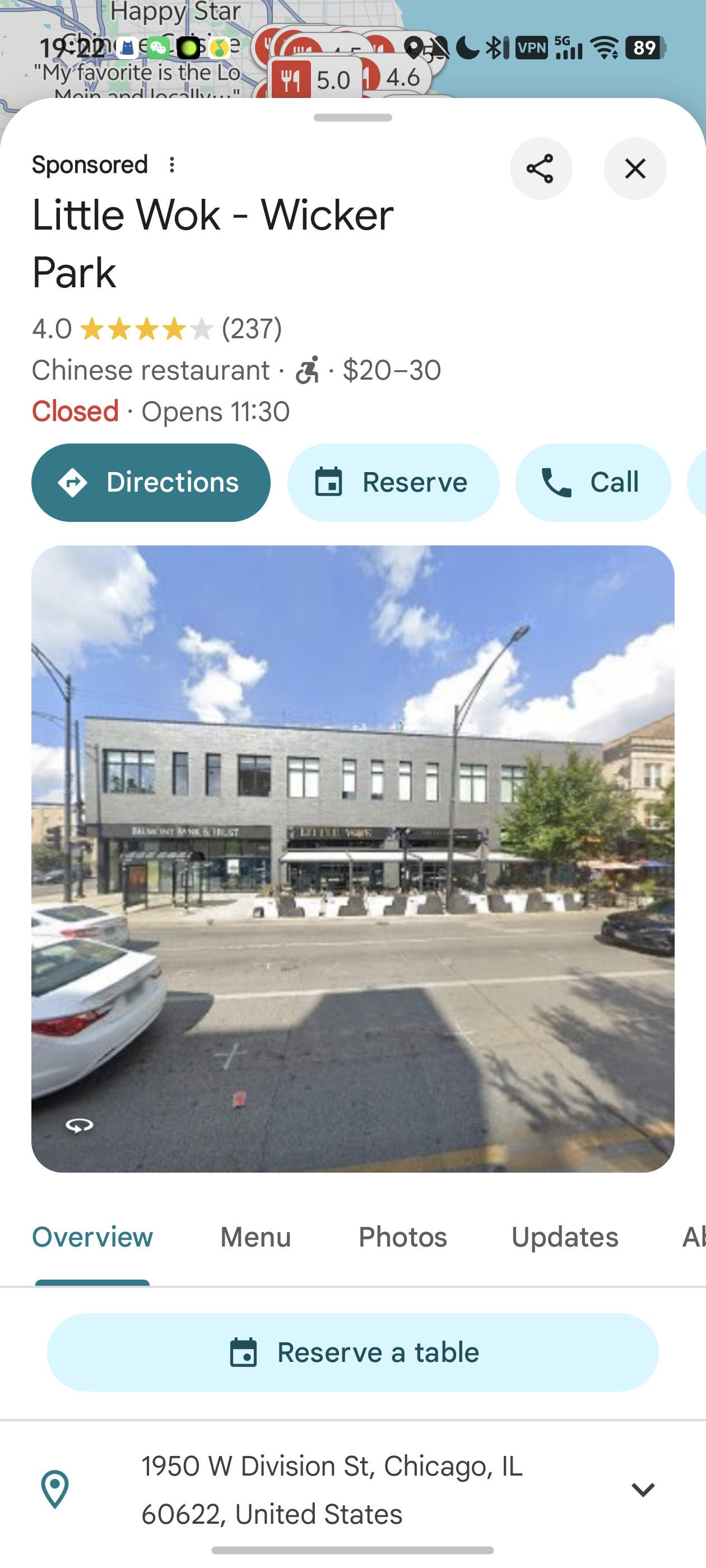}
    
    \textbf{Subtask:} Tap ``Photos''. \\ \textbf{Action:} Tap at \{``x'': 717, ``y'': 2207\}
\end{minipage} \\
\midrule

\begin{minipage}[t]{\linewidth}
    \centering
    \includegraphics[width=0.23\linewidth]{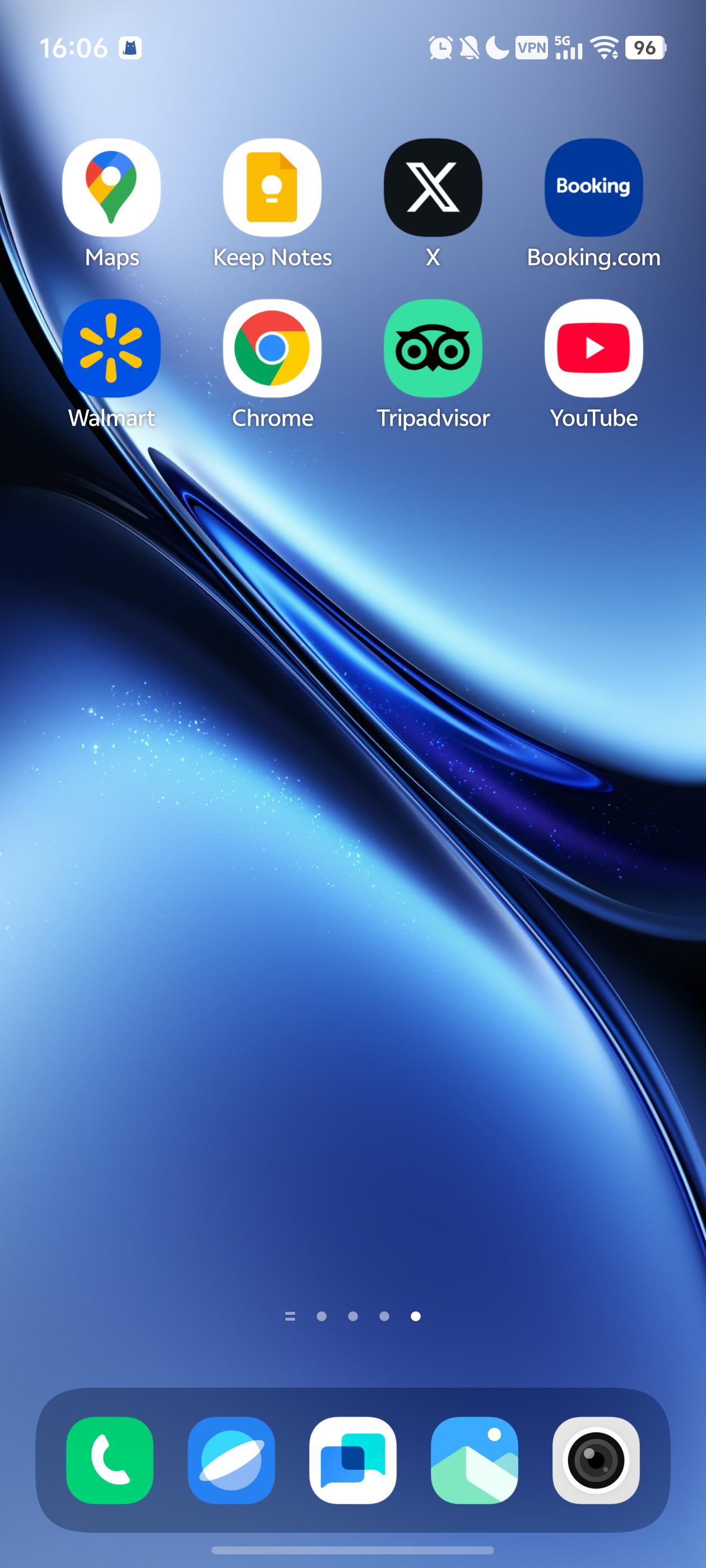}
    
    \textbf{Subtask:} Tap Maps app. \\ \textbf{Action:} Tap at \{``x'': 197, ``y'': 333\}
\end{minipage} &
\begin{minipage}[t]{\linewidth}
    \centering
    \includegraphics[width=0.23\linewidth]{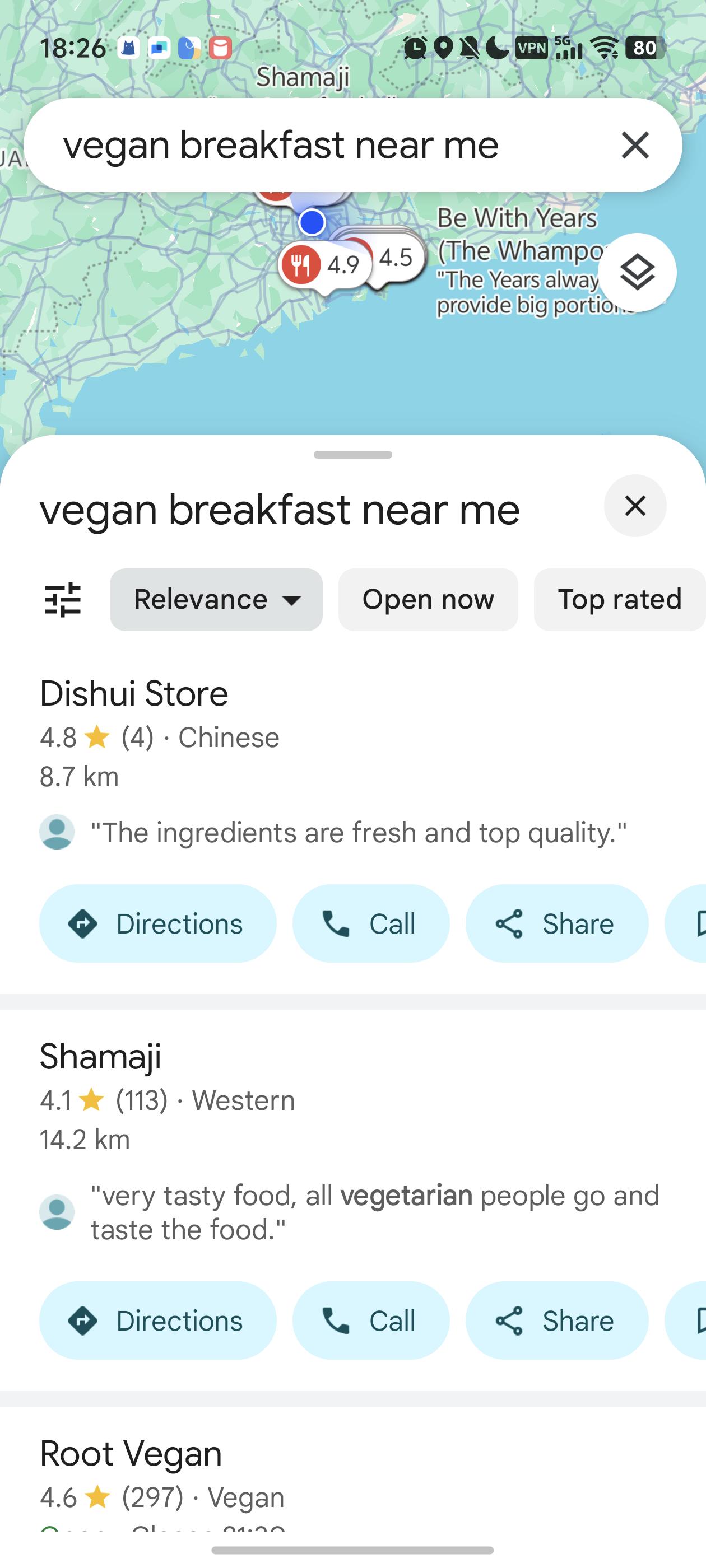}
    
    \textbf{Subtask:} Tap ``Top rated''. \\ \textbf{Action:} Tap at \{``x'': 1102, ``y'': 1069\}
\end{minipage} &
\begin{minipage}[t]{\linewidth}
    \centering
    \includegraphics[width=0.23\linewidth]{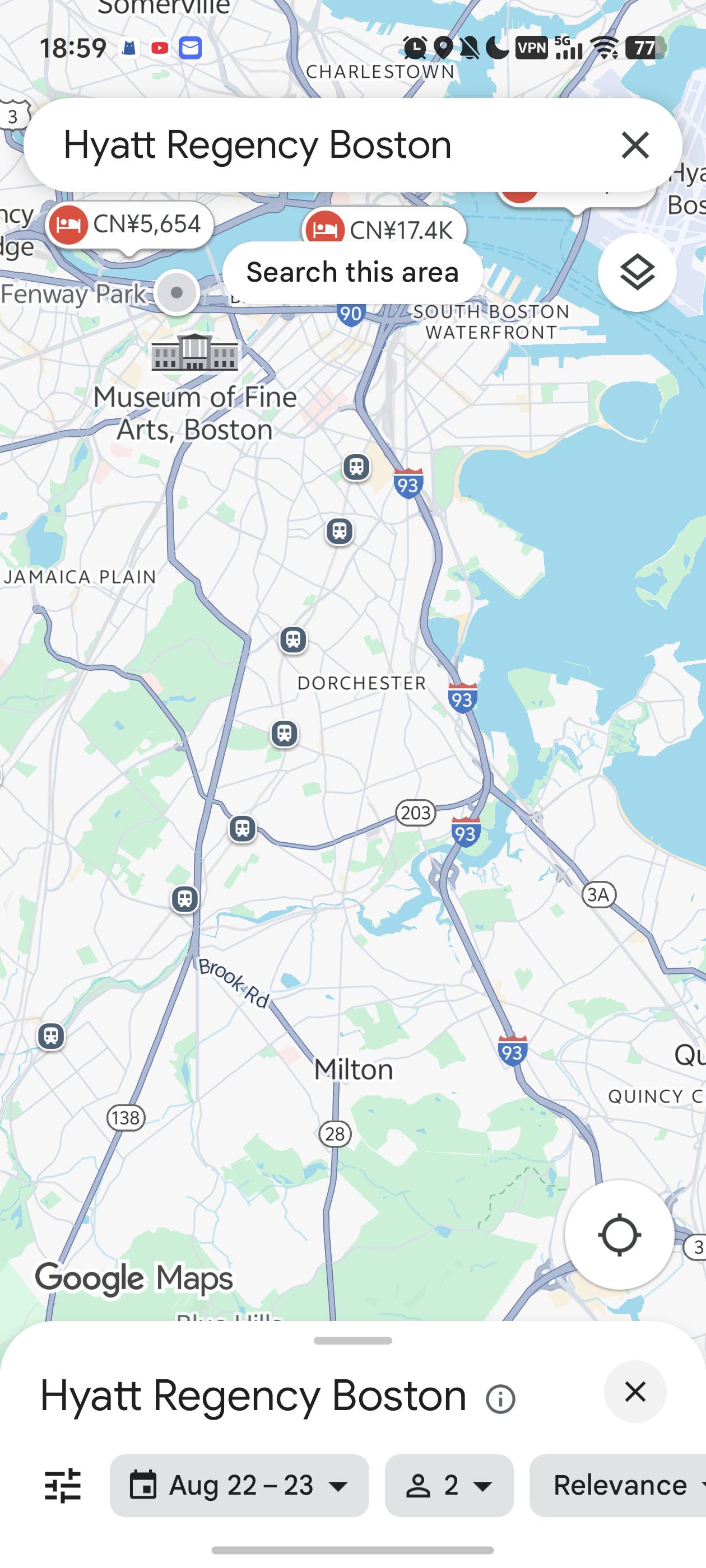}
    
    \textbf{Subtask:} Tap ``Attractions''. \\ \textbf{Action:} Tap at \{``x'': 777, ``y'': 153\}
\end{minipage} &
\begin{minipage}[t]{\linewidth}
    \centering
    \includegraphics[width=0.23\linewidth]{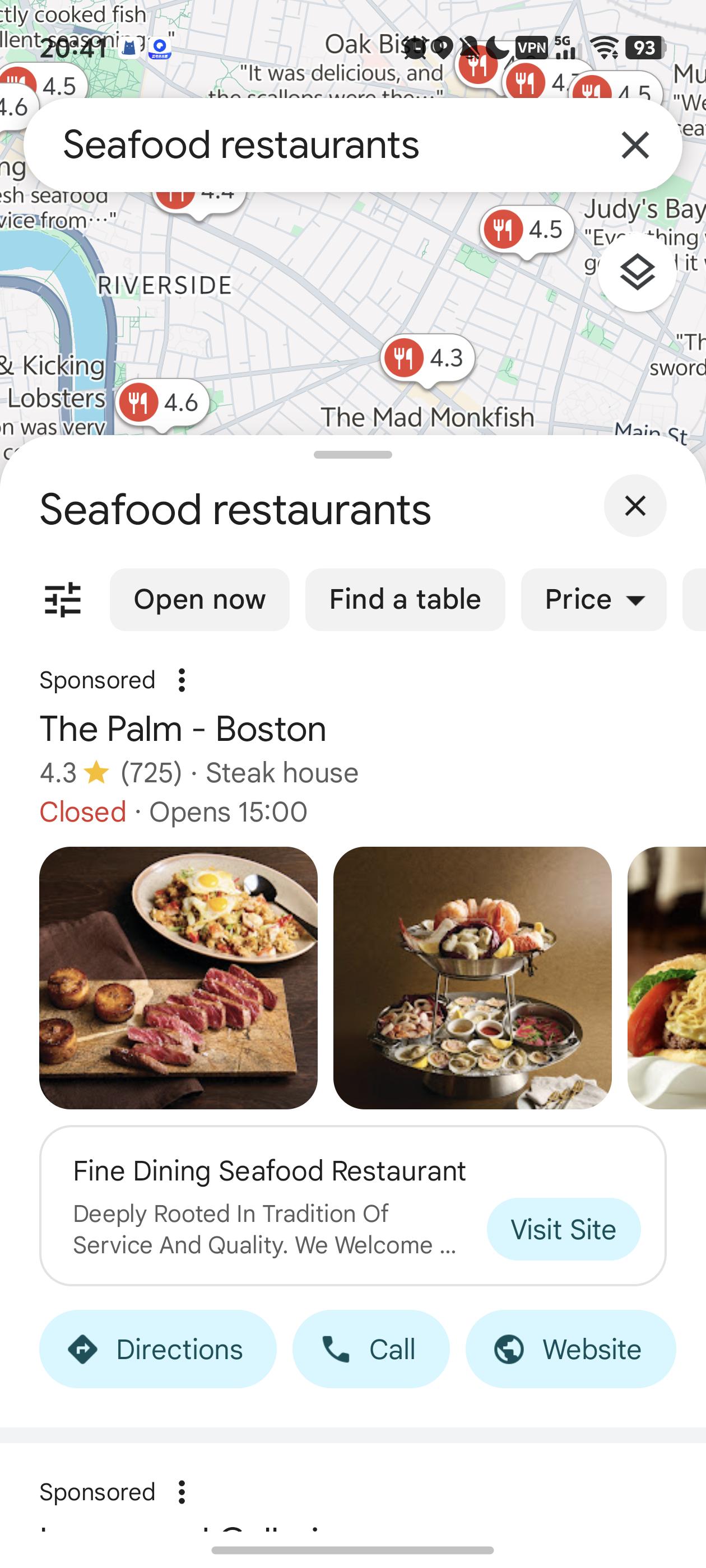}
    
    \textbf{Subtask:} Tap on ``Live \& Kicking Lobsters'' from the search results. \\ \textbf{Action:} Tap at \{``x'': 100, ``y'': 650\}
\end{minipage} \\
\bottomrule
\end{tabular}
\caption{Representative documents with \textbf{subtasks}, \textbf{screenshots}, and \textbf{actions} collected from executing Maps app task and used to construct the ``Maps'' part of the \textbf{Operator-RAG} knowledge base.}
\label{operator-rag}
\end{table*}

\section{F. Further Details for Mobile-Eval-RAG Construction}

This section provides a detailed overview of the proposed benchmark dataset, namely \textbf{Mobile-Eval-RAG}, which is specifically designed to assess multi-app collaboration in long-horizon mobile agents. We outline the key features of the dataset and highlight its unique design philosophy, {which makes it particularly suitable for evaluating the generalization capabilities of the proposed RAG systems.} Additionally, we provide a detailed comparison with the popular benchmark, e.g., \textbf{Mobile-Eval-E}, to clarify our evaluation focus.

\paragraph{More Key Features of Mobile-Eval-RAG} Mobile-Eval-RAG is a comprehensive benchmark dataset designed to evaluate the capability of multi-app collaboration across mobile agents. It simulates real-life scenarios that require agents to use multiple applications simultaneously to complete complex, daily tasks. The core idea is to test a system's practical operational ability in a cross-app mobile device environment, moving beyond single-app proficiency.
Additionally, a significant feature of Mobile-Eval-RAG is its emphasis on information integration and cross-platform collaboration. Most tasks require collecting and synthesizing data from multiple sources before performing analysis, comparison, and summarization. For example, in a restaurant recommendation task, the agent must search for restaurants in a map application, review ratings and comments, and then write a summary in a notes app. Similarly, an online shopping task might involve searching for products on a platform like Walmart, watching related reviews on YouTube, and finally summarizing the pros and cons in a notes app.

\paragraph{Comparison Between Mobile-Eval-RAG and Existing Benchmarks}

Table~\ref{tab:benchmark_comparison} presents a comparative analysis between our proposed benchmark, \textbf{Mobile-Eval-RAG}, and several representative mobile automation benchmarks. Mobile-Eval-RAG is divided into two subsets: \textbf{Mobile-Eval-RAG (Simple)} and \textbf{Mobile-Eval-RAG (Complex)}, which correspond to relatively simple and complex task scenarios within the same evaluation framework.

Compared to existing datasets such as Mobile-Eval, DroidTask, and AppAgent, Mobile-Eval-RAG offers notable advantages in several key dimensions. First, it contains a higher proportion of cross-app tasks (100\% in both subsets), which more accurately reflects real-world multi-app collaboration. Second, it features significantly longer task horizons, with an average execution length of 14.05 steps for simple tasks and up to 18.80 steps for complex tasks—highlighting its suitability for evaluating long-horizon reasoning capabilities. Third, unlike most prior datasets (with the exception of Mobile-Eval-E), our benchmark incorporates well-defined \textbf{Completion Rate Evaluation Criteria} (see \underline{\textbf{Appendix I}}), enabling more objective and fine-grained evaluation of task progress.

Furthermore, while both Mobile-Eval-RAG and Mobile-Eval-E aim to support complex automation tasks, they differ significantly in their design focus. Mobile-Eval-E features a broader range of task types and a progressively distributed difficulty spectrum, making it well-suited for evaluating self-learning and lifelong learning in open-ended scenarios. In contrast, Mobile-Eval-RAG emphasizes generalization under controlled task variations, making it especially suitable for benchmarking \textbf{RAG-based systems}. Each task category (e.g., ``Restaurant Recommendation'') in Mobile-Eval-RAG consists of highly similar applications and interaction patterns, but with subtle variations in content and context. This concentrated design yields two key benefits: (1) it enables the reuse of generalized retrieval strategies across tasks within a category, and (2) it prevents systems from relying on rote memorization, thereby requiring deeper understanding and adaptive reasoning. As a result, Mobile-Eval-RAG serves as a more rigorous testbed for evaluating the generalization capabilities of mobile agents in RAG settings.

\begin{table*}[t]
\setlength{\tabcolsep}{1mm} 
\centering
\small 
\begin{tabular}{p{0.2\textwidth}|p{0.2\textwidth}|p{0.55\textwidth}} 
\toprule
\textbf{Category} & \textbf{Apps} & \textbf{Task Instruction} \\
\midrule
\multirow{10}{*}[-48pt]{\textbf{Information Searching}}
& Chrome, Notes & Research the latest green energy innovations from 2025 and summarize top 3 technologies in Notes. \\
& YouTube, Notes & Find 3 healthy breakfast recipes on YouTube that are under 10 mins to cook. Write the summary in Notes. \\
& Chrome, Notes & Find recent space discoveries from NASA or SpaceX. Summarize 2 major ones in Notes. \\
& Chrome, Notes & Find recommended books for beginners in machine learning. Summarize book titles and author recommendations in Notes. \\
& Chrome, Notes & Research top 5 internet safety tips for teenagers in 2025. Write a short guideline in Notes. \\
& Chrome, Notes & Search for electric car models under \$35k in Chrome. Note 3 options and main features in Notes. \\
& YouTube, Notes & Find a beginner-friendly daily yoga video on YouTube. Note down the video title and channel name and write a routine summary in Notes. \\
& Chrome, Notes & Find 2023–2024 news or articles about global plastic bans. Summarize 3 countries' policies in Notes. \\
& Chrome, Notes & Research next 3 major space missions. Write a timeline summary in Notes. \\
& Chrome, Notes & Find articles or guides comparing coffee brewing methods. Summarize key differences and ideal use cases for each method in Notes. \\
\midrule
\multirow{10}{*}[-48pt]{\textbf{What's Trending}}
& X, Notes & Search for 3 fun or useful mobile apps trending on X in 2025. Summarize features in Notes.\\
& Chrome, X, Notes & Look for trending tech startups in 2025. Use X and Chrome to summarize 3 promising ones. \\
& YouTube, Notes & Find the top trending music video on YouTube. Analyze the comments and summarize what people like in Notes. \\
& X, Notes & Check what's trending for holidays in Tokyo. Search on X and summarize top places or events. \\
& YouTube, Notes & Find 3 popular vloggers who post daily life or travel content. Summarize what makes their videos engaging. \\
& X, Notes & Search on X for discussions on 2025 metaverse chat tools. Summarize 3 tools in Notes. \\
& X, Notes & Check recent posts about AI-generated music. Find 2 popular songs or tools and write a summary. \\
& Chrome, Notes & Research what types of games are trending in 2025. Use Chrome and summarize 3 trends in Notes. \\
& YouTube, Notes & Find 3 YouTube Shorts creators with viral content in 2025. Summarize what makes their content engaging. \\
& X, Notes & Look for hype around 2024 memecoins on X. Note top 2 trending coins and community sentiment. \\

\bottomrule
\end{tabular}

\caption{Examples of \textbf{simple} operation tasks proposed by \textbf{Mobile-Eval-RAG}, covering two categories: \textit{Information Searching} and \textit{What's Trending}. Each task specifies a real-world mobile scenario involving 2–3 apps and requires users to retrieve, analyze, and summarize content in Notes.}

\label{tab:Mobile-Agent-RAG_bench_part1}
\end{table*}

\begin{table*}[t]
\small
\setlength{\tabcolsep}{1mm}

\centering

\begin{tabular}{p{0.25\textwidth}|p{0.2\textwidth}|p{0.5\textwidth}} 
\toprule
\textbf{Category} & \textbf{Apps} & \textbf{Task Instruction} \\
\midrule
\multirow{10}{*}[-48pt]{\textbf{Restaurant Recommendation}}
& Maps, Notes & Find the best ramen place in Chicago Loop with at least 500 reviews and rating over 4.5. Write a review summary in Notes. \\
& Maps, Notes & Look for a family-friendly restaurant in Urbana suitable for kids. Write a short summary in Notes. \\
& Maps, Notes & Search for breakfast buffet places near me with good reviews. Compare 2 and write in Notes. \\
& Maps, Notes & Find a hotpot restaurant near a university campus. Write the address and the average user rating into Notes. \\
& Maps, Notes & Find a Chinese restaurant in Chicago with rating over 4.5 that offers takeout. Save 3 dishes and their prices in Notes. \\
& Maps, Notes & Search for nearby vegan breakfast spots. Pick one with best rating and write a short review in Notes. \\
& Maps, Notes & Find a seafood restaurant suitable for a romantic dinner. Include menu highlight in Notes. \\
& Maps, X, Notes & Find a trending brunch place in Chicago in Maps. Check user posts on X and summarize in Notes. \\
& Maps, Notes & Find 3 burger restaurants within 5km. Write a comparison of reviews and prices in Notes. \\
& Maps, Notes & Search for a dessert shop open after 10pm. Check user reviews and note recommended items. \\
\midrule
\multirow{10}{*}[-48pt]{\textbf{Online Shopping}}
& Walmart, Notes & Find a tablet under \$150 on Walmart. Compare 2 brands and summarize specs in Notes. \\
& Walmart, YouTube & Search for a portable speaker under \$100. Watch 1–2 YouTube reviews and write pros/cons in Notes. \\
& Walmart, Notes & Look for an affordable desk lamp for study with eye-care mode. Compare ratings \& product descriptions  and note in Notes. \\
& Walmart, Notes & Find top-rated pet supplies under \$30 for dogs on Walmart. Compare options, prices and descriptions in Notes. \\
& Walmart,YouTube, Notes & Find a student laptop under \$400. Check YouTube reviews and summarize 3 models in Notes. \\
& Walmart,YouTube, Notes & Find an ergonomic chair under \$120 on Walmart. Watch YouTube reviews and write pros/cons in Notes. \\
& Walmart, Notes & Find a blender under \$50 with good reviews on Walmart. Save a note with specs and use cases. \\
& Walmart, Notes & Find a mechanical keyboard with WiFi wireless under \$80. Compare two models and write a summary in Notes. \\
& Walmart, Notes & Look for a budget monitor for under \$150 on Walmart. Research reviews and compare specs in Notes. \\
& Walmart, Notes & Find a student-friendly printer under \$100. Summarize pros/cons and printing cost in Notes. \\
\midrule
\multirow{10}{*}[-48pt]{\textbf{Travel Planning}}
& Booking, Notes & Find a breakfast and Free-WiFi including hotel in Florida under \yen 1750/night for 3 people. Summarize final choice in Notes.\\
& Tripadvisor, Notes & Plan a weekend foodie trip to Chicago. Select 3 top-rated food places from Tripadvisor and write your plan in Notes. \\
& Tripadvisor, Notes & Use Tripadvisor to plan a route to visit 4 museums in Washington, D.C. Include notes on entry fees and hours. \\
& Tripadvisor, Notes & Find 3 top-rated beautiful spots in Arizona using Tripadvisor. Summarize cost, facilities, and activities in Notes. \\
& Tripadvisor, Maps, Notes & Search for a hotel in Seoul close to cultural landmarks. Summarize the hotel info and nearby attractions in Notes. \\
& Booking, Maps, Notes & Find a top-rated hotel in Boston under \yen 1840/night, and note nearby attractions in Notes. \\
& Booking, Maps, Notes & Book a hotel in Orlando for tonight under \yen 800. Check on Maps if it's close to amusement parks. Summarize in Notes.\\
& Tripadvisor, Notes & Plan a 3-day itinerary to San Francisco. Use Tripadvisor to find places to visit, eat, and stay. Summarize in Notes. \\
& Booking, Notes & Find 2 top-rated hotels in Vail, Colorado. Write a comparison of the hotels in Notes. \\
& Tripadvisor, Notes & Find 3 trendy restaurants in Tokyo on Tripadvisor. Summarize their highlights in Notes. \\
\bottomrule
\end{tabular}

\caption{Examples of \textbf{complex} operation tasks proposed by \textbf{Mobile-Eval-RAG}, covering three categories: \textit{Restaurant Recommendation}, \textit{What's Trending} and \textit{Travel Planning}. Each task specifies a real-world mobile scenario involving 2–3 apps and requires users to retrieve, analyze, and summarize content in Notes.}
\label{tab:Mobile-Agent-RAG_bench_part2}
\end{table*}

\section{G. Further Details on Evaluation Metrics}

To comprehensively evaluate the performance of mobile agents in complex automation tasks, we draw inspiration from classic frameworks such as \texttt{Mobile-Agent-E} and \texttt{Mobile-Agent-v2} to construct a multidimensional evaluation metric system. This system encompasses task completion effectiveness (\textbf{Success Rate} and \textbf{Completion Rate}) and fine-grained operational accuracy (\textbf{Operator Accuracy} and \textbf{Reflector Accuracy}), aiming to assess whether the agent can perform tasks correctly and with high quality. While these metrics effectively reflect the agent's competence, they are insufficient for capturing execution efficiency.

Considering the practical importance of efficiency in real-world scenarios such as mobile device automation, we further introduce two complementary metrics: \textbf{Steps} and \textbf{Efficiency}, which measure the operational cost of task completion and the contribution of each step to the task completion progress. Together, these metrics form a systematic, fine-grained, and objective evaluation framework that assesses agent performance from three perspectives—task effectiveness, action-level accuracy, and execution efficiency—making it suitable for the evaluation of long-horizon, multi-step tasks.

\begin{enumerate}

    \item \textbf{Success Rate (SR, \%):}
    The Success Rate evaluates whether a task is completed successfully under three essential conditions: (1) The task is completed within 30 steps; (2) The agent does not make any erroneous task completion judgments; and (3) The agent does not repeat the same action more than five consecutive times. A task is counted as a ``success'' only if all three of these conditions are met. If any one of the conditions is not satisfied, the task is counted as a ``failure''.

    This metric reflects whether the agent can accomplish a goal efficiently, without becoming trapped in repetitive behavior or terminating prematurely.

    \item \textbf{Completion Rate (CR, \%):}
    {Completion Rate} is used to quantify the degree to which a task has been completed. Metrics like SR have been adopted in prior work, such as \texttt{Mobile-Agent-v2}. However, as illustrated in \underline{\textbf{Appendix I}}, its original definition exhibits certain limitations in terms of accuracy and applicability.
    To address this, we tackle the shortcomings of existing metrics by introducing a tailored \textbf{Completion Rate Evaluation Criteria} for each task. Based on these criteria, the Completion Rate is recalculated to more precisely reflect task progress. The formal definition of Completion Rate is provided as follows.

    \begin{equation}
        \text{CR} = \frac{\text{Number of completed items}}{\text{Number of total items}}
        \label{eq:cr}
    \end{equation}
    This metric is particularly useful when tasks are partially completed or vary in structure and length.

    \item \textbf{Operator Accuracy (OA, \%):}
    Operator Accuracy measures how accurately the Operator module selects and executes the correct atomic actions required to advance each subtask. An action is considered correct if it is successfully executed on screen and contributes to progress. The metric is defined as:
    \begin{equation}
        \text{OA} = \frac{\text{Number of correct operations in the task}}{\text{Total steps in the task}}
        \label{eq:oa}
    \end{equation}
    This metric directly reflects the precision of the agent's action selection and execution.

    \item \textbf{Reflector Accuracy (RA, \%):}
    Reflector Accuracy evaluates whether the Action Reflector module can correctly judge the outcome of the Operator's action. It captures the proportion of steps in which the reflection correctly determines whether the action has advanced the current subtask. The metric is defined as:
    \begin{equation}
        \text{RA} = \frac{\text{Number of correct reflections in the task}}{\text{Total steps in the task}}
        \label{eq:ra}
    \end{equation}
    This metric is essential for understanding the system's capacity for self-assessment and timely error correction.

    \item \textbf{Steps:}
    This metric records the number of core operational steps required to complete a task. It serves as a direct measure of task execution cost. In our framework, the maximum number of steps is capped at 30 in accordance with the Success Rate condition.

    \item \textbf{Efficiency:}
    Efficiency captures the average per-step contribution to task completion. It measures how effectively the agent advances toward the goal with each step. A higher efficiency value indicates that each step contributes more significantly to task progress. It is defined as:
    \begin{equation}
        \text{Efficiency} = \frac{\text{CR of the task}}{\text{Total steps in the task}}
        \label{eq:efficiency}
    \end{equation}
    This metric reflects the overall quality of the agent's exploration and decision-making process.

\end{enumerate}

In summary, these evaluation metrics constitute a rigorous and comprehensive framework for assessing mobile agent systems. They jointly capture various aspects of performance, including task-level success, progression detail, operation precision, reflection correctness, and behavioral efficiency. This framework is especially well suited for evaluating Mobile-Agent-RAG under human-in-the-loop settings, where understanding nuanced task behavior and adaptation strategies is essential.

\section{H. Completion Rate Evaluation Criteria}
\label{Completion_Rate_Evaluation_Criterion}

To more accurately assess agent performance in multi-step mobile tasks, this work introduces a fine-grained evaluation mechanism referred to as the \textbf{Completion Rate Evaluation Criteria}. This mechanism is designed to address the limitations of the traditional Success Rate (SR) metric, which provides only a coarse-grained view of task outcomes in complex scenarios.

Conventional SR metrics evaluate task execution using a binary outcome—either ``success'' or ``failure''—for the entire task. This approach fails to capture whether an agent has made partial progress or completed key subcomponents of the task. In real-world mobile settings, tasks often involve a sequence of operations, such as opening applications, searching for content, filling out forms, or interacting across multiple apps. Even when an agent does not complete all steps, it may successfully carry out a majority of essential actions. Therefore, relying solely on SR lacks the granularity needed to fully understand agent performance and hinders targeted model improvements.

To overcome this limitation, each evaluation task is decomposed into a set of clearly defined and independently verifiable subgoals, referred to as \textit{completion items}, which collectively constitute the task's Completion Criteria. The number of subgoals is determined by task complexity: tasks involving two applications are assigned 8 completion items, while tasks involving three applications are assigned 10. Each subgoal can be individually assessed, enabling a fine-grained quantification of task progress. The number of completed subgoals directly maps to the agent's \textbf{Completion Rate (CR)}, providing a more nuanced indicator of intermediate performance.

The initial set of completion items is automatically generated using the large language model \textbf{Gemini-2.5-pro}, which is prompted to identify objective and critical progress points. These model-generated items were then rigorously reviewed and refined by human annotators to ensure clarity, accuracy, and comprehensive task coverage. All subgoals are assigned equal weight, ensuring that the final CR score is both fair and comparable across different tasks.

This evaluation framework brings three key benefits:
\begin{itemize}
    \item \textbf{Fine-grained Measurement:} Offers detailed insight into agent behavior by quantifying intermediate progress, improving objectivity and expressiveness.
    \item \textbf{Complexity-aware Comparison:} Accounts for task complexity by using a standardized structure, enabling fair cross-task comparisons.
    \item \textbf{Interpretability and Reproducibility:} Human-validated subgoals make results more transparent and replicable, helping researchers diagnose performance bottlenecks and improve system design.
\end{itemize}

In summary, the Completion Criteria provide a more stable, precise, and extensible foundation for systematic evaluation of mobile agents, addressing the limitations of binary metrics in multi-step task environments.

\section{I. Experimental Implementations}  
We use Android Debug Bridge\footnote{\url{https://developer.android.google.cn/tools/adb}} (ADB) to allow the Operator Agent to perform atomic actions on real mobile devices. We select 8 widely used mobile applications and design corresponding tasks for multi-application scenarios. Each task allows up to 30 steps, with no more than five consecutive repetitions of the same action. To ensure reproducibility, MLLM API calls are limited to 2048 tokens with a temperature of 0. Human annotators monitor and evaluate system performance in real time using predefined metrics.
For consistency with our \texttt{Mobile-Agent-RAG}'s implementation, we use only the default ``Shortcuts'' and ``Tips'' components of \texttt{Mobile-Agent-E} across all frameworks. For methods requiring pre-training, such as \texttt{Mobile-Agent-RAG} and \texttt{AppAgent}, we allocate 50\% of the Mobile-Eval-RAG dataset for knowledge base construction and use the remaining 50\% as a unified test set to ensure fair and consistent evaluation. Empirically, we set $k=3$ in all exmperiments.

\begin{table*}[t]
\setlength{\tabcolsep}{1mm} 
\centering 
\small 

\begin{tabular}{p{0.17\textwidth}|p{0.3\textwidth}||p{0.17\textwidth}|p{0.3\textwidth}} 
\toprule
\textbf{Task} & \textbf{Completion Items} & \textbf{Task} & \textbf{Completion Items} \\
\midrule

Research the latest green energy innovations from 2025 and summarize top 3 technologies in Notes. &
    \begin{itemize}[leftmargin=*, noitemsep, topsep=0pt, partopsep=0pt]
    \item Opened Chrome
    \item Searched for ``2025 green energy innovations''
    \item Identified at least 3 innovations
    \item Checked technical descriptions or use cases
    \item Compared key features or applications
    \item Opened Notes
    \item Created a new note
    \item Wrote a summary of 3 innovations
    \end{itemize}
& Search for electric car models under \$35k in Chrome. Note 3 options and main features in Notes. &
    \begin{itemize}[leftmargin=*, noitemsep, topsep=0pt, partopsep=0pt]
    \item Opened Chrome
    \item Searched for ``electric cars under \$35k''
    \item Checked top 3 results
    \item Compared specs of at least 3 models
    \item Verified prices are under \$35k
    \item Opened Notes
    \item Created a new note
    \item Wrote key features of 3 models
    \end{itemize} \\ 
\midrule

Find 3 healthy breakfast recipes on YouTube that are under 10 mins to cook. Write the summary in Notes. &
    \begin{itemize}[leftmargin=*, noitemsep, topsep=0pt, partopsep=0pt]
    \item Opened YouTube
    \item Searched for ``healthy breakfast under 10 mins''
    \item Identified 3 suitable recipes
    \item Verified recipe durations
    \item Identified key ingredients
    \item Opened Notes
    \item Created a new note
    \item Wrote a summary of 3 recipes
    \end{itemize}
& Find a beginner-friendly daily yoga video on YouTube. Note down the video title and channel name and write a routine summary in Notes. &
    \begin{itemize}[leftmargin=*, noitemsep, topsep=0pt, partopsep=0pt]
    \item Opened YouTube
    \item Searched for ``daily yoga beginner video''
    \item Selected one video
    \item Verified it's beginner-friendly
    \item Verified title, channel name and routine
    \item Opened Notes
    \item Created a new note
    \item Summarized the routine structure
    \end{itemize} \\ 
\midrule

Find recent space discoveries from NASA or SpaceX. Summarize 2 major ones in Notes. &
    \begin{itemize}[leftmargin=*, noitemsep, topsep=0pt, partopsep=0pt]
    \item Opened Chrome
    \item Searched for ``recent discoveries NASA / SpaceX''
    \item Tap into a relevant article
    \item Selected 2 major discoveries
    \item Checked dates and agencies involved
    \item Opened Notes
    \item Created a new note
    \item Summarized both discoveries
    \end{itemize}
& Find 2023–2024 news or articles about global plastic bans. Summarize 3 countries' policies in Notes. &
    \begin{itemize}[leftmargin=*, noitemsep, topsep=0pt, partopsep=0pt]
    \item Opened Chrome
    \item Searched for ``plastic ban policies in 2023-2024''
    \item Tap into a relevant article
    \item Identified 3 country-specific sources
    \item Noted banned items and start dates
    \item Opened Notes
    \item Created a new note
    \item Summarized each policy
    \end{itemize} \\ 
\midrule

Find recommended books for beginners in machine learning. Summarize book titles and author recommendations in Notes. &
    \begin{itemize}[leftmargin=*, noitemsep, topsep=0pt, partopsep=0pt]
    \item Opened Chrome
    \item Searched for ``beginner machine learning books''
    \item Visited at least one curated list
    \item Selected 3 recommended books
    \item Checked author information
    \item Opened Notes
    \item Created a new note
    \item Listed book titles and authors
    \end{itemize}
& Research next 3 major space missions. Write a timeline summary in Notes. &
    \begin{itemize}[leftmargin=*, noitemsep, topsep=0pt, partopsep=0pt]
    \item Opened Chrome
    \item Searched for ``upcoming space missions 2025''
    \item Tap into a relevant official article
    \item Found missions from different agencies
    \item Noted launch dates and goals
    \item Opened Notes
    \item Created a new note
    \item Wrote mission timeline
    \end{itemize} \\ 
\midrule

Research top 5 internet safety tips for teenagers in 2025. Write a short guideline in Notes. &
    \begin{itemize}[leftmargin=*, noitemsep, topsep=0pt, partopsep=0pt]
    \item Opened Chrome
    \item Searched for ``internet safety tips teenagers 2025''
    \item Tap into a relevant article
    \item Read multiple guides
    \item Selected 5 actionable tips
    \item Opened Notes
    \item Created a new note
    \item Listed 5 safety tips
    \end{itemize}
& Find articles or guides comparing coffee brewing methods. Summarize key differences and ideal use cases for each method in Notes. &
    \begin{itemize}[leftmargin=*, noitemsep, topsep=0pt, partopsep=0pt]
    \item Opened Chrome
    \item Searched for ``coffee brewing methods comparison''
    \item Found a detailed article
    \item Found more detailed articles
    \item Compared time, flavor, ease
    \item Opened Notes
    \item Created a new note
    \item Wrote a method-by-method summary and compare differences and use cases
    \end{itemize} \\
\bottomrule
\end{tabular}
\caption{
Examples of \textbf{Completion Rate Evaluation Criteria} for the \textit{Information Searching} task category in the Multi-Eval-RAG dataset. Each task is decomposed into a sequential list of atomic \textbf{Completion Items}, serving as step-wise indicators for objectively evaluating task Completion Rate (CR). The table is structured into pairs of columns: the \textbf{Task} column shows the natural language instruction, while the corresponding \textbf{Completion Items} column enumerates the necessary steps to fulfill the task. Tasks involving two applications (e.g., Chrome and Notes) contain 8 items, while those involving three applications contain 10 items, enabling more fine-grained progress tracking for long-horizon or multi-app tasks.}

\label{criteria1}
\end{table*}

\begin{table*}[t] 
 \setlength{\tabcolsep}{1mm} 
 \centering 
 \small 

 \begin{tabular}{p{0.17\textwidth}|p{0.3\textwidth}||p{0.17\textwidth}|p{0.3\textwidth}} 
 \toprule
 \textbf{Task} & \textbf{Completion Items} & \textbf{Task} & \textbf{Completion Items} \\ 
 \midrule 

 Search for 3 fun or useful mobile apps trending on X in 2025. Summarize features in Notes. & 
    \begin{itemize}[leftmargin=*, noitemsep, topsep=0pt, partopsep=0pt]
    \item Opened X 
    \item Searched for ``trending mobile apps 2025''
    \item Identified 3 different apps 
    \item Reviewed posts describing app features 
    \item Verified popularity through likes 
    \item Opened Notes 
    \item Created a new note 
    \item Wrote short feature summary for the 3 
    \end{itemize}
 & Search on X for discussions on 2025 metaverse chat tools. Summarize 3 tools in Notes. & 
    \begin{itemize}[leftmargin=*, noitemsep, topsep=0pt, partopsep=0pt]
    \item Opened X 
    \item Searched for ``metaverse chat tools 2025'' 
    \item Found 3 frequently mentioned tools 
    \item Read user discussions 
    \item Identified key features and opinions 
    \item Opened Notes 
    \item Created a new note 
    \item Wrote a short summary of the 3 tools
    \end{itemize} \\ 
 \midrule 

 Look for trending tech startups in 2025. Use X and Chrome to summarize 3 promising ones. & 
    \begin{itemize}[leftmargin=*, noitemsep, topsep=0pt, partopsep=0pt]
    \item Opened Chrome 
    \item Searched for ``trending tech startups 2025'' 
    \item Identified 3 promising companies 
    \item Opened X 
    \item Searched each startup name 
    \item Reviewed user sentiment or press mentions 
    \item Opened Notes 
    \item Created a new note 
    \item Summarized startup strengths 
    \item Mentioned industries 
    \end{itemize}
 & Check recent posts about AI-generated music. Find 2 popular songs or tools and write a summary. & 
    \begin{itemize}[leftmargin=*, noitemsep, topsep=0pt, partopsep=0pt]
    \item Opened X 
    \item Searched for ``AI-generated music in 2025''  
    \item Found 2 popular songs/tools 
    \item Reviewed likes/comments for verifying popularity
    \item Opened Notes 
    \item Created a new note 
    \item Mentioned AI tool or artist 
    \item Summarized feedback 
    \end{itemize} \\ 
 \midrule 

 Find the top trending music video on YouTube. Analyze the comments and summarize what people like in Notes. & 
    \begin{itemize}[leftmargin=*, noitemsep, topsep=0pt, partopsep=0pt]
    \item Opened YouTube 
    \item Searched for ``top trending music video'' 
    \item Clicked top video 
    \item Reviewed likes/multiple user comments for checking trending
    \item Identified common praise points 
    \item Opened Notes 
    \item Created a new note 
    \item Summarized viewer highlights 
    \end{itemize}
 & Research what types of games are trending in 2025. Use Chrome and summarize 3 trends in Notes. & 
    \begin{itemize}[leftmargin=*, noitemsep, topsep=0pt, partopsep=0pt]
    \item Opened Chrome 
    \item Searched for ``game trends 2025'' 
    \item Found 1–2 gaming articles 
    \item Identified 3 distinct trends 
    \item Reviewed game titles as examples 
    \item Opened Notes 
    \item Created a new note 
    \item Wrote short trend summary
    \end{itemize} \\ 
 \midrule 

 Check what's trending for holidays in Tokyo. Search on X and summarize top places or events. & 
    \begin{itemize}[leftmargin=*, noitemsep, topsep=0pt, partopsep=0pt]
    \item Opened X 
    \item Searched ``holidays in Tokyo'' 
    \item Found relevant trending posts 
    \item Identified 2–3 locations/events 
    \item Checked photo/video content 
    \item Opened Notes 
    \item Created a new note 
    \item Summarized top places or events 
    \end{itemize}
 & Find 3 YouTube Shorts creators with viral content in 2025. Summarize what makes their content engaging. & 
    \begin{itemize}[leftmargin=*, noitemsep, topsep=0pt, partopsep=0pt]
    \item Opened YouTube 
    \item Searched for ``YouTube Shorts creators with viral content''
    \item Tap into Shorts
    \item Identified 3 viral creators 
    \item Noted visual/editing style 
    \item Opened Notes 
    \item Created a new note 
    \item Summarized engaging features
    \end{itemize} \\ 
 \midrule 

 Find 3 popular vloggers who post daily life or travel content. Summarize what makes their videos engaging. & 
    \begin{itemize}[leftmargin=*, noitemsep, topsep=0pt, partopsep=0pt]
    \item Opened YouTube 
    \item Searched for ``top vloggers 2025 travel/daily life'' 
    \item Picked 3 with high views/subscribers 
    \item Identified common themes 
    \item Opened Notes 
    \item Created a new note 
    \item Summarized engagement factors 
    \item Mentioned creator names 
    \end{itemize}
 & Look for hype around 2024 memecoins on X. Note top 2 trending coins and community sentiment. & 
    \begin{itemize}[leftmargin=*, noitemsep, topsep=0pt, partopsep=0pt]
    \item Opened X 
    \item Searched for ``2024 memecoins'' 
    \item Identified 2 frequently mentioned coins 
    \item Reviewed comments for popular threads 
    \item Analyzed tone and hype level 
    \item Opened Notes 
    \item Created a new note 
    \item Wrote short summary for top 2 trending coins and community sentiment
    \end{itemize} \\ 

 \bottomrule 
 \end{tabular} 

 \caption{
Examples of \textbf{Completion Rate Evaluation Criteria} for the \textit{What's Trending} task category in the Multi-Eval-RAG dataset. Each task is decomposed into a sequential list of atomic \textbf{Completion Items}, serving as step-wise indicators for objectively evaluating task Completion Rate (CR). The table is structured into pairs of columns: the \textbf{Task} column shows the natural language instruction, while the corresponding \textbf{Completion Items} column enumerates the necessary steps to fulfill the task. Tasks involving two applications (e.g., Chrome and Notes) contain 8 items, while those involving three applications contain 10 items, enabling more fine-grained progress tracking for long-horizon or multi-app tasks.
}

 \label{criteria2}
 \end{table*}

\begin{table*}[t]
\setlength{\tabcolsep}{1mm} 
\centering
\small

\begin{tabular}{p{0.17\textwidth}|p{0.3\textwidth}||p{0.17\textwidth}|p{0.3\textwidth}}
\toprule 
\textbf{Task} & \textbf{Completion Items} & \textbf{Task} & \textbf{Completion Items} \\
\midrule 
Find the best ramen place in Chicago Loop with at least 500 reviews and rating over 4.5. Write a review summary in Notes. &
    \begin{itemize}[leftmargin=*, noitemsep, topsep=0pt, partopsep=0pt]
    \item Opened Maps
    \item Searched for ``ramen restaurants in Chicago Loop''
    \item Identified restaurants with $> 500$ reviews
    \item Identified restaurants with rating $> 4.5$
    \item Selected the best-rated candidate
    \item Opened Notes
    \item Created a new note
    \item Wrote a review summary mentioning key strengths, rating and review count in summary
    \end{itemize}
& Search for nearby vegan breakfast spots. Pick one with best rating and write a short review in Notes. &
    \begin{itemize}[leftmargin=*, noitemsep, topsep=0pt, partopsep=0pt]
    \item Opened Maps
    \item Searched for ``vegan breakfast near me''
    \item Evaluated ratings and reviews
    \item Verified vegan menu items
    \item Selected the top-rated spot
    \item Opened Notes
    \item Created a new note    
    \item Wrote a short review
    \end{itemize} \\
\midrule

Look for a family-friendly restaurant in Urbana suitable for kids. Write a short summary in Notes. &
    \begin{itemize}[leftmargin=*, noitemsep, topsep=0pt, partopsep=0pt]
    \item Opened Maps
    \item Searched for ``family-friendly restaurant Urbana''
    \item Checked at least one restaurant's rating
    \item Verified that restaurant offers kids menu
    \item Selected one preferred restaurant
    \item Opened Notes
    \item Created a new note
    \item Wrote a summary about the selected restaurant
    \end{itemize}
&Find a seafood restaurant suitable for a romantic dinner. Include menu highlight in Notes. &
    \begin{itemize}[leftmargin=*, noitemsep, topsep=0pt, partopsep=0pt]
    \item Opened Maps
    \item Searched for ``seafood restaurant''
    \item Checked ambiance-related reviews for suitability
    \item Checked menu options
    \item Identified one with good romantic setting
    \item Opened Notes
    \item Created a new note
    \item Wrote menu highlights in Notes
    \end{itemize} \\ 
\midrule

Search for breakfast buffet places near me with good reviews. Compare 2 and write in Notes. &
    \begin{itemize}[leftmargin=*, noitemsep, topsep=0pt, partopsep=0pt]
    \item Opened Maps
    \item Searched for ``breakfast buffet near me''
    \item Identified a buffet place
    \item Identified two buffet places with good reviews
    \item Compared menu offerings and prices
    \item Opened Notes
    \item Created a new note
    \item Wrote a comparison between the two places 
    \end{itemize}
&Find a trending brunch place in Chicago in Maps. Check user posts on X and summarize in Notes. &
    \begin{itemize}[leftmargin=*, noitemsep, topsep=0pt, partopsep=0pt]
    \item Opened Maps
    \item Searched for ``trending brunch place in Chicago''
    \item Identified 1-2 candidate restaurants
    \item Opened X
    \item Searched restaurant names on X
    \item Reviewed relevant user posts
    \item Collected highlights or common opinions
    \item Opened Notes
    \item Created a new note
    \item Wrote summary of user opinions
    \end{itemize} \\
\midrule

Find a hotpot restaurant near a university campus. Write the address and the average user rating into Notes. &
    \begin{itemize}[leftmargin=*, noitemsep, topsep=0pt, partopsep=0pt]
    \item Opened Maps
    \item Searched for ``hotpot restaurant near university campus''
    \item Verify at least one university on the map
    \item Checked restaurant ratings
    \item Selected one hotpot restaurant
    \item Opened Notes
    \item Created a new note
    \item Wrote address and rating into note 
    \end{itemize}
& Find 3 burger restaurants within 5km. Write a comparison of reviews and prices in Notes. &
    \begin{itemize}[leftmargin=*, noitemsep, topsep=0pt, partopsep=0pt]
    \item Opened Maps
    \item Searched for ``burger restaurants''
    \item Applied distance filter
    \item Selected 3 distinct restaurants
    \item Reviewed user comments and prices
    \item Opened Notes
    \item Created a new note
    \item Wrote review and price comparison
    \end{itemize} \\
\midrule

Find a Chinese restaurant in Chicago with rating over 4.5 that offers takeout. Save 3 dishes and their prices in Notes. &
    \begin{itemize}[leftmargin=*, noitemsep, topsep=0pt, partopsep=0pt]
    \item Opened Maps
    \item Searched for ``Chinese restaurant in Chicago''
    \item Filtered for 4.5 star
    \item Filtered for takeout
    \item Checked menu
    \item Opened Notes
    \item Created a new note
    \item Wrote a summary explaining why it's liked 
    \end{itemize}
& Search for a dessert shop open after 10pm. Check user reviews and note recommended items. &
    \begin{itemize}[leftmargin=*, noitemsep, topsep=0pt, partopsep=0pt]
    \item Opened Maps
    \item Searched for ``dessert shop open after 10pm''
    \item Verified business hours
    \item Checked user reviews
    \item Identified recommended items
    \item Opened Notes
    \item Created a new note
    \item Wrote summary of user recommendations
    \end{itemize} \\

\bottomrule
\end{tabular}

\caption{
Examples of \textbf{Completion Rate Evaluation Criteria} for the \textit{Restaurant Recommendation} task category in the Multi-Eval-RAG dataset. Each task is decomposed into a sequential list of atomic \textbf{Completion Items}, serving as step-wise indicators for objectively evaluating task Completion Rate (CR). The table is structured into pairs of columns: the \textbf{Task} column shows the natural language instruction, while the corresponding \textbf{Completion Items} column enumerates the necessary steps to fulfill the task. Tasks involving two applications (e.g., Chrome and Notes) contain 8 items, while those involving three applications contain 10 items, enabling more fine-grained progress tracking for long-horizon or multi-app tasks.
}
\label{criteria3}
\end{table*}

\begin{table*}[t]
\setlength{\tabcolsep}{1mm} 
\centering 
\small 

\begin{tabular}{p{0.17\textwidth}|p{0.3\textwidth}||p{0.17\textwidth}|p{0.3\textwidth}}  
\toprule 
\textbf{Task} & \textbf{Completion Items} & \textbf{Task} & \textbf{Completion Items} \\
\midrule

Find a tablet under \$150 on Walmart. Compare 2 brands and summarize specs in Notes. &
    \begin{itemize}[leftmargin=*, noitemsep, topsep=0pt, partopsep=0pt]
    \item Opened Walmart
    \item Searched for ``tablet under \$150''
    \item Found at least two brand options
    \item Compared screen size and storage
    \item Reviewed technical specs
    \item Opened Notes
    \item Created a new note
    \item Wrote brand comparison
    \end{itemize}
& Find an ergonomic chair under \$120 on Walmart. Watch YouTube reviews and write pros/cons in Notes. &
    \begin{itemize}[leftmargin=*, noitemsep, topsep=0pt, partopsep=0pt]
    \item Opened Walmart
    \item Searched for ``ergonomic chair under \$120''
    \item Chose 1–2 chairs
    \item Open YouTube
    \item Searched for selected chairs
    \item Reviewed videos comment on YouTube
    \item Identified comfort and features
    \item Opened Notes
    \item Created a new note
    \item Summarized strengths and weaknesses
    \end{itemize} \\ 
\midrule

Search for a portable speaker under \$100. Watch 1–2 YouTube reviews and write pros/cons in Notes. &
    \begin{itemize}[leftmargin=*, noitemsep, topsep=0pt, partopsep=0pt]
    \item Opened Walmart
    \item Searched for ``portable speaker under \$100''
    \item Selected 1–2 speaker models
    \item Opened YouTube
    \item Searched for selected speakers
    \item Reviewed videos comment on YouTube
    \item Identified pros and cons
    \item Opened Notes
    \item Created a new note
    \item Wrote pros and cons summary
    \end{itemize}
& Find a blender under \$50 with good reviews on Walmart. Save a note with specs and use cases. &
    \begin{itemize}[leftmargin=*, noitemsep, topsep=0pt, partopsep=0pt]
    \item Opened Walmart
    \item Searched for ``blender under \$50''
    \item Filtered by high ratings
    \item Read reviews
    \item Verified blender features
    \item Opened Notes
    \item Created a new note
    \item Wrote down key specs and use cases
    \end{itemize} \\ 
\midrule

Look for an affordable desk lamp for study with eye-care mode. Compare ratings \& product descriptions  and note in Notes. &
    \begin{itemize}[leftmargin=*, noitemsep, topsep=0pt, partopsep=0pt]
    \item Opened Walmart
    \item Searched for ``eye-care desk lamp for study''
    \item Found multiple lamp options
    \item Checked product descriptions
    \item Reviewed customer ratings
    \item Opened Notes
    \item Created a new note
    \item Wrote lamp comparison
    \end{itemize}
& Find a mechanical keyboard with WiFi wireless under \$80. Compare two models and write a summary in Notes. &
    \begin{itemize}[leftmargin=*, noitemsep, topsep=0pt, partopsep=0pt]
    \item Opened Walmart
    \item Searched for ``mechanical keyboard under \$80''
    \item Filter Wifi wireless
    \item Choose 2 models
    \item Compared features
    \item Opened Notes
    \item Created a new note
    \item Summarized model comparison
    \end{itemize} \\ 
\midrule

Find top-rated pet supplies under \$30 for dogs on Walmart. Compare options, prices and descriptions in Notes. &
    \begin{itemize}[leftmargin=*, noitemsep, topsep=0pt, partopsep=0pt]
    \item Opened Walmart
    \item Searched for ``dog supplies under \$30''
    \item Filtered by rating
    \item Selected multiple items
    \item Checked item descriptions
    \item Opened Notes
    \item Created a new note
    \item Compare top 2–3 products 
    \end{itemize}
& Look for a budget monitor for under \$150 on Walmart. Research reviews and compare specs in Notes. &
    \begin{itemize}[leftmargin=*, noitemsep, topsep=0pt, partopsep=0pt]
    \item Opened Walmart
    \item Searched for ``budget monitor under \$150''
    \item Found multiple options
    \item Compared screen resolution and size
    \item Checked product details
    \item Opened Notes
    \item Created a new note
    \item Summarized and compared specs
    \end{itemize} \\ 
\midrule

Find a student laptop under \$400. Check YouTube reviews and summarize 3 models in Notes. &
    \begin{itemize}[leftmargin=*, noitemsep, topsep=0pt, partopsep=0pt]
    \item Opened Walmart
    \item Searched for ``laptop under \$400''
    \item Identified 3 potential models
    \item Compared specs (RAM, CPU, storage)
    \item Opened YouTube
    \item Searched for selected laptops
    \item Reviewed videos comment on YouTube
    \item Opened Notes
    \item Created a new note
    \item Wrote summary of 3 laptops
    \end{itemize}
& Find a student-friendly printer under \$100. Summarize pros/cons and printing cost in Notes. &
    \begin{itemize}[leftmargin=*, noitemsep, topsep=0pt, partopsep=0pt]
    \item Opened Walmart
    \item Searched for ``student-friendly printer under \$100''
    \item Read a product page
    \item Checked prices, printing speed and type
    \item Reviewed comments for pros/cons
    \item Opened Notes
    \item Created a new note
    \item Summary the selected printers' pros/cons and cost
    \end{itemize} \\

\bottomrule
\end{tabular}
\caption{
Examples of \textbf{Completion Rate Evaluation Criteria} for the \textit{Online Shopping} task category in the Multi-Eval-RAG dataset. Each task is decomposed into a sequential list of atomic \textbf{Completion Items}, serving as step-wise indicators for objectively evaluating task Completion Rate (CR). The table is structured into pairs of columns: the \textbf{Task} column shows the natural language instruction, while the corresponding \textbf{Completion Items} column enumerates the necessary steps to fulfill the task. Tasks involving two applications (e.g., Chrome and Notes) contain 8 items, while those involving three applications contain 10 items, enabling more fine-grained progress tracking for long-horizon or multi-app tasks.
}
\label{criteria4}
\end{table*}

\begin{table*}[t]
\setlength{\tabcolsep}{1mm}
\centering
\small

\begin{tabular}{p{0.17\textwidth}|p{0.3\textwidth}||p{0.17\textwidth}|p{0.3\textwidth}}
\toprule
\textbf{Task} & \textbf{Completion Items} & \textbf{Task} & \textbf{Completion Items} \\
\midrule

Find a breakfast and Free-WiFi including hotel in Florida under \yen 1750/night for 3 people. Summarize final choice in Notes. &
    \begin{itemize}[leftmargin=*, noitemsep, topsep=0pt, partopsep=0pt]
    \item Opened Booking
    \item Filter for 3 people
    \item Searched for ``Florida''
    \item Filter for breakfast and Free-WiFi
    \item Reviewed at least 2 options under the prices
    \item Opened Notes
    \item Created a new note
    \item Wrote hotel summary
    \end{itemize}
& Find a top-rated hotel in Boston under \yen 1840/night, and note nearby attractions in Notes. &
    \begin{itemize}[leftmargin=*, noitemsep, topsep=0pt, partopsep=0pt]
    \item Opened Booking
    \item Searched for ``Boston''
    \item Applied top-rated filters
    \item Selected best matching hotel
    \item Open Maps
    \item Searched for selected hotel
    \item Checked nearby attractions
    \item Opened Notes
    \item Created a new note
    \item Summarized hotel and nearby spots
    \end{itemize} \\
\midrule

Plan a weekend foodie trip to Chicago. Select 3 top-rated food places from Tripadvisor and write your plan in Notes. &
    \begin{itemize}[leftmargin=*, noitemsep, topsep=0pt, partopsep=0pt]
    \item Opened Tripadvisor
    \item Searched for ``top-rated food places in Chicago''
    \item Applied filters for cuisine or rating
    \item Selected 3 restaurants
    \item Reviewed descriptions and ratings
    \item Opened Notes
    \item Created a new note
    \item Wrote a foodie trip plan
    \end{itemize}
& Book a hotel in Orlando for tonight under \yen 800. Check on Maps if it's close to amusement parks. Summarize in Notes. &
    \begin{itemize}[leftmargin=*, noitemsep, topsep=0pt, partopsep=0pt]
    \item Opened Booking
    \item Searched for ``Orlando''
    \item Selected a hotel under \yen 800
    \item Open Maps
    \item Searched for the selected hotel
    \item Checked distance to amusement parks
    \item Opened Notes
    \item Created a new note
    \item Wrote booking summary
    \item Listed name, price and the distances to amusement parks
    \end{itemize} \\
\midrule

Use Tripadvisor to plan a route to visit 4 museums in Washington, D.C. Include notes on entry fees and hours. &
    \begin{itemize}[leftmargin=*, noitemsep, topsep=0pt, partopsep=0pt]
    \item Opened Tripadvisor
    \item Searched for ``museums in Washington D.C.''
    \item Selected 4 nearby museums
    \item Checked entry fees and hours
    \item Planned a route
    \item Opened Notes
    \item Created a new note
    \item Wrote museum route plan and 
    \item Listed fees and open times
    \end{itemize}
& Plan a 3-day itinerary to San Francisco. Use Tripadvisor to find places to visit, eat, and stay. Summarize in Notes. &
    \begin{itemize}[leftmargin=*, noitemsep, topsep=0pt, partopsep=0pt]
    \item Opened Tripadvisor
    \item Searched for ``San Francisco''
    \item Found places to visit for 3 days
    \item Looked for dining and hotel options
    \item Opened Notes
    \item Created a new note
    \item Wrote 3-day plan
    \item Listed sites, restaurants, and hotels
    \end{itemize} \\
\midrule

Find 3 top-rated beautiful spots in Arizona using Tripadvisor. Summarize cost, facilities, and activities in Notes. &
    \begin{itemize}[leftmargin=*, noitemsep, topsep=0pt, partopsep=0pt]
    \item Opened Tripadvisor
    \item Searched for ``Arizona''
    \item Selected 3 top-rate spots
    \item Reviewed costs and facilities
    \item Checked available activities
    \item Opened Notes
    \item Created a new note
    \item Summarized info for each spots
    \end{itemize}
& Find 2 top-rated hotels in Vail, Colorado. Write a comparison of the hotels in Notes. &
    \begin{itemize}[leftmargin=*, noitemsep, topsep=0pt, partopsep=0pt]
    \item Opened Booking
    \item Searched for ``Colorado''
    \item Filtered for ``Vail''
    \item Filtered for top-rated
    \item Selected 2 relevant options
    \item Opened Notes
    \item Created a new note
    \item Listed features for each
    \end{itemize} \\
\midrule

Search for a hotel in Seoul close to cultural landmarks. Summarize the hotel info and nearby attractions in Notes. &
    \begin{itemize}[leftmargin=*, noitemsep, topsep=0pt, partopsep=0pt]
    \item Opened Tripadvisor
    \item Searched ``Seoul hotels near cultural landmarks''
    \item Found relevant listings
    \item Open Maps
    \item Searched selected hotels
    \item Verified landmarks near the hotels
    \item Opened Notes
    \item Created a new note
    \item Summarized hotel info
    \item Listed nearby cultural spots
    \end{itemize}
& Find 3 trendy restaurants in Tokyo on Tripadvisor. Summarize their highlights in Notes. &
    \begin{itemize}[leftmargin=*, noitemsep, topsep=0pt, partopsep=0pt]
    \item Opened Tripadvisor
    \item Searched ``trendy restaurants Tokyo''
    \item Selected 3 distinct options
    \item Read user comments
    \item Noted dish highlights or decor themes
    \item Opened Notes
    \item Created a new note
    \item Summarized 3 restaurants
    \end{itemize} \\
\bottomrule
\end{tabular}

\caption{
Examples of \textbf{Completion Rate Evaluation Criteria} for the \textit{Travel Planning} task category in the Multi-Eval-RAG dataset. Each task is decomposed into a sequential list of atomic \textbf{Completion Items}, serving as step-wise indicators for objectively evaluating task Completion Rate (CR). The table is structured into pairs of columns: the \textbf{Task} column shows the natural language instruction, while the corresponding \textbf{Completion Items} column enumerates the necessary steps to fulfill the task. Tasks involving two applications (e.g., Chrome and Notes) contain 8 items, while those involving three applications contain 10 items, enabling more fine-grained progress tracking for long-horizon or multi-app tasks.
}
\label{criteria5}
\end{table*}

\section{J. Computational Cost}
\label{appendix:costs_and_kb}

This section provides further details on the computational costs of the proposed framework during execution and the resources required for constructing the knowledge bases.

\paragraph{Execution Latency and Computational Cost}
To assess the real-time performance of our dual retrieval system, we measured the average latency and token consumption for each component during task execution. All API-based agents (Manager, Operator, ActionReflector, Notetaker) were run using the Gemini-1.5-pro-latest API. The Perceptor module runs locally, utilizing models such as Qwen-VL-Plus, and thus does not incur API token costs.

As detailed in Table~\ref{tab:execution_costs}, the total average time for a complete ``core loop'' (all components) is approximately \textbf{38.71 seconds}. While this is acceptable for a research prototype, optimizing this latency is a key direction for future work.

\begin{table*}[h]
\centering

\setlength{\tabcolsep}{6pt}
\begin{tabular}{l c c}
\toprule
\textbf{Component} & \textbf{Average Latency (s)} & \textbf{Average Tokens (Input + Output)} \\
\midrule
Manager + Manager-RAG & 6.87  & 1059 + 210 \\
Operator + Operator-RAG & 6.63  & 2098 + 63 \\
Perceptor & 11.14 s & N/A  \\
Action Reflector & 5.91  & 2114 + 66 \\
Notetaker & 6.99  & 1272 + 191 \\
\midrule
\textbf{Total Core Loop (Avg)} & \textbf{38.71} & 6543 + 530 (Perceptor removed) \\
\bottomrule
\end{tabular}
\caption{Average latency and token consumption per component during real-time task execution. API token costs are not applicable (N/A) for the locally-run Perceptor module.}
\label{tab:execution_costs}

\end{table*}

\paragraph{KB Construction Cost}
We provide further details on the size and construction cost of the two knowledge bases (KBs). The KBs were constructed using a semi-automatic process from the 25 tasks in the training split (50\% of Mobile-Eval-RAG).

\begin{itemize}
    \item \textbf{KB Size.} The \textbf{Manager-RAG KB} contains \textbf{50} high-level (task instruction, human steps) documents. The \textbf{Operator-RAG KB} is partitioned by app, with document counts for our evaluation set as follows: Booking (25), Chrome (17), Notes (5), Tripadvisor (27), Walmart (23), X (21), YouTube (21), and Maps (28).

    \item \noindent\textbf{Construction Cost.} The entire process required approximately \textbf{5 hours} (avg. 12.13 minutes per task). The process is lightweight and runnable on a single NVIDIA 4060 GPU (8G VRAM), as the bottleneck is API latency, not local computation. Using Gemini 1.5 Pro, data generation averaged 10,028 tokens per step (188,526 per task). This resulted in an estimated total construction cost of \textbf{\$74.25} (approx. \$2.97 per task) for the 25-task training set.
    
\end{itemize}

\section{K. Additional Ablation Study on Core Components}
\label{appendix:ablation_core_components}

In this section,  we conducted additional ablation analysis to quantify the impact of removing other core components: \textbf{Notetaker} and \textbf{Action Reflector}.
This study quantifies their non-retrieval contributions to task success. Results are summarized in Table~\ref{tab:core_ablation}. As illustrated,  the Notetaker is essential for maintaining contextual working memory, while the Action Reflector is indispensable for self-correction and robust execution.

\begin{table*}[h]
\centering

\setlength{\tabcolsep}{6pt}
\begin{tabular}{l c c c c}
\toprule
\textbf{Ablation Configuration} & \textbf{SR (\%)} & \textbf{CR (Change)} & \textbf{OA (Change)} & \textbf{RA (Change)} \\
\midrule
Action Reflector & 24\% & -23.51\% & No Sig. Change & N/A \\
Notetaker & 20\% & -11.67\% & -5.32\% & No Sig. Change \\

\bottomrule
\end{tabular}
\caption{Impact of removing core non-RAG components. SR values are the absolute resulting scores, while CR and OA represent the percentage drop relative to the full framework. ``No Sig. Change'' indicates no significant variation was observed. ``N/A'' (Not Applicable) is used for RA as the component was removed.}
\label{tab:core_ablation}

\end{table*}

The results demonstrate the critical, distinct functions of these modules:
\begin{itemize}
    \item The removal of \textbf{Notetaker} causes a catastrophic drop in \textbf{Success Rate (SR)} to 20\%. This is because most long-horizon tasks require referencing information gathered in previous steps (e.g., writing a summary), which is impossible without the Notetaker's working memory. This dependency also led to a \textbf{11.67\%} drop in \textbf{Completion Rate (CR)} and a \textbf{5.32\%} drop in \textbf{Operator Accuracy (OA)}.
    
    \item Removing \textbf{Action Reflector} also causes a major \textbf{SR} drop (to 24\%) and an even larger \textbf{CR} drop (\textbf{-23.51\%}). This highlights that the agent frequently makes minor errors that, without the Reflector's feedback loop, cascade into task failure. As the Reflector acts *after* the Operator, its removal did not significantly impact \textbf{OA}, but it made recovery from errors impossible. The \textbf{RA} metric becomes Not Applicable (N/A) as the component was removed.
\end{itemize}

\section{L. Case Study}
\label{sec:case-study}

\begin{figure*}[t]\centering\includegraphics[width=0.80\textwidth]{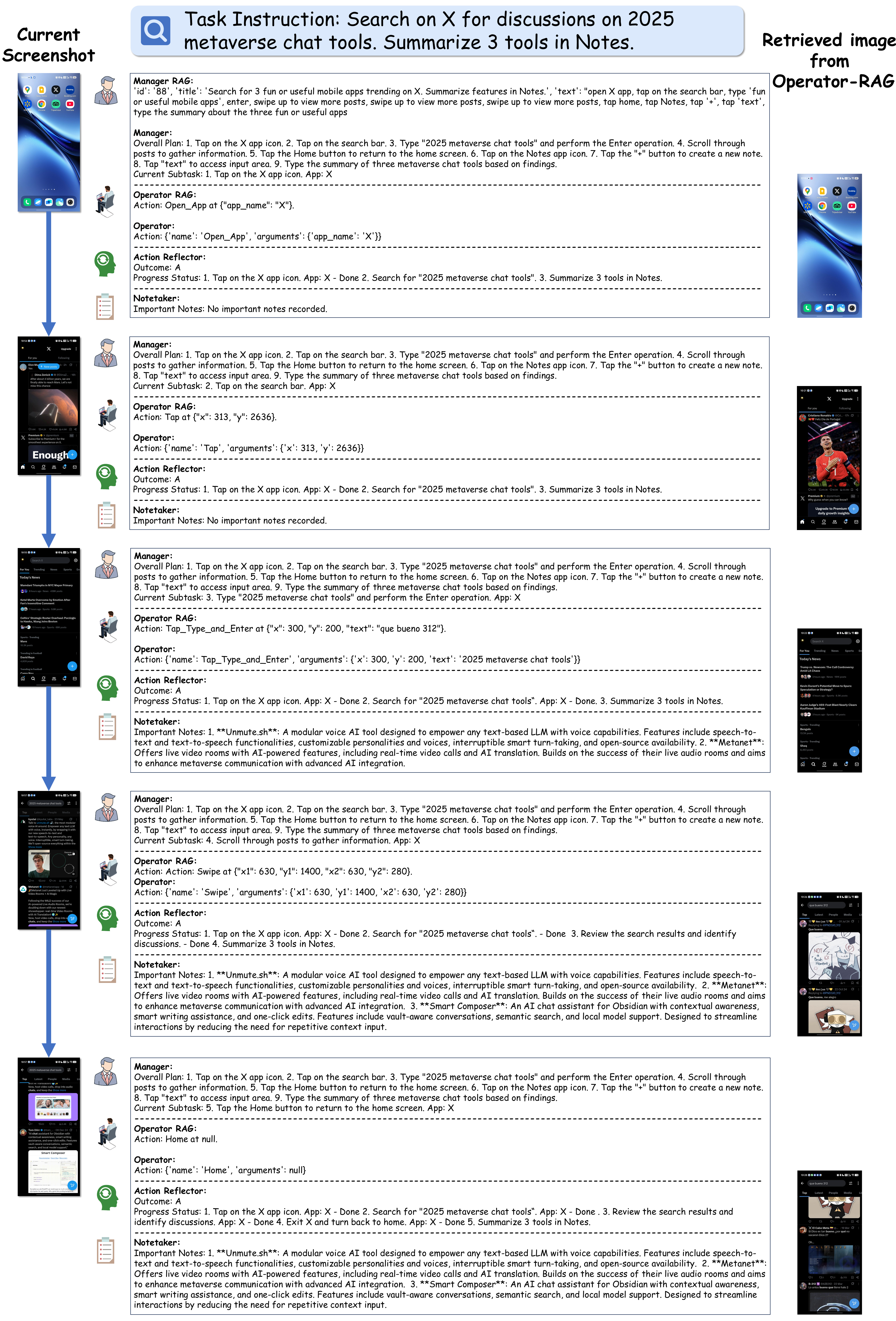}\caption{\small An end-to-end execution case of a task. Each box represents an iterative loop, progressing chronologically from top to bottom. The ``\textbf{Current Screenshot}'' column on the left shows the mobile device's real-time UI screen state. The central section details the inter-agent communication, where a ``\textbf{Manager-RAG}'' retrieves the most relevant document (we set k=1 here) to provide high-level guidance. The ``\textbf{Manager}'' section presents the overall plan and the current subtask. The ``\textbf{Operator-RAG}'' section shows the best-matched action retrieved from its knowledge base for the current subtask, while the ``Retrieved image from Operator-RAG'' on the right provides reference screenshot. The ``\textbf{Operator}'' then synthesizes this information to generate a precise action with its argument. After that, The ``\textbf{Action Reflector}'' records the outcome and updates the progress status. Finally, the ``\textbf{Notetaker}'' records important information as ``Important Notes'' for future context. This process clearly depicts how the agents, through a ``\textbf{Planning\,→\,Execution\,→\,Reflection}'' loop, retrieval augmentation, step-by-step execution, and continuous feedback, progressively complete complex, cross-app tasks. (Process continued with next figure)}\label{fig:cs1}\end{figure*}

\begin{figure*}[t]\centering\includegraphics[width=0.80\textwidth]{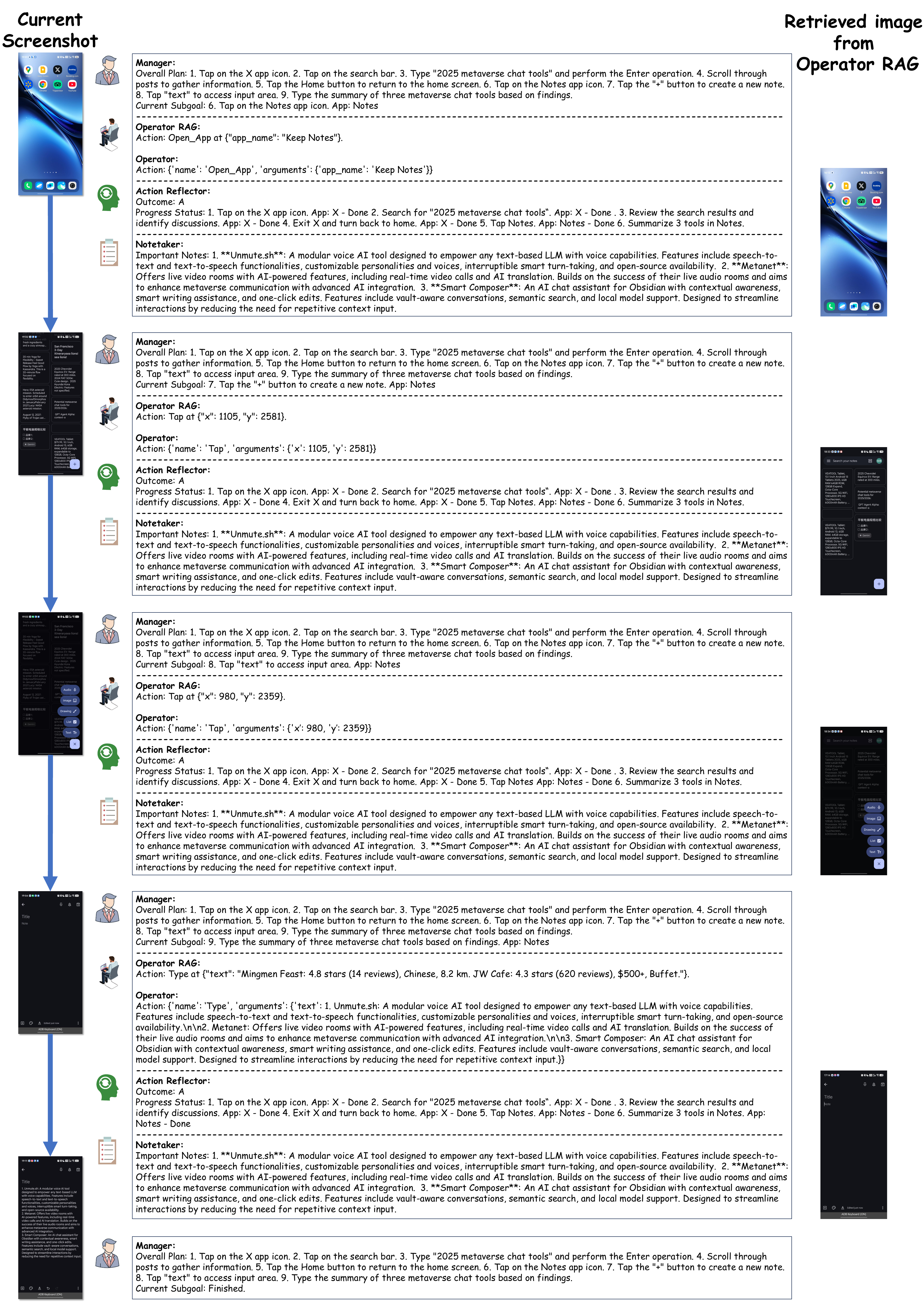}\caption{\small (Process continued from previous figure) An end-to-end execution case of a task. Each box represents an iterative loop, progressing chronologically from top to bottom. The ``\textbf{Current Screenshot}'' column on the left shows the mobile device's real-time UI screen state. The central section details the inter-agent communication. The ``\textbf{Manager}'' section presents the overall plan and the current subtask. The ``\textbf{Operator-RAG}'' section shows the best-matched action retrieved from its knowledge base for the current subtask, while the ``Retrieved image from Operator-RAG'' on the right provides reference screenshot. The ``\textbf{Operator}'' then synthesizes this information to generate a precise action with its argument. After that, The ``\textbf{Action Reflector}'' records the outcome and updates the progress status. Finally, the ``\textbf{Notetaker}'' records important information as ``Important Notes'' for future context. This process clearly depicts how the agents, through a ``\textbf{Planning\,→\,Execution\,→\,Reflection}'' loop, retrieval augmentation, step-by-step execution, and continuous feedback, progressively complete complex, cross-app tasks.}\label{fig:cs2}\end{figure*}

\begin{figure*}[t]\centering\includegraphics[width=\textwidth]{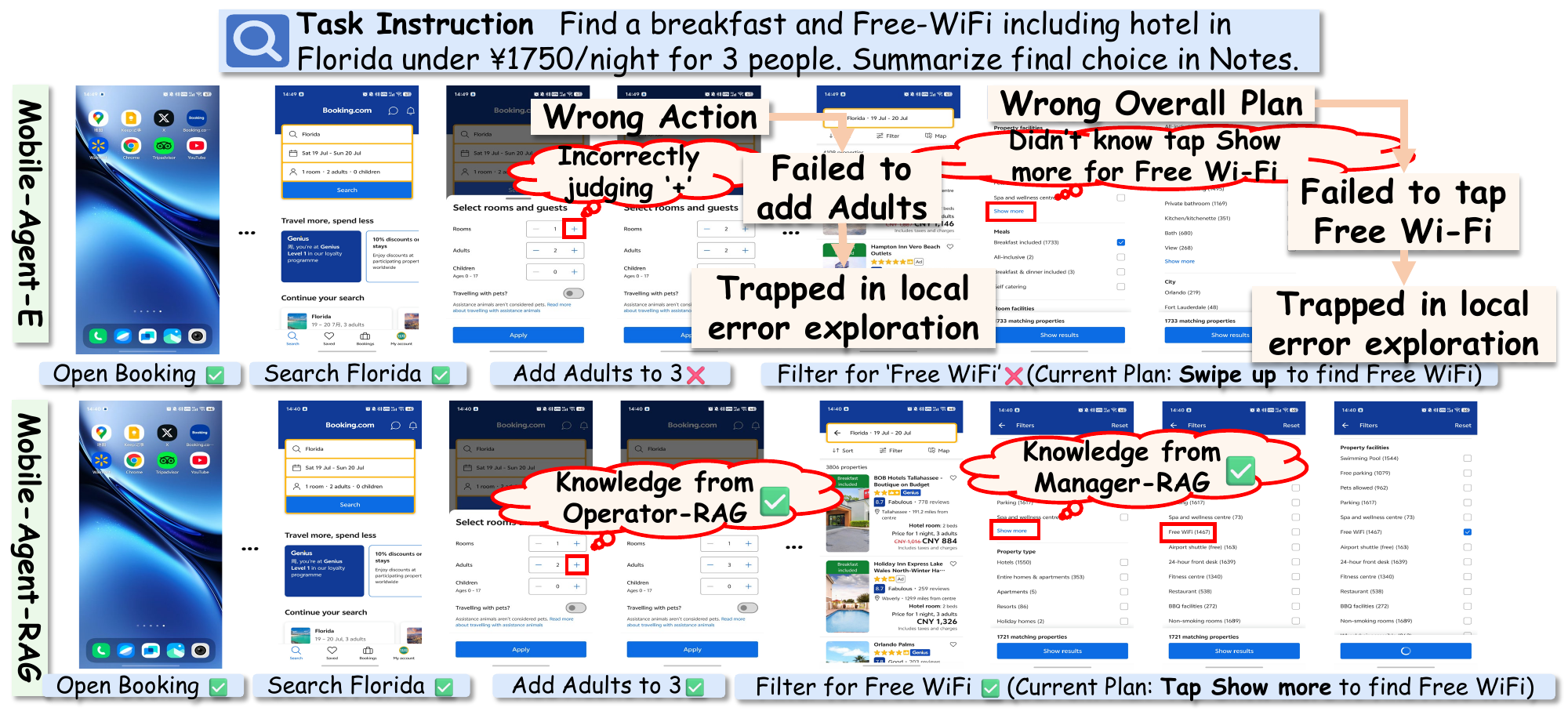}\caption{Qualitative comparison on a challenging cross-app task. Without retrieval augmentation, \texttt{Mobile-Agent-E} misidentifies visually similar buttons, enters local exploration, and requires a greater number of atomic steps and replans, which impacts task execution efficiency. Augmented by \textbf{Manager-RAG} and \textbf{Operator-RAG}, \texttt{Mobile-Agent-RAG} completes the same task with precise and decisive actions, demonstrating higher action accuracy, plan coherence, and overall efficiency.}\label{fig:cs3}\end{figure*}

In this section, we perform case studies to evaluate \texttt{Mobile-Agent-RAG} in a multi-app mobile setting that mirrors real-world ``app-hopping'' scenarios. 

\paragraph{Case Study of End-to-End Task Execution} In this case, we provide a comprehensive, step-by-step walkthrough of how the agent completes a complex long-horizon cross-app task on a real-world smartphone in Figures~\ref{fig:cs1} and~\ref{fig:cs2}. Throughout the process, the system repeatedly follows an iterative ``\textbf{Planning\,→\,Execution\,→\,Reflection}'' loop, as illustrated below.\par

\begin{itemize}
    \item \textbf{Phase 1 - Planning (\textit{Manager Agent + Manager-RAG})}
Before any UI action is issued, Manager-RAG intervenes; guided by the overall task objective, it queries its knowledge base and retrieves relevant Manager-RAG documents (To ensure a clear demonstration in the figure, we set $k=1$ in this case.) that contains the task instruction together with the corresponding human steps, as shown in the ``Manager-RAG'' section of the figure. The exemplar constrains the search space, steers the Manager's high-level strategy, and prevents sub-optimal plans.\

    \item \textbf{Phase 2 - Execution (\textit{Operator Agent + Operator-RAG})}
After the Manager provides the high-level plan and current sub-goal, the Operator must emit atomic UI actions to satisfy that sub-goal. It first captures the live screenshot shown on the left under ``Current Screenshot''; it then consults Operator-RAG, which returns the top-1 (subtask, screenshot, action) triplet. The screenshot is located under the ``Retrieved Image from Operator RAG'' section on the right, while the retrieved action guidance is within the ``Operator-RAG'' section. By fusing the live screenshot with the retrieved exemplar, the Operator accurately locates the target UI widget and generates a precise atomic action with all required arguments. The full list of actions and their arguments is enumerated in the ``Operator'' section of the figure.

    \item \textbf{Phase 3 - Reflection (\textit{Action Reflector})}
Once an atomic action completes, the Action Reflector evaluates the result by analysing the post-action screenshot and its fine-grained metadata, confirms whether the desired effect was achieved, updates the progress tracker, and decides whether to proceed, retry, or trigger a plan revision. The ``Action Reflector'' panel logs every verdict and the updated task status, forming a dynamic feedback loop that allows \texttt{Mobile-Agent-RAG} to adapt to real execution conditions.
\end{itemize}

\paragraph{Case-study Comparison with \texttt{Mobile-Agent-E}} To verify \texttt{Mobile-Agent-RAG}'s effectiveness of retrieval augmentation on complex tasks, we conduct a head-to-head case study, as shown in Figure~\ref{fig:cs3}. In this scenario, \texttt{Mobile-Agent-E}'s Operator frequently misidentifies visually similar buttons and drifts into local exploration, while its Manager occasionally produces sub-optimal plans on unseen screens, both of which lower overall efficiency. By introducing Operator-RAG and Manager-RAG—thereby injecting human-annotated experience—\texttt{Mobile-Agent-RAG} issues more accurate atomic actions and maintains a coherent plan, significantly accelerating execution and demonstrating improved robustness and generalisation.\par

\section{M. More Analysis and Limitations}

This section presents a deeper analysis of the mechanisms that drive the effectiveness of \texttt{Mobile-Agent-RAG}, identifies key limitations observed during experimentation, and compares the retrieval-based framework with evolutionary approaches. These insights provide guidance for future system improvements and the exploration of hybrid designs.

\paragraph{How Retrieval Enhances Planning and Execution}

The strong performance of \texttt{Mobile-Agent-RAG} is rooted in its ability to ground agent behavior in verified human trajectories, which substantially mitigates the hallucination problem that commonly affects autonomous agents.
\begin{itemize}
    \item \textbf{Manager-RAG} strengthens high-level planning by retrieving similar task patterns that offer strategic guidance. These examples effectively reduce the planning search space and prevent the agent from engaging in inefficient or suboptimal strategies. The retrieved task templates are not static—they are adapted to fit the nuances of novel task variations while maintaining consistency with proven strategies.

    \item \textbf{Operator-RAG} enhances low-level execution by retrieving contextually relevant examples of successful UI interactions. This is especially crucial in visually complex mobile environments, where subtle differences in UI layout can significantly impact action validity. The retrieved cases help the agent infer correct actions from the current visual context, providing robust grounding for precise operations.
\end{itemize}

\paragraph{Limitations and Failure Modes} Despite its overall effectiveness, \texttt{Mobile-Agent-RAG} reveals two primary types of failure that highlight important limitations:
\begin{itemize}
    \item \textbf{Limitations in Knowledge Base Coverage:} Failures may arise when the agent encounters task contexts or UI states that are underrepresented or missing from the retrieval corpus. These cases highlight the need for a more comprehensive and diverse knowledge base. Leveraging active learning to identify and fill such gaps could significantly enhance retrieval effectiveness and overall system robustness.

    \item 
    \textbf{Challenges in Visual Perception:} In some cases, the agent struggles to interpret complex or unfamiliar UI layouts, even when relevant examples are successfully retrieved. These failures indicate that retrieval alone is insufficient to overcome fundamental visual understanding limitations, emphasizing the need for more capable visual perception modules to support retrieval-based reasoning.

\end{itemize}
